\documentclass{article}

\usepackage[T1]{fontenc}

\usepackage{amssymb}

\usepackage{amsmath}

\usepackage[inline]{enumitem}

\usepackage[usenames,dvipsnames]{xcolor}
\definecolor{shadecolor}{gray}{0.9}

\usepackage{graphicx}
\usepackage[labelfont=bf]{caption}
\usepackage[format=hang]{subcaption}

\usepackage{booktabs, array, multirow}
\usepackage{wrapfig}

\usepackage{natbib}
\usepackage[colorlinks,linktoc=all]{hyperref}
\usepackage[all]{hypcap}
\hypersetup{citecolor=MidnightBlue}
\hypersetup{linkcolor=MidnightBlue}
\hypersetup{urlcolor=MidnightBlue}
\usepackage[nameinlink]{cleveref}
\creflabelformat{equation}{#2\textup{#1}#3}  

\usepackage
[acronym,nowarn,section,nogroupskip,nonumberlist]{glossaries}
\glsdisablehyper{}

\input{maths_preamble}
\usepackage{tikz}
\usetikzlibrary{positioning, arrows.meta, shapes.multipart}

\tikzstyle{mhsa}=[rectangle, draw, align=center, rounded corners, fill=LimeGreen!60]
\tikzstyle{mhca}=[rectangle, draw, align=center, rounded corners, fill=SkyBlue!60]
\tikzstyle{mha}=[rectangle, draw, align=center, rounded corners, fill=BurntOrange!60]
\tikzstyle{mlp}=[rectangle, draw, align=center, rounded corners, fill=BurntOrange!60]
\tikzstyle{arrow}=[->, >=Stealth, shorten >=2pt, shorten <=2pt]
\tikzstyle{every node}=[font=\small]

\usepackage{pifont}
\newcommand{\cmark}{\ding{51}}%
\newcommand{\xmark}{\ding{55}}%

\usepackage{microtype}

\usepackage[accepted]{icml2024}

\theoremstyle{plain}
\newtheorem{theorem}{Theorem}[section]

\theoremstyle{definition}

\theoremstyle{remark}

\icmltitlerunning{Translation Equivariant Transformer Neural Processes}

\newacronym[longplural={transformer neural processes}]{tnp}{TNP}{transformer neural process}
\newacronym{te-tnp}{TE-TNP}{translation equivariant TNP}
\newacronym{pt-tnp}{PT-TNP}{pseudo-token TNP}
\newacronym{te-pt-tnp}{TE-PT-TNP}{translation equivariant PT-TNP}
\newacronym[longplural={neural processes}]{np}{NP}{neural process}
\newacronym{cnp}{CNP}{conditional NP}
\newacronym{anp}{ANP}{attentive NP}
\newacronym{tnp-d}{TNP-D}{diagonal TNP}
\newacronym{tnp-ar}{TNP-AR}{autoregressive TNP}
\newacronym{tnp-nd}{TNP-ND}{non-diagonal TNP}
\newacronym{eq-tnp}{EQTNP}{efficient-query TNP}
\newacronym{lbanp}{LBANP}{latent-bottlenecked ANP}
\newacronym{ist}{IST}{induced set transformer}
\newacronym{convcnp}{ConvCNP}{convolutional conditional NP}
\newacronym{rcnp}{RCNP}{relational CNP}
\newacronym{mhsa}{MHSA}{multi-head self attention}
\newacronym{mhca}{MHCA}{multi-head cross attention}
\newacronym{te-mhsa}{TE-MHSA}{translation equivariant multi-head self attention}
\newacronym{te-mhca}{TE-MHCA}{translation equivariant multi-head cross attention}

\begin{document}

\twocolumn[
\icmltitle{Translation Equivariant Transformer Neural Processes}



\icmlsetsymbol{equal}{*}

\begin{icmlauthorlist}
\icmlauthor{Matthew Ashman}{cam}
\icmlauthor{Cristiana Diaconu}{cam}
\icmlauthor{Junhyuck Kim}{cam}
\icmlauthor{Lakee Sivaraya}{cam}
\icmlauthor{Stratis Markou}{cam}
\icmlauthor{James Requeima}{toronto}
\icmlauthor{Wessel P.\ Bruinsma}{microsoft}
\icmlauthor{Richard E.\ Turner}{cam,microsoft}
\end{icmlauthorlist}

\icmlaffiliation{cam}{Department of Engineering, University of Cambridge, Cambridge, UK}
\icmlaffiliation{toronto}{Vector Institute, University of Toronto, Toronto, Canada}
\icmlaffiliation{microsoft}{Microsoft Research AI for Science, Cambridge, UK}

\icmlcorrespondingauthor{Matthew Ashman}{mca39@cam.ac.uk}

\icmlkeywords{Machine Learning, ICML}

\vskip 0.3in
]



\printAffiliationsAndNotice{}  

\begin{abstract}
The effectiveness of \glspl{np} in modelling posterior prediction maps---the mapping from data to posterior predictive distributions---has significantly improved since their inception. This improvement can be attributed to two principal factors: 
\begin{enumerate*}[label={(\arabic*)}]
    \item advancements in the architecture of permutation invariant set functions, which are intrinsic to all \glspl{np}; and
    \item leveraging symmetries present in the true posterior predictive map, which are problem dependent.
\end{enumerate*} 
Transformers are a notable development in permutation invariant set functions, and their utility within \glspl{np} has been demonstrated through the family of models we refer to as \glspl{tnp}. Despite significant interest in \glspl{tnp}, little attention has been given to incorporating symmetries. Notably, the posterior prediction maps for data that are stationary---a common assumption in spatio-temporal modelling---exhibit translation equivariance. In this paper, we introduce of a new family of \glspl{te-tnp} that incorporate \emph{translation equivariance}. Through an extensive range of experiments on synthetic and real-world spatio-temporal data, we demonstrate the effectiveness of \glspl{te-tnp} relative to their non-translation-equivariant counterparts and other \gls{np} baselines.

\end{abstract}

\section{Introduction}
Transformers have emerged as an immensely effective architecture for natural language processing and computer vision tasks \citep{vaswani2017attention,dosovitskiy2020image}. They have become the backbone for many state-of-the-art models---such ChatGPT \citep{achiam2023gpt} and DALL-E \citep{betker2023improving}---owing to their ability to learn complex dependencies amongst input data.
More generally, transformers can be understood as permutation equivariant set functions. This abstraction has led to the deployment of transformers in domains beyond that of sequence modelling, including particle physics, molecular modelling, climate science, and Bayesian inference \citep{lee2019set,fuchs2020se,muller2021transformers}.


\glspl{np} \citep{garnelo2018conditional, garnelo2018neural} are a broad family of meta-learning models which learn the mapping from sets of observed datapoints to predictive stochastic processes \citep{foong2020meta}. They are straightforward to train, handle off-the-grid data and missing observations with ease, and can be easily adapted for different data modalities. 
This flexibility makes them an attractive choice for a wide variety of problem domains, including spatio-temporal modelling, healthcare, and few-shot learning \citep{jha2022neural}. 
Exchangeability in the predictive distribution with respect to the context set is achieved through the use of permutation invariant set functions, which, in \glspl{np}, map from the sets of observations to some representation space. Given the utility of transformers as set functions, it is natural to consider their use within \glspl{np}. This gives rise to \glspl{tnp}.

The family of \glspl{tnp} include the \gls{anp} \citep{kim2019attentive}, \gls{tnp-d}, \gls{tnp-ar}, and \gls{tnp-nd} \citep{nguyen2022transformer}, and the \gls{lbanp} \citep{kim2019attentive}. 
Despite a significant amount of interest in \glspl{tnp} from the research community, there are certain properties that we may wish our model to possess that have not yet been addressed. In particular, for spatio-temporal problems the data is often roughly stationary, in which case it is desirable to equip our model with translation equivariance: if the data are translated in space or time, then the predictions of our model should be translated correspondingly. 
Although translation equivariance has been incorporated into other families of \gls{np} models, such as the \gls{convcnp} \citep{gordon2019convolutional} and \gls{rcnp} \citep{huang2023practical}, it is yet to be incorporated into the \gls{tnp}. The key ingredient to achieving this is to establish effective translation equivariant attention layers that can be used in place of the standard attention layers within the transformer encoder.
In this paper, we develop the \gls{te-tnp}.  
Our contributions are as follows:
\begin{enumerate}[leftmargin=15pt]
    \item We develop an effective method for incorporating translation equivariance into the attention mechanism of transformers, developing the \gls{te-mhsa} and \gls{te-mhca} operations. These operations replace standard MHSA and MHCA operations within transformer encoders to obtain a new family of translation equivariant \glspl{tnp}.
    \item We use pseudo-tokens to reduce the quadratic computational complexity of \glspl{te-tnp}, developing \glspl{te-pt-tnp}.
    \item We demonstrate the efficacy of \glspl{te-tnp} relative to existing NPs---including the \gls{convcnp} and the \gls{rcnp}---on a number of synthetic and real-world spatio-temporal modelling problems.
\end{enumerate}

\section{Background}
\label{sec:background}
Throughout this section, we will use the following notation. Let $\mcX = \R^{D_x}$, $\mcY = \R^{D_y}$ denote the input and output spaces, and let $(\bfx, \bfy) \in \mcX \times \mcY$ denote an input--output pair. Let $\mcS = \bigcup_{N=0}^\infty (\mcX \times \mcY)^N$ be a collection of all finite data sets, which includes the empty set $\varnothing$, the data set containing no data points.
We denote a context and target set with $\mcD_c,\ \mcD_t \in \mcS$, where $|\mcD_c| = N_c$, $|\mcD_t| = N_t$.
Let $\bfX_c \in \R^{N_c \times D_x}$, $\bfY_c \in \R^{N_c \times D_y}$ be the inputs and corresponding outputs of $\mcD_c$, with $\bfX_t \in \R^{N_t \times D_x}$, $\bfY_t \in \R^{N_t \times D_y}$ defined analogously. We denote a single task as $\xi = (\mcD_c, \mcD_t) = ((\bfX_c, \bfY_c), (\bfX_t, \bfY_t))$. Let $\mcP(\mcX)$ denote the collection of stochastic processes on $\mcX$. 

\subsection{Neural Processes}
\label{subsec:neural-processes}
\glspl{np} \citep{garnelo2018conditional, garnelo2018neural} aim to learn the mapping from context sets $\mcD_c$ to ground truth posterior distributions over the target outputs, $\mcD_c \mapsto p(\bfY_t | \bfX_t, \mcD_c)$, using meta-learning. This mapping is known as the \emph{posterior prediction map} $\pi_P: \mcS \rightarrow \mcP(\mcX)$, where $P$ denotes the ground truth stochastic process over functions mapping from $\mcX$ to $\mcY$.
Common to all \gls{np} architectures is an encoder and decoder. The encoder maps from $\mcD_c$ and $\bfX_t$ to some representation, $e(\mcD_c, \bfX_t)$.\footnote{In many \gls{np} architectures, including the original \gls{cnp} and \gls{np}, the representation does not depend on the target inputs $\bfX_t$.} The decoder takes as input the representation and target inputs $\bfX_t$ and outputs $d(\bfX_t, e(\mcD_c, \bfX_t))$, which are the parameters of the predictive distribution over the target outputs $\bfY_t$: $p(\bfY_t | \bfX_t, \mcD_c) = p(\bfY_t | d(\bfX_t, e(\mcD_c, \bfX_t)))$.
An important requirement of the predictive distribution is permutation invariance with respect to the elements of $\mcD_c$.
We shall focus on \glspl{cnp} \citep{garnelo2018conditional}, which factorise the predictive distribution as $p(\bfY_t | \bfX_t, \mcD_c) = \prod_{n=1}^{N_t} p(\bfy_{t, n} | d(\bfx_{t, n}, e(\mcD_c, \bfx_{t, n})))$. \glspl{cnp} are trained by maximising the posterior predictive likelihood:
\begin{equation} \textstyle
    \label{eq:cnp-objective}
    \!\!\mcL_{\text{ML}} \!=\! \mathbb{E}_{p(\xi)}\Big[\!\sum_{n=1}^{N_t} \log p(\bfy_{t, n} | d(\bfx_{t, n}, e(\mcD_c, \bfx_{t, n})))\Big].
\end{equation}
Here, the expectation is taken with respect to the distribution of tasks $p(\xi)$. As shown in \citet{foong2020meta}, the global maximum is achieved if and only if the model recovers the ground-truth posterior prediction map.
When training a \gls{cnp}, we often approximate the expectation with an average over the finite number of tasks available.

\subsection{Transformers}
\label{subsec:transformers}
A useful perspective is to understand transformers as a permutation equivariant set function $f$.\footnote{Note that not all permutation equivariant set functions can be represented by transformers. For example, the family of informers \citep{garnelo2023exploring} cannot be represented by transformers, yet are permutation equivariant set functions. However, transformers are universal approximators of permutation equivariant set functions \citep{lee2019set,wagstaff2022universal}.} They take as input a set of $N$ tokens, $\bfZ^0 \in \R^{N\times D_z}$, output a set of $N$ tokens of the same cardinality: $f\colon (\R^{D_z})^N \rightarrow (\R^{D_z})^N$. If the input set is permuted, then the output set is permuted accordingly:
$f(\bfz_1, \ldots, \bfz_N)_n = f(\bfz_{\sigma(1)}, \ldots, \bfz_{\sigma(N)})_{\sigma(n)}$
for all permutations $\sigma \in \mbbS^N$ of $N$ elements. At the core of each layer of the transformer architecture is the \gls{mhsa} operation \citep{vaswani2017attention}. Let $\bfZ^{\ell} \in \R^{N\times D_z}$ denote the input set to the $\ell$-th MHSA operation. The MHSA operation updates the $n$\textsuperscript{th} token $\bfz^{\ell}_n$ as
\begin{equation}\textstyle
\label{eq:self-attention}
    \!\!\tilde{\bfz}^{\ell}_n \!=\! \operatorname{cat}\!\Big(\Big\{\sum_{m=1}^N \alpha^{\ell}_h(\bfz^{\ell}_n, \bfz^{\ell}_m) {\bfz^{\ell}_m\!}^T \bfW^{\ell}_{V, h}\Big\}_{h=1}^{H^{\ell}}\Big)\bfW^{\ell}_O
\end{equation}
where $\operatorname{cat}$ denotes the concatenation operation across the last dimension. Here, $\bfW^{\ell}_{V, h} \in \R^{D_z\times D_V}$ and $\bfW^{\ell}_{O} \in \R^{H^{\ell}D_V \times D_z}$ are the value and projection weight matrices, where $H^{\ell}$ denotes the number of `heads' in layer $\ell$. Note that permutation equivariance is achieved through the permutation invariant summation operator. As this is the only mechanism through which the tokens interact with each other, permutation equivariance for the overall model is ensured. The attention mechanism, $\alpha^{\ell}_h$, is implemented as
\begin{equation}
\label{eq:attention-mechanism}
    \alpha^{\ell}_h(\bfz^{\ell}_n, \bfz^{\ell}_m) = \frac{e^{{\bfz^{\ell}_n}^T\bfW^{\ell}_{Q, h}\left[\bfW^{\ell}_{K, h}\right]^T\bfz^{\ell}_m}}{\sum_{m=1}^N e^{{\bfz^{\ell}_n}^T\bfW^{\ell}_{Q, h}\left[\bfW^{\ell}_{K, h}\right]^T\bfz^{\ell}_m}}
\end{equation}
where $\bfW^{\ell}_{Q, h} \in \R^{D_z \times D_{QK}}$ and $\bfW^{\ell}_{K, h} \in \R^{D_z\times D_{QK}}$ are the query and key weight matrices. The softmax-normalisation ensures that $\sum_{m=1}^N \alpha^{\ell}_h(\bfz^{\ell}_n, \bfz^{\ell}_m) = 1 \ \forall n, h, \ell$.
Often, conditional independencies amongst the set of tokens---in the sense that the set $\{\bfz^{\ell}_n\}^{\ell=L}_{\ell=1}$ do not depend on the set $\{\bfz^{\ell}_m\}^{\ell=L}_{\ell=1}$ given some other set of tokens for some $n,\ m \in \{1, \ldots, N\}$---are desirable. Whilst this is typically achieved through masking, if the same set of tokens are conditioned on for every $n$, then it is more computationally efficient to use \gls{mhca} operations together with \gls{mhsa} operations than it is to directly compute \Cref{eq:self-attention}. The \gls{mhca} operation updates the $n$\textsuperscript{th} token $\bfz^{\ell}_n$ using the set of tokens $\{\hat{\bfz}^{\ell}_m\}_{m=1}^{M}$ as 
\begin{equation}\textstyle
\label{eq:mhca}
    \!\!\tilde{\bfz}^{\ell}_n \!=\! \operatorname{cat}\!\Big(\Big\{\sum_{m=1}^M \alpha^{\ell}_h(\bfz^{\ell}_n, \hat{\bfz}^{\ell}_m) {\hat{\bfz}^{\ell}_m\!}^T \bfW^{\ell}_{V, h}\Big\}_{h=1}^{H^{\ell}}\Big)\bfW^{\ell}_O.
\end{equation}
Note that all tokens updated in this manner are conditionally independent of each other given $\{\hat{\bfz}^{\ell}_m\}_{m=1}^{M}$.
We discuss this in more detail in \Cref{app:masked-attention}. \gls{mhca} operations are at the core of the pseudo-token-based transformers such as the perceiver \citep{jaegle2021perceiver} and \gls{ist} \citep{lee2019set}. We describe these differences in the following section.

\gls{mhsa} and \gls{mhca} operations are used in combination with layer-normalisation operations and pointwise MLPs to obtain \gls{mhsa} and \gls{mhca} blocks. Unless stated otherwise, we shall adopt the order used by \citet{vaswani2017attention}.

\subsection{Pseudo-Token-Based Transformers}
\label{subsec:pseudo-tokens}
Pseudo-token based transformers reduce the quadratic computational complexity of the standard transformer through the use of pseudo-tokens. Concretely, let $\bfU \in \R^{M \times D_z}$ denote an initial set of $M \ll N$ tokens we call pseudo-tokens. There are two established methods for incorporating information about the set of observed tokens ($\bfZ)$ into these pseudo-tokens in a computationally efficient manner: the perceiver-style approach of \citet{jaegle2021perceiver} and the \gls{ist} style approach of \citet{lee2019set}. The perceiver-style approach iterates between applying $\operatorname{MHCA}(\bfU^{\ell}, \bfZ^{\ell})$ and $\operatorname{MHSA}(\bfU^{\ell})$, outputting a set of $M$ pseudo-tokens, and has a computational complexity of $\order{MN}$ at each layer. The \gls{ist}-style approach iterates between applying $\operatorname{MHCA}(\bfU^{\ell}, \bfZ^{\ell})$ and $\operatorname{MHCA}(\bfZ^{\ell}, \bfU^{\ell})$, outputting a set of $N$ tokens and $M$ pseudo-tokens, and also has a computational complexity of $\order{MN}$ at each layer. We provide illustrations these differences \Cref{app:pseudo-token-transformers}.

\subsection{Transformer Neural Processes}
\label{subsec:tnps}
Given the utility of transformers as set functions, it is natural to consider their use in the encoder of a \gls{np}---we describe this family of \glspl{np} as \glspl{tnp}. Let $\bfZ^0_c \in \R^{N_c \times D}$ denote the initial set-of-token representation of each input-output pair $(\bfx_{c, n}, \bfy_{c, n}) \in \mcD_c$, and $\bfZ^0_{t, n} \in \R^{N_t \times D}$ denote the initial set-of-token representation of each input $\bfx_{t, n} \in \bfX_t$. The encoding $e(\mcD_c, \bfX_t)$ of \glspl{tnp} is is achieved by passing the union of initial context and target tokens, $\bfZ^0 = [\bfZ^0_c,\ \bfZ^0_t]$, through a transformer-style architecture, and keeping only the output tokens corresponding to the target inputs, $\bfZ^L_t$. 

The specific transformer-style architecture is unique to each \gls{tnp} variant. However, they generally consist of MHSA operations acting on the context tokens and MHCA operations acting to update the target tokens, given the context tokens.\footnote{As discussed in \Cref{subsec:transformers}, this is often implemented as a single MHSA operation with masking operating.} The combination of MHSA and MHCA operations is a permutation invariant function with respect to the context tokens. We provide an illustration of this in \Cref{fig:tnp}. Enforcing these conditional independencies ensures that the final target token $\bfz^L_{t, n}$ depends only on $\mcD_c$ and $\bfx_{t, n}$, i.e.\ $\left[e(\mcD_c, \bfX_t)\right]_n = e(\mcD_c, \bfx_{t, n})$. This is required for the factorisation of the predictive distribution $p(\bfY_t | \bfX_t, \mcD_c) = \prod_{n=1}^{N_t} p(\bfy_{t, n} | d(\bfx_{t, n}, e(\mcD_c, \bfx_{t, n})))$. We denote \glspl{pt-tnp} as the family of \glspl{tnp} which use pseudo-token based transformers. Currently, this family is restricted to the \gls{lbanp}, which uses a perceiver-style architecture; however, it is straightforward to use an IST-style architecture instead.

\subsection{Translation Equivariance}
\label{subsec:translation-equivariance}

\begin{figure*}[htb]
\begin{minipage}[b]{.6\textwidth}
    \centering
    \begin{subfigure}[t]{0.37\textwidth}
        \centering
        \resizebox{\textwidth}{!}{
        \begin{tikzpicture}[transform shape]
        \node[align=center] (dc) {$\mcD_c$};
        \node[align=center, right=2.5cm of dc] (dt) {$\bfX_t$};

        \node[mlp, above=0.5cm of dc] (context_encoder) {$\operatorname{MLP}(\bfX_c, \bfY_c)$};
        \node[mlp, above=0.5cm of dt] (target_encoder) {$\operatorname{MLP}(\bfX_t)$};

        \node[align=center, above=0.5cm of context_encoder] (context) {$\bfZ^0_c$};
        \node[align=center, above=0.5cm of target_encoder] (target) {$\bfZ^0_t$};
        
        \node[mhsa, above=0.5cm of context] (mhsa1) {$\operatorname{MHSA}(\bfZ^0_{c})$};
        \node[above=0.5cm of mhsa1] (mhsa_elipse) {\rotatebox{90}{...}};
        \node[mhsa, above=0.5cm of mhsa_elipse] (mhsa2) {$\operatorname{MHSA}(\bfZ^{L - 1}_{c})$};
    
        \node[mhca, above=0.5cm of target] (mhca1) {$\operatorname{MHCA}(\bfZ^0_{t}, \bfZ^1_{c})$};
        \node[above=0.5cm of mhca1] (mhca_elipse) {\rotatebox{90}{...}};
        \node[mhca, above=0.5cm of mhca_elipse] (mhca2) {$\operatorname{MHCA}(\bfZ^{L-1}_{t}$, $\bfZ^L_{c})$};
        
        \node[align=center, above=0.5cm of mhca2] (output) {$e(\mcD_c, \bfX_t)$};
        
        \draw[arrow] (dc) -- (context_encoder);
        \draw[arrow] (dt) -- (target_encoder);
        \draw[arrow] (context_encoder) -- (context);
        \draw[arrow] (target_encoder) -- (target);
        \draw[arrow] (context) -- (mhsa1);
        \draw[arrow] (target) -- (mhca1);
        \draw[arrow](mhca2) -- (output);
        \foreach \i [evaluate={\next=int(\i+1)}] in {1,2}
        {
            \draw[arrow] (mhsa\i) -- (mhca\i);
            \ifnum\next<2
                \draw[arrow] (mhsa\i) -- (mhsa\next);
                \draw[arrow] (mhca\i) -- (mhca\next);
            \fi
        }
        \draw[arrow] (mhsa1) -- (mhsa_elipse);
        \draw[arrow] (mhsa_elipse) -- (mhsa2);
        \draw[arrow] (mhca1) -- (mhca_elipse);
        \draw[arrow] (mhca_elipse) -- (mhca2);
        
        
        \end{tikzpicture}
        }
        \caption{\gls{tnp}.}
        \label{fig:tnp}
    \end{subfigure}
    \hfill
    %
    \begin{subfigure}[t]{0.6\textwidth}
        \centering
        \resizebox{\textwidth}{!}{
        \begin{tikzpicture}[transform shape]
        
        \node[mhsa] (mhsa1) {$\operatorname{te-MHSA}(\bfZ^0_{c}, \bfX^0_c)$};
        \node[above=0.5cm of mhsa1] (mhsa_elipse) {\rotatebox{90}{...}};
        \node[mhsa, above=0.5cm of mhsa_elipse] (mhsa2) {$\operatorname{te-MHSA}(\bfZ^{L - 1}_{c}, \bfX^{L-1}_c)$};
    
        \node[mhca, right=1.5cm of mhsa1] (mhca1) {$\operatorname{te-MHCA}(\bfZ^0_{t}, \bfZ^1_{c}, \bfX^0_t, \bfX^0_c)$};
        \node[above=0.5cm of mhca1] (mhca_elipse) {\rotatebox{90}{...}};
        \node[mhca, above=0.5cm of mhca_elipse] (mhca2) {$\operatorname{te-MHCA}(\bfZ^{L-1}_{t}$, $\bfZ^L_{c}, \bfX^{L-1}_t, \bfX^{L-1}_c)$};
        
        \node[align=center, above=0.5cm of mhca2] (output) {$e(\mcD_c, \bfX_t)$};


        \coordinate[left=0.75cm of mhsa1.south] (mhsa1_south_west);
        \coordinate[right=0.75cm of mhsa1.south] (mhsa1_south_east);
        \coordinate[left=0.75cm of mhsa1.north] (mhsa1_north_west);
        \coordinate[right=0.75cm of mhsa1.north] (mhsa1_north_east);
        \coordinate[left=0.75cm of mhsa2.south] (mhsa2_south_west);
        \coordinate[right=0.75cm of mhsa2.south] (mhsa2_south_east);
        \coordinate[left=0.75cm of mhsa_elipse.south] (mhsa_elipse_south_west);
        \coordinate[right=0.75cm of mhsa_elipse.south] (mhsa_elipse_south_east);
        \coordinate[left=0.75cm of mhsa_elipse.north] (mhsa_elipse_north_west);
        \coordinate[right=0.75cm of mhsa_elipse.north] (mhsa_elipse_north_east);

        \draw[arrow] (mhsa1_north_west) -- (mhsa_elipse_south_west);
        \draw[arrow] (mhsa1_north_east) -- (mhsa_elipse_south_east);
        \draw[arrow] (mhsa_elipse_north_west) -- (mhsa2_south_west);
        \draw[arrow] (mhsa_elipse_north_east) -- (mhsa2_south_east);

        \coordinate[left=0.75cm of mhca1.south] (mhca1_south_west);
        \coordinate[right=0.75cm of mhca1.south] (mhca1_south_east);
        \coordinate[left=0.75cm of mhca1.north] (mhca1_north_west);
        \coordinate[right=0.75cm of mhca1.north] (mhca1_north_east);
        \coordinate[left=0.75cm of mhca2.south] (mhca2_south_west);
        \coordinate[right=0.75cm of mhca2.south] (mhca2_south_east);
        \coordinate[left=0.75cm of mhca_elipse.south] (mhca_elipse_south_west);
        \coordinate[right=0.75cm of mhca_elipse.south] (mhca_elipse_south_east);
        \coordinate[left=0.75cm of mhca_elipse.north] (mhca_elipse_north_west);
        \coordinate[right=0.75cm of mhca_elipse.north] (mhca_elipse_north_east);

        \draw[arrow] (mhca1_north_west) -- (mhca_elipse_south_west);
        \draw[arrow] (mhca1_north_east) -- (mhca_elipse_south_east);
        \draw[arrow] (mhca_elipse_north_west) -- (mhca2_south_west);
        \draw[arrow] (mhca_elipse_north_east) -- (mhca2_south_east);

        \node[align=center, below=0.5cm of mhsa1_south_west] (context) {$\bfZ^0_c$};
        \node[align=center, below=0.5cm of mhca1_south_west] (target) {$\bfZ^0_t$};

        \node[mlp, below=0.5cm of context] (context_encoder) {$\operatorname{MLP}(\bfY_c)$};

        \node[align=center, below=0.5cm of context_encoder] (dc) {$\bfY_c$};

        \node[align=center, below=0.5cm of mhsa1_south_east] (context_inputs) {$\bfX^0_c = \bfX_c$};
        \node[align=center, below=0.5cm of mhca1_south_east] (target_inputs) {$\bfX^0_t = \bfX_t$};
        
        \draw[arrow] (dc) -- (context_encoder);
        \draw[arrow] (context_encoder) -- (context);
        \draw[arrow] (context) -- (mhsa1_south_west);
        \draw[arrow] (target) -- (mhca1_south_west);
        \draw[arrow] (context_inputs) -- (mhsa1_south_east);
        \draw[arrow] (target_inputs) -- (mhca1_south_east);
        \draw[arrow](mhca2) -- (output);
        \foreach \i [evaluate={\next=int(\i+1)}] in {1,2}
        {
            \draw[arrow] (mhsa\i) -- (mhca\i);
        }
        
        \end{tikzpicture}
        }
        \caption{\gls{te-tnp}.}
        \label{fig:te-tnp}
    \end{subfigure}
    \caption{Block diagrams illustrating the \gls{tnp} and \gls{te-tnp} encoder architectures. For both models, we pass individual datapoints through pointwise MLPs to obtain the initial token representations, $\bfZ^0_c$ and $\bfZ^0_t$. These are then passed through multiple attention layers, with the context tokens interacting with the target tokens through cross-attention. The output of the encoder depends on $\mcD_c$ \emph{and} $\bfX_t$. The \gls{te-tnp} encoder updates the input locations at each layer, in addition to the tokens.}
    \label{fig:encoder-comparison}
\end{minipage}
\hfill
\begin{minipage}[b]{0.377\textwidth}
    \centering
    \includegraphics[width=\linewidth]{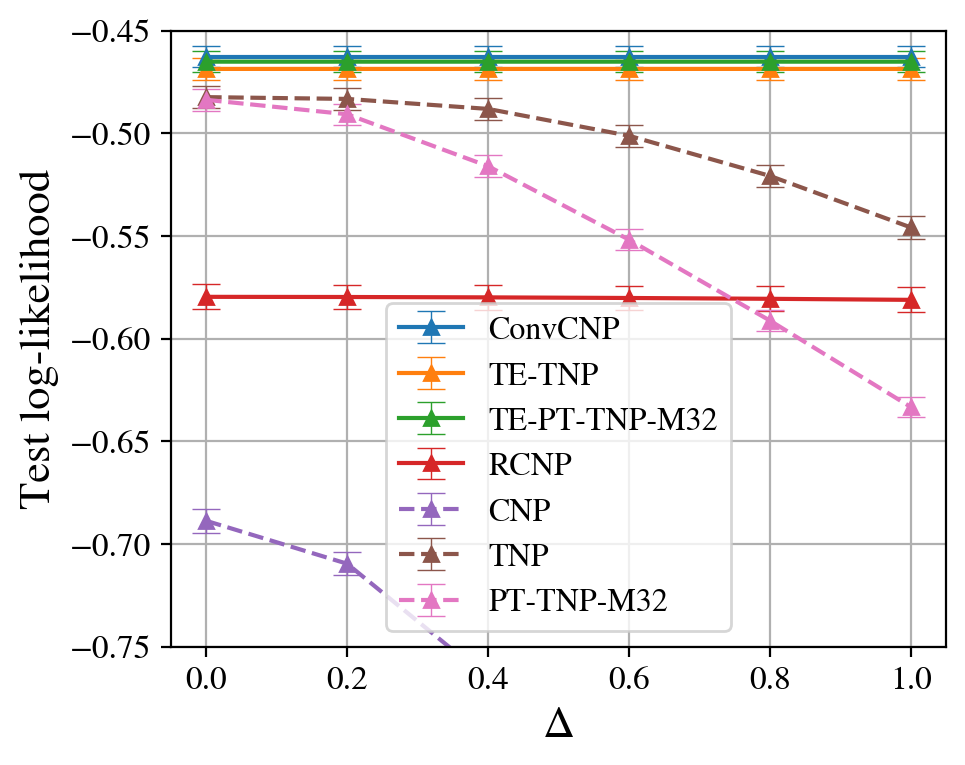}
    \caption{Average log-likelihood (\textcolor{OliveGreen}{$\*\uparrow$}) on the test datasets for the synthetic 1-D regression experiment. $\Delta$ denotes the amount by which the range from which the context and target inputs and sampled from is shifted at test time. Standard errors are shown.}
    \label{fig:1d-regression-results}
\end{minipage}
\vspace{-1em}
\end{figure*}
Here, we provide new theoretical results which show the importance of translation equivariance as an inductive bias in \glspl{np}. In particular, we first show that \emph{if, and only if,} the ground-truth stochastic process is stationary, the corresponding predictive map is translation equivariant (\Cref{thm:te_iff_stat}). Second, we show the importance of translation equivariance in the ability of our models to generalise to settings outside of the training distribution (\Cref{thm:generalisation}), for which \Cref{fig:generalisation} provides some intuition.

Let $\mathsf{T}_{\*\tau}$ denote a translation by $\*\tau \in \R^{D_x}$.
For a data set $\mathcal{D} \in \mathcal{S}$, $\mathsf{T}_{\*\tau} \mathcal{D} \in \mathcal{S}$ translates the data set by adding $\*\tau$ to all inputs.
For a function $f\colon \mcX \to Z$, $\mathsf{T}_{\*\tau} f$ translates $f$ by producing a new function $\mcX \to Z$ such that $\mathsf{T}_{\*\tau} f(\bfx) = f(\bfx - \*\tau)$ for all $\bfx \in \R^{D_x}$.
For a stochastic process $\mu \in \mcP(\mcX)$, $\mathsf{T}_{\*\tau}(\mu)$ denotes the pushforward measure of pushing $\mu$ through $\mathsf{T}_{\*\tau}$.
A \emph{prediction map} $\pi$ is a mapping $\pi \colon\mcS \to \mcP(\mcX)$ from data sets $\mcS$ to stochastic processes $\mcP(\mcX)$.
Prediction maps are mathematical models of neural processes.
Say that \emph{a prediction map $\pi$ is translation equivariant} if $\mathsf{T}_{\*\tau} \circ \pi = \pi \circ \mathsf{T}_{\*\tau}$ for all translations $\*\tau \in \R^{D_x}$.

The ground-truth stochastic process $P$ is stationary if and only if the prediction map $\pi_P$ is translation equivariant.
\Citet{foong2020meta} provide a simple proof of the ``only if''-direction.
We provide a rigorous proof in both directions.
Consider $\mcD \in \mcS$.
Formally define $\pi_P(\mcD)$ by integrating $P$ against a density $\pi_P'(\mcD)$ that depends on $\mcD$, so
$\mathrm{d} \pi_P(\mcD) = \pi_P'(\mcD)\,\mathrm{d} P$.\footnote{
    Intuitively, $\pi_P'(\mcD)(f) = p(\mcD | f) / p(\mcD)$, so $\pi_P'(\mcD)(f)$ is the modelling assumption that specifies how observations are generated from the ground-truth stochastic process.
    A simple example is $\pi_P'(\mcD)(f) \propto \prod_{(\bfx, \bfy) \in \mcD} \mathcal{N}(\bfy\mid f(\bfx),\sigma^2)$, which adds independent Gaussian noise with variance $\sigma^2$.
}
Assume that $\pi'_P(\varnothing) \propto 1$, so $\pi_P(\varnothing) = P$.
Say that $\pi_P'$ is \emph{translation invariant} if, for all $\mcD \in \mcS$ and $\*\tau \in \mcX$, $\pi_P'(\mathsf{T}_{\*\tau} \mcD) \circ \mathsf{T}_{\*\tau} = \pi_P'(\mcD)$ $P$--almost surely.\footnote{
For example, the usual Gaussian likelihood is translation invariant:
$
    \mathcal{N}(\bfy\mid (\mathsf{T}_{\*\tau} f)(\bfx + \*\tau), \sigma^2)
    = \mathcal{N}(\bfy\mid f(\*\tau), \sigma^2).
$
}

\begin{theorem} \label{thm:te_iff_stat}
    (1) The ground-truth stochastic process $P$ is stationary and $\pi'_P$ is translation invariant if and only if
    (2) $\pi_P$ is translation equivariant.
\end{theorem}
\vspace{-0.5em}  

See \Cref{sec:spatial-generalisation} for the proof.
If the ground-truth stochastic process is stationary, it is helpful to build translation equivariance into the neural process:
this greatly reduces the model space to search over, 
which can significantly improve data efficiency \citep{foong2020meta}.
In addition, it is possible to show that translation equivariant \glspl{np} generalise spatially.
We formalise this in the following theorem, which we present in the one-dimensional setting ($D_x=1$) for notational simplicity, and we provide an illustration of these ideas in \Cref{fig:generalisation}.

\textbf{Definitions for theorem.}
For a stochastic process $f \sim \mu$ with $\mu \in \mcP(\mcX)$, for every $\bfx \in \R^N$, denote the distribution of $(f(x_1), \ldots, f(x_N))$ by $P_\bfx \mu$.
We now define the notion of the \emph{receptive field}.
For two vectors of inputs $\bfx_1 \in \R^{N_1}$, $\bfx_2 \in \R^{N_2}$, and $R >0$, let $\bfx_1|_{\bfx_2,R}$ be the subvector of $\bfx_1$ with inputs at most distance $R$ away from any input in $\bfx_2$.
Similarly, for a data set $\mcD = (\bfx, \bfy) \in \mcS$, let $\mcD|_{\bfx_2,R} \in \mcS$ be the subset of data points of $\mcD$ with inputs at most distance $R$ away from any input in $\bfx_2$.
With these definitions, say that a \emph{stochastic process $f \sim \mu$ with $\mu \in \mcP(\mcX)$ has receptive field $R > 0$} if, for all $N_1,N_2 \in \N$, $\bfx_1 \in \R^{N_1}$, and $\bfx_2 \in \R^{N_2}$,
$f(\bfx_2) \mid f(\bfx_1) \overset{\text{d}}{=} f(\bfx_2) \mid f(\bfx_1|_{\bfx_2,\frac12R}$).
Intuitively, $f$ only has local dependencies.
Moreover, say that a \emph{prediction map $\pi\colon \mcS \to \mcP(\mcX)$ has receptive field $R>0$} if, for all $\mcD \in \mcS$, $N\in \N$, and $\bfx \in \R^{N}$, $P_\bfx \pi(\mathcal{D}) = P_\bfx \pi(\mathcal{D}|_{\bfx,\frac12 R})$.
Intuitively, predictions by the neural process $\pi$ are only influenced by context points at most distance $\frac12 R$ away.\footnote{
For example, if $f$ is a Gaussian process with a kernel compactly supported on $[-\frac12 R, \frac12 R]$, then the mapping $\mathcal{D} \mapsto p(f \mid \mathcal{D})$ is a prediction map which (a) has receptive field $R$ and (b) maps to stochastic processes with receptive field $R$.
}

\begin{theorem} \label{thm:generalisation}
    Let $\pi_1, \pi_2\colon \mcS \rightarrow \mcP(\mcX)$ be translation equivariant prediction maps with receptive field $R > 0$. Assume that, for all $\mathcal{D} \in \mcS$, $\pi_1(\mathcal{D})$ and $\pi_2(\mathcal{D})$ also have receptive field $R > 0$. Let $\epsilon > 0$ and fix $N \in \N$. Assume that, for all $\bfx \in \bigcup_{n=1}^N [0, 2R]^n$ and $\mathcal{D} \in \mcS \cap \bigcup_{n=0}^{\infty} \left([0, 2R] \times \R\right)^n$,
    \begin{equation}
        \KL{P_{\bfx}\pi_1(\mathcal{D})}{P_{\bfx}\pi_2(\mathcal{D}))} \leq \epsilon.
    \end{equation}
    Then, for all $M > 0$, $\bfx \in \bigcup_{n=1}^N [0, M]^n$, and $\mathcal{D} \in \mcS \cap \bigcup_{n=0}^{\infty} \left([0, M] \times \R\right)^n$,
    \begin{equation}
        \KL{P_{\bfx}\pi_1(\mathcal{D})}{P_{\bfx}\pi_2(\mathcal{D}))} \leq \lceil 2M / R\rceil \epsilon.
    \end{equation}
\end{theorem}

\begingroup
\DeclarePairedDelimiter\floor{\lfloor}{\rfloor}  
\DeclarePairedDelimiter\ceil{\lceil}{\rceil}     
\DeclarePairedDelimiter\set{\{}{\}}              
\DeclarePairedDelimiter\abs{|}{|}                
\newcommand{\e}{\epsilon}
\newcommand{\cond}{\mid}
\newcommand{\union}{\bigcup}
\newcommand{\vx}{\bfx}
\newcommand{\vy}{\bfy}
\newcommand{\E}{\mathbb{E}}
\newcommand{\T}{\mathsf{T}}
\renewcommand{\KL}[2]{\operatorname{KL}[#1\|#2]}

\tikzset{
    line/.style = {
        thick,
        ->,
        > = {
            Triangle[length=1.5mm, width=1.5mm]
        }
    }, 
}

\begin{figure*}[t]
    \centering
    \small
    \begin{subfigure}[t]{0.49\linewidth}
        \centering
        \begin{tikzpicture}[
            xscale=0.8,
        ]
            \draw [thick] (0, 0) node [anchor=east] {\strut$x_c$} -- ++(8, 0); 
            \draw [thick] (0, -1) node [anchor=east] {\strut$x_t$} -- ++(8, 0); 
            \node [circle, fill=black, inner sep=0pt, minimum size=3pt] at (2.67, 0) {};
            \path [draw, fill=black!5, thick]
                (2.67, 0) node [circle, fill=black, inner sep=0pt, minimum size=3pt] {}
                -- (1.67, -1)
                -- (3.67, -1)
                -- cycle;
            \draw [thick]
                (1.67, -1.1)
                -- ++(0, -0.1)
                -- ++(2, 0) node [pos=0.5, anchor=north] {$R$}
                -- ++(0, 0.1);
            \path [draw, fill=black!5, thick]
                (5.33, -1) node [circle, fill=black, inner sep=0pt, minimum size=3pt] {}
                -- (4.33, 0)
                -- (6.33, 0)
                -- cycle;
            \draw [thick]
                (4.33, 0.1)
                -- ++(0, 0.1)
                -- ++(2, 0) node [pos=0.5, anchor=south] {$R$}
                -- ++(0, -0.1);
        \end{tikzpicture}
        \caption{
            For a model with receptive field $R > 0$,
            a context point at $x_c$ influences predictions at target inputs only limitedly far away.
            Conversely, a prediction at a target input $x_t$ is influenced by context points only limitedly far away.
        }
        \label{fig:receptive_field}
    \end{subfigure}
    \hfill
    \begin{subfigure}[t]{0.49\linewidth}
        \centering
        \begin{tikzpicture}[xscale=0.8]
            \draw [thick] (0, 0) node [anchor=east] {\strut$x_c$} -- ++(8, 0);
            \draw [thick] (0, -1) node [anchor=east] {\strut$x_t$} -- ++(8, 0);
            \draw [thick] (0.65, -1.05) -- ++(0, 1.1);
            \draw [thick] (3.35, -1.05) node (text) [anchor=north, xshift=-1.15cm] {training range} -- ++(0, 1.1);
            \node [yshift=-7pt] at (text) {\hphantom{$R$}};
            \path [draw, fill=black!5, thick]
                (2, -1) node [circle, fill=black, inner sep=0pt, minimum size=3pt] {}
                -- (1, 0)
                -- (3, 0)
                -- cycle;
            \draw [thick]
                (1, 0.1)
                -- ++(0, 0.1)
                -- ++(2, 0) node [pos=0.5, anchor=south] {$R$}
                -- ++(0, -0.1);
            \path [draw, fill=black!5, thick]
                (5.33, -1) node [circle, fill=black, inner sep=0pt, minimum size=3pt] {}
                -- (4.33, 0)
                -- (6.33, 0)
                -- cycle;
            \draw [thick]
                (4.33, 0.1)
                -- ++(0, 0.1)
                -- ++(2, 0) node [pos=0.5, anchor=south] {$R$}
                -- ++(0, -0.1);
            \draw [line, ->] (4.33, -0.5) -- node [pos=0.4, anchor=south] {TE} (3, -0.5);
        \end{tikzpicture}
        \caption{
            If a model is translation equivariant,
            then all context points and targets inputs can simultaneously be shifted left or right
            without changing the output of the model.
            Intuitively, this means that triangles in the figures can just be ``shifted left or right''. 
        }
    \end{subfigure}
    \caption[
       Translation equivariance can help generalisation
    ]{
        Translation equivariance in combination with a limited receptive field (see (a)) can help generalisation performance.
        Consider a translation equivariant (TE) model which performs well within a \emph{training range} (see (b)).
        Consider a prediction for a target input outside the training range (right triangle in (b)).
        If the model has receptive field $R > 0$ and the training range is bigger than $R$, then TE can be used to ``shift that prediction back into the training range'' (see (b)).
        Since the model performs well within the training range, the model also performs well for the target input outside the training range.
    }
    \label{fig:generalisation}
\end{figure*}


See \Cref{sec:spatial-generalisation} for the proof.
The notion of receptive field is natural to CNNs and corresponds to the usual notion of the receptive field.
The notion is also inherent to transformers that adopt sliding window attention: the size of the window multiplied by the number of transformer layers gives the receptive field of the model.
Intuitively, this theorem states that, if (a) the ground-truth stochastic process and our model are translation equivariant and (b) everything has receptive field size $R >0$, then, whenever our model is accurate on $[0, 2R]$, it is also accurate on any bigger interval $[0, M]$.
Note that this theorem accounts for dependencies between target points, so it also applies to latent-variable neural processes \citep{garnelo2018neural} and Gaussian neural processes \citep{Bruinsma:2021:The_Gaussian_Neural_Process,Markou:2022:Practical_Conditional_Neural_Processes_for_Tractable}.
In practice, we do not explicitly equip our transformer neural processes with a fixed receptive field size by using sliding window attention, although this would certainly be possible.
The main purpose of this theorem is to elucidate the underlying principles that enable translation equivariant neural processes to generalise.

\section{Translation Equivariant Transformer Neural Processes}
\label{sec:tetnp}

Let $\bfZ^{\ell} \in \R^{N\times D_z}$ and $\tilde{\bfZ}^{\ell} \in \R^{N\times D_z}$ denote the inputs and outputs at layer $\ell$, respectively. To achieve translation equivariance, we must ensure that each token $\bfz^{\ell}_n$ is translation invariant with respect to the inputs, which can be achieved through dependency solely on the pairwise distances $\bfx_i - \bfx_j$ (see \Cref{app:te-pe-functions}). We choose to let our initial token embeddings $\bfz^0_n$ depend solely on the corresponding output $\bfy_n$, and introduce dependency on the pairwise distances through the attention mechanism.
For permutation and translation equivariant operations, we choose updates of the form $\tilde{\bfz}^{\ell}_n = \phi\left(\oplus_{m=1}^N \psi\left(\bfz^{\ell}_n, \bfz^{\ell}_m, \bfx_n - \bfx_m\right)\right)$,
where $\oplus$ is a permutation invariant operation. Adopting the MHSA approach, we instantiate this as
\begin{equation}\textstyle
\label{eq:te-mhsa}
    \tilde{\bfz}^{\ell}_n = \operatorname{cat}\!\Big(\Big\{\sum_{m=1}^N \alpha^{\ell}_{h, n, m} {\bfz^{\ell}_m}^T\bfW^{\ell}_{V, h}\Big\}_{h=1}^{H^{\ell}}\Big) \bfW^{\ell}_O
\end{equation}
where $\alpha^{\ell}_{h, n, m} = \alpha^{\ell}_{h}(\bfz_n^{\ell}, \bfz_m^{\ell}, \bfx_n - \bfx_m)$, the attention mechanism, depends on the difference $\bfx_n - \bfx_m$ as well as the input tokens $\bfz^{\ell}_n$ and $\bfz^{\ell}_m$. This differs to the standard attention mechanism in \Cref{eq:self-attention}, which depends solely on the input tokens.
There exist a number of possible choices for the attention mechanism. Again, following the MHSA approach, we implement this as
\begin{equation}
\label{eq:te-attention-mechanism}
    \alpha^{\ell}_{h, n, m} = \frac{e^{\rho^{\ell}_h\left({\bfz^{\ell}_n}^T \bfW^{\ell}_{Q, h}\left[\bfW^{\ell}_{K, h}\right]^T\bfz^{\ell}_m, \bfx_n - \bfx_m\right)}}{\sum_{n=1}^N e^{\rho^{\ell}_h\left({\bfz^{\ell}_n}^T \bfW^{\ell}_{Q, h}\left[\bfW^{\ell}_{K, h}\right]^T\bfz^{\ell}_m, \bfx_n - \bfx_m\right)}}
\end{equation}
where $\rho^{\ell}_h\colon \R \times \R^{D_x} \rightarrow \R$ is a learnable function, which we parameterise by an MLP.\footnote{We implement $\left\{\rho^{\ell}_h\right\}_{h=1}^H$ as a single MLP with $H$ output dimensions.} 
We can also choose to update the input $\bfx_n$ used in \Cref{eq:te-attention-mechanism} with a function $\bff^{\ell}_n(\{\bfx_m\}_{m=1}^N)$, which itself needs to satisfy both translation equivariance and permutation invariance with respect to the set of inputs. A general form for functions 
satisfying translation equivariance and permutation invariance is
\begin{equation} \textstyle
    \label{eq:te-f}
    \bff^{\ell}_n(\{\bfx_m\}_{m=1}^N)\!=\! \sum_{i=1}^N b_{ni} \left(\bfx_i \!+\! g_n\!\left(\sum_{j=1}^N h_n\!\left(\bfx_i \!-\! \bfx_j\right)\right) \right)
\end{equation}
where $b_{n1},\ldots, b_{nN} \in \R$ are \emph{any} set of weights, possibly negative or even dependent on $\bfx_1, \ldots, \bfx_N$, that satisfy $\sum_{i=0}^M b_{ni} = 1$. See \Cref{app:te-pe-functions} for proof. A convenient choice is similar to that used by \citet{satorras2021n}:
\begin{equation} \textstyle
\label{eq:te-x-update}
    \bfx^{\ell + 1}_n = \bfx^{\ell}_n + C\sum_{h=1}^{H^{\ell}} \sum_{m=1}^N (\bfx^{\ell}_n - \bfx^{\ell}_m) \phi^{\ell}_h\left(\alpha^{\ell}_{h, n, m}\right)
\end{equation}
where we have reused the computations of the attention-mechanism. Again, $\phi^{\ell}_h$ is a learnable function, typically parameterised by an MLP, and $C$ is a constant which we choose to be $1 / N$. For the \gls{mhca} layers, we update the target inputs using the context points only.\footnote{i.e.,\ $\bfx^{\ell + 1}_{t, n} = \bfx^{\ell}_{t, n} + C\sum_{h} \sum_{m} (\bfx^{\ell}_{t, n} - \bfx^{\ell}_{c, n})\phi^{\ell}_h(\alpha^{\ell}_{h, n, m})$.}


The attention mechanism defined in \Cref{eq:te-attention-mechanism} in combination with \Cref{eq:te-mhsa} defines the \gls{te-mhsa} operation, which together with layer normalisation and pointwise MLPs described in \Cref{subsec:transformers} forms a \gls{te-mhsa} block. We can define a \gls{te-mhca} operation in an identical manner to the MHCA operation in place of masking when conditional independencies are required. The \gls{te-tnp} shares an identical architecture to the regular \gls{tnp}, with MHSA and MHCA blocks replaced by \gls{te-mhsa} and \gls{te-mhca} blocks, which include both the translation equivariant operation and input location update steps. We illustrate the \gls{te-tnp} and \gls{tnp} encoders in \Cref{fig:encoder-comparison}.

It is worthwhile highlighting the connections between the translation equivariant attention mechanism and the use of relative positional encodings (e.g.\ RoPE \citep{su2024roformer}) in transformer-based large language models. Indeed, relative positional encodings are a specific implementation of our translation equivariant attention mechanism applied to a discrete, regular input domain that words exist on. Our work builds upon these methods, enabling relative positional encodings to be applied to more general, continuous input domains. In both cases, they serve as a useful inductive bias that can improve performance.


\subsection{Translation Equivariance with Pseudo Tokens}
\label{subsec:telbanp}
As with \glspl{pt-tnp}, we can introduce pseudo-tokens $\bfU \in \R^{M \times D_z}$ into the \gls{te-tnp} to reduce the computational complexity from $\mcO(N_c^2 + N_cN_T)$ to $\mcO(MN_c + MN_t)$ (assuming $M \ll N_c, N_t$). To do so, we must also introduce corresponding pseudo-locations $\bfV \in \R^{M\times D_z}$. At each layer, we perform cross-attention between the pseudo-tokens and context-set tokens as in \Cref{eq:te-mhsa}. For the overall architecture to be translation equivariant, we require the initial pseudo-locations to be translation equivariant with respect to the inputs. As before, this is satisfied by functions of the form shown in \Cref{eq:te-f}. We choose to implement this as 
\begin{equation}\textstyle
\label{eq:pseudo-locations}
    \bfv_m = \bfv^0_m + \sum_{n=1}^{N} \psi\left(\bfu_m, \bfz_n\right) \bfx_n
\end{equation}
where $\sum_{n=1}^{N} \psi\left(\bfu_m, \bfz_n\right) = 1\ \forall m$. We exclude dependency on the pairwise differences between the observed input locations as this introduces an $\order{N^2}$ complexity. \Cref{eq:pseudo-locations} can be thought of as constructing a weighted average of the inputs, where the weights depend on the initial token values, which in turn depend on observations. 
We implement $\psi$ using the attention mechanism in \Cref{eq:attention-mechanism}. It should be noted that there exists scenarios in which this initialisation results in undesirable behaviour. For example, suppose that $\bfv^0_m = \*0$, $\phi = 1 / N$. If you observe 100 data points in $[-101, -100]$ and 100 data points in $[100, 101]$, then the pseudo-points will be near the origin, far from the data. In general, we suspect that for any $\psi$, there exists a pathological data set that breaks this initialisation. 
Nonetheless, we found this initialisation scheme to be effective in practice and did not observe pathological behaviour such as this. 

In \Cref{subapp:dynamic-input-updates}, we perform an ablation study to determine the effectiveness of dynamically updating input-locations using \Cref{eq:pseudo-locations} and \Cref{eq:te-x-update}. We find that for simple input distributions (e.g.\ uniformly distributed), dynamically updating input locations has little effect. However, for more complex input distributions (e.g.\ bimodal), dynamically updating input locations significantly improves performance.

\section{Experiments}
\label{sec:experiments}
In this section, we evaluate the performance of both translation equivariant and non-translation-equivariant \gls{np} models on modelling both synthetic and real-world data. We provide more detailed descriptions of the architectures and datasets used in \Cref{app:experimental-details}. In preliminary experiments, we found that \glspl{pt-tnp} using the IST-style architecture outperformed \glspl{pt-tnp} using the perceiver-style architecture. We therefore only include results using the former. 

\begin{figure*}[t]
    \centering
    \begin{subfigure}[t]{\textwidth}
        \centering\includegraphics[width=0.5\textwidth,trim={5 430 10 10},clip]{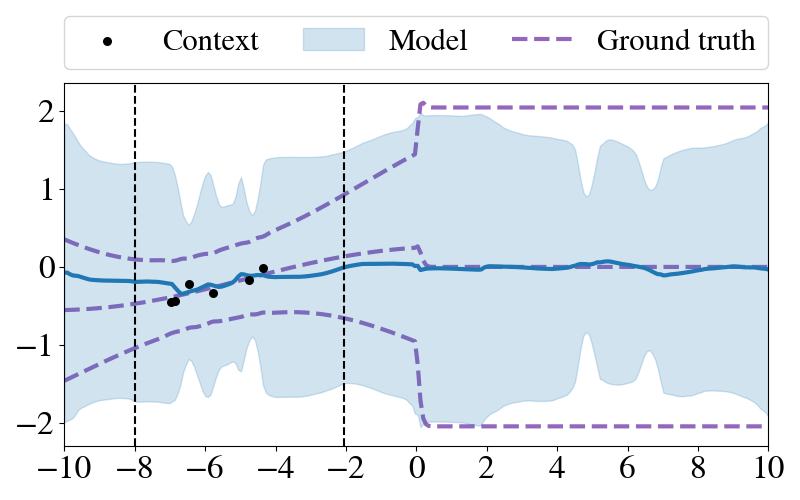}
    \end{subfigure}
    \begin{subfigure}[t]{0.24\textwidth}
        \includegraphics[width=\textwidth,trim={10 10 10 10},clip]{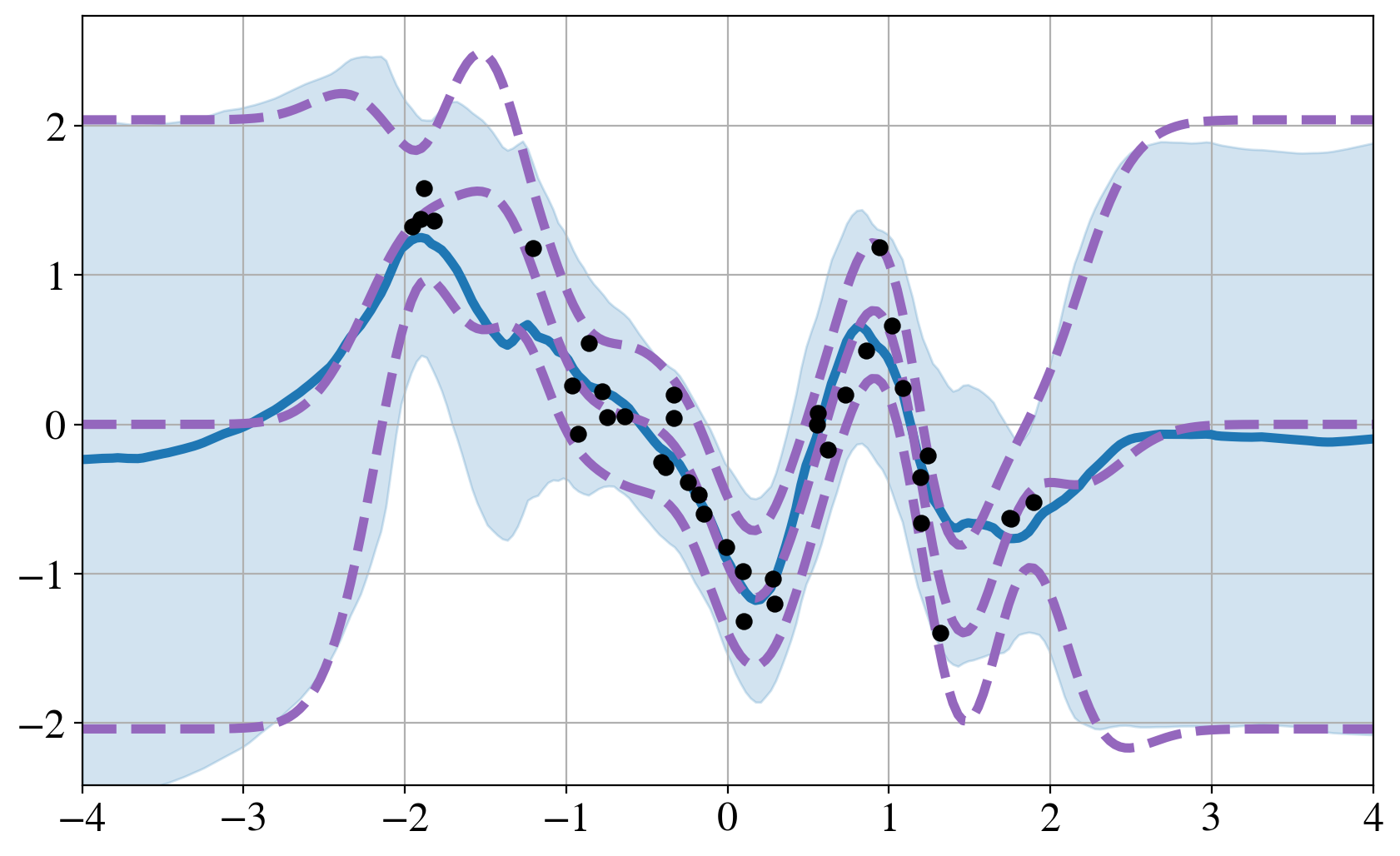}
    \end{subfigure}
    \begin{subfigure}[t]{0.24\textwidth}
        \includegraphics[width=\textwidth,trim={10 10 10 10},clip]{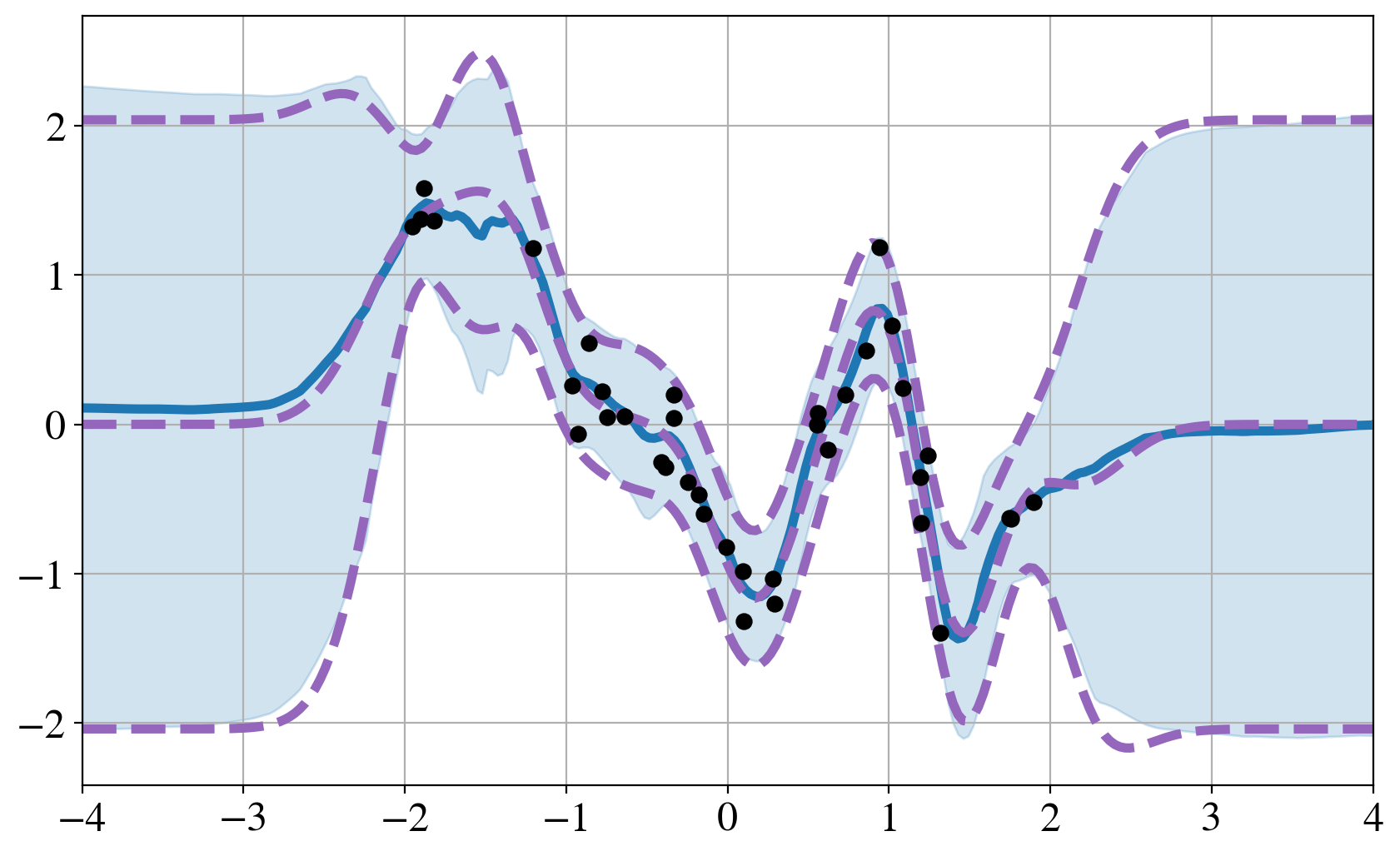}
    \end{subfigure}
    \begin{subfigure}[t]{0.24\textwidth}
        \includegraphics[width=\textwidth,trim={10 10 10 10},clip]{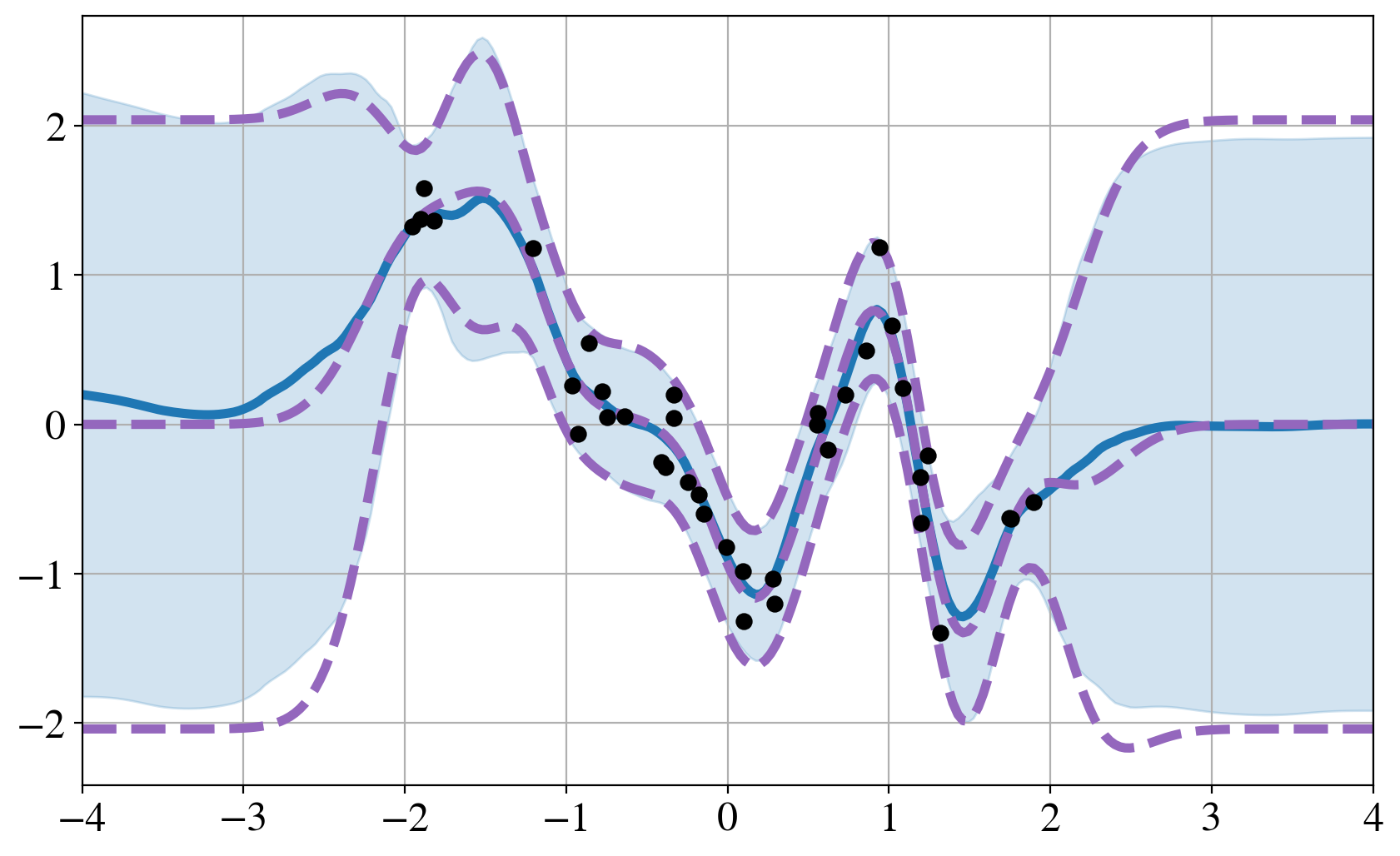}
    \end{subfigure}
    \begin{subfigure}[t]{0.24\textwidth}
        \includegraphics[width=\textwidth,trim={10 10 10 10},clip]{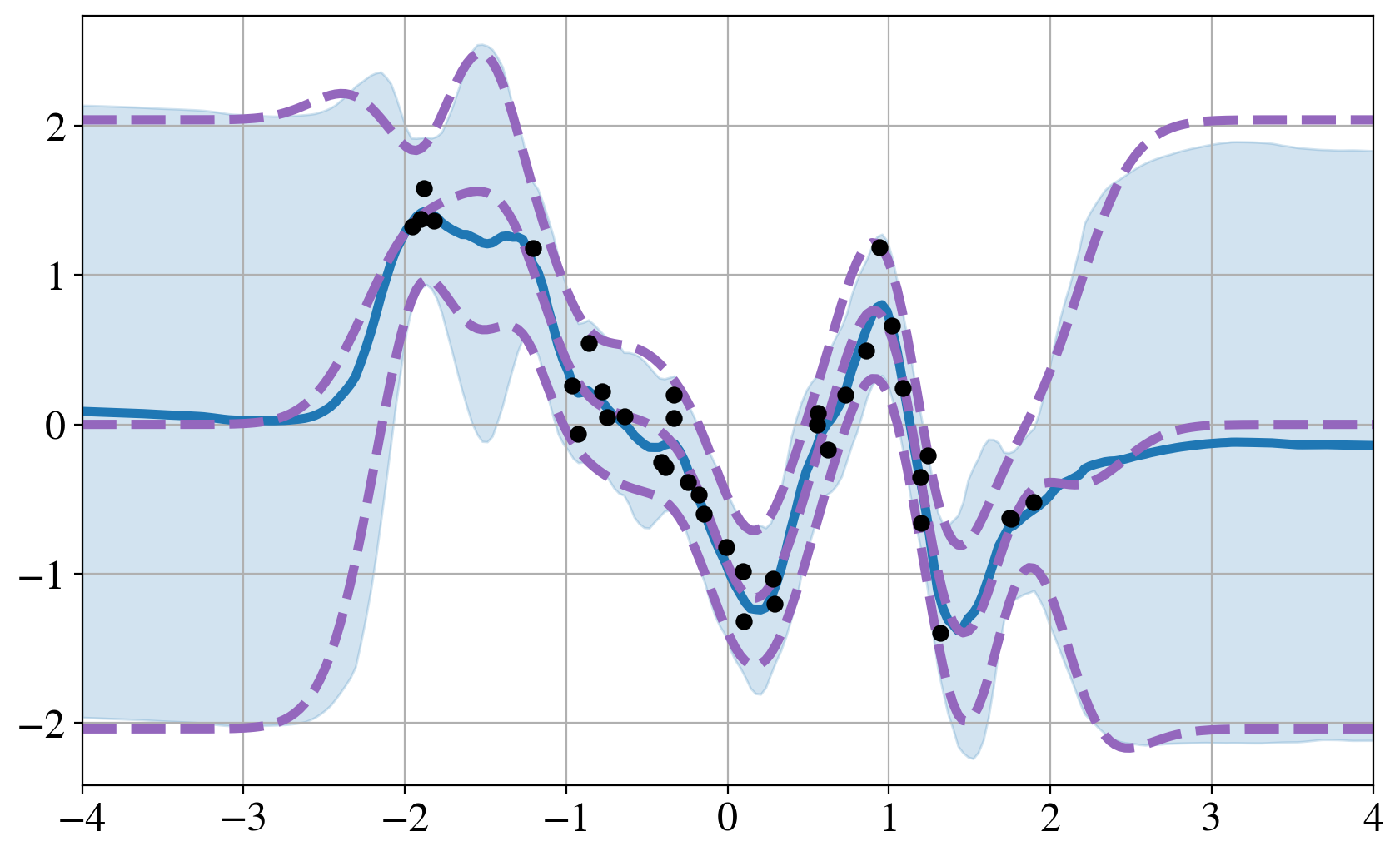}
    \end{subfigure}
    \begin{subfigure}[t]{0.24\textwidth}
        \includegraphics[width=\textwidth,trim={10 10 10 10},clip]{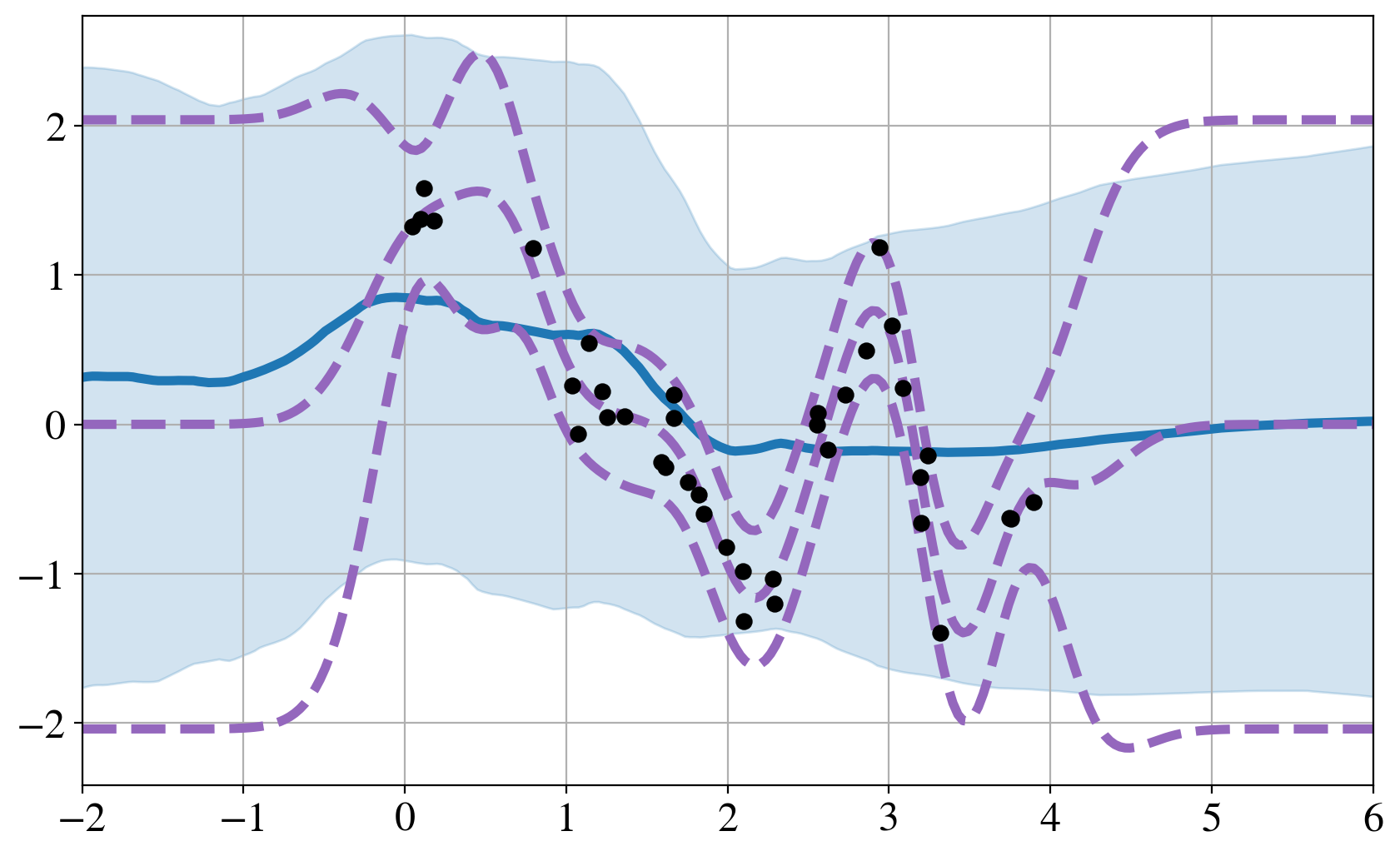}
        \caption{\gls{cnp}.}
    \end{subfigure}
    \begin{subfigure}[t]{0.24\textwidth}
        \includegraphics[width=\textwidth,trim={10 10 10 10},clip]{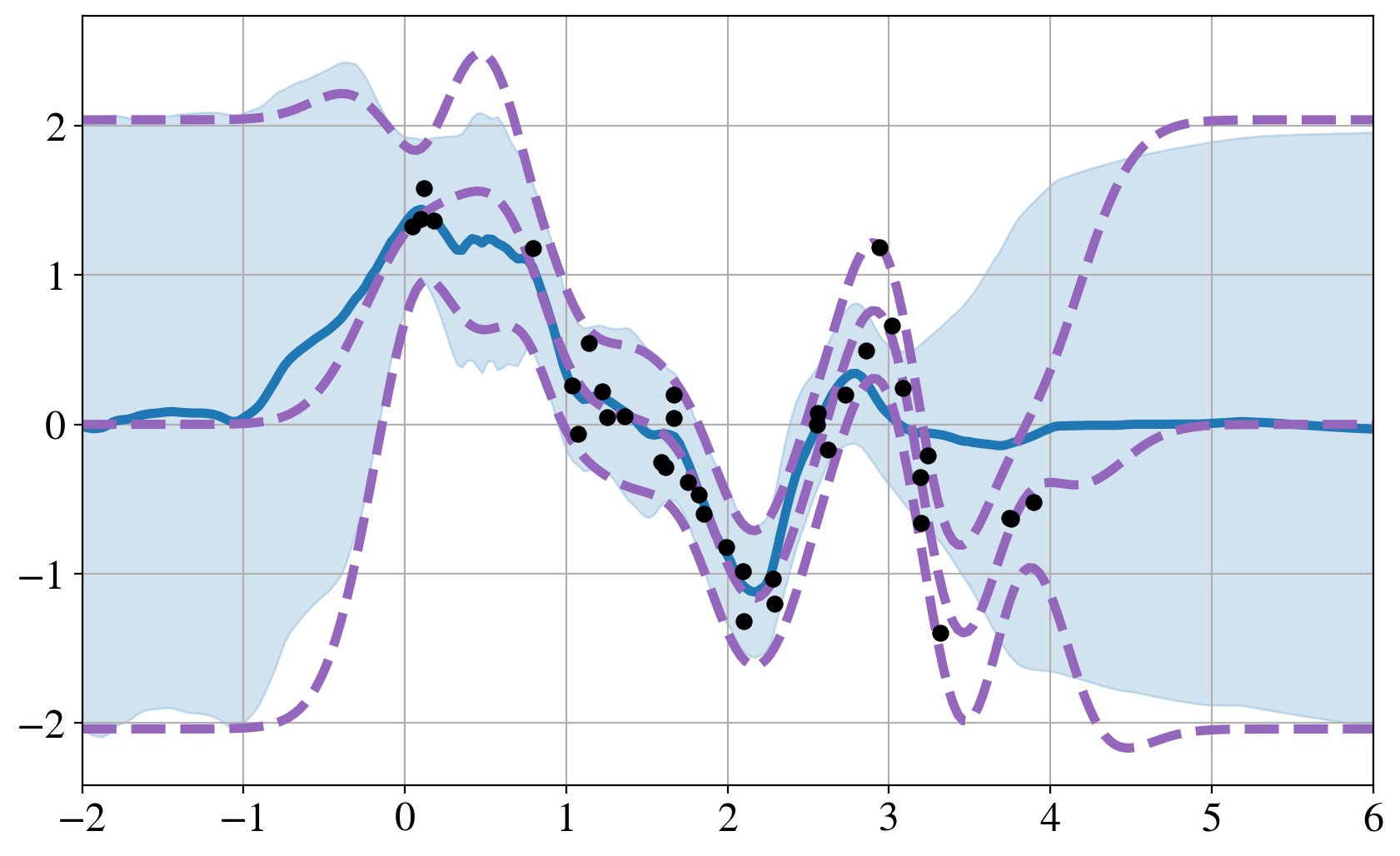}
        \caption{\gls{tnp}.}
    \end{subfigure}
    \begin{subfigure}[t]{0.24\textwidth}
        \includegraphics[width=\textwidth,trim={10 10 10 10},clip]{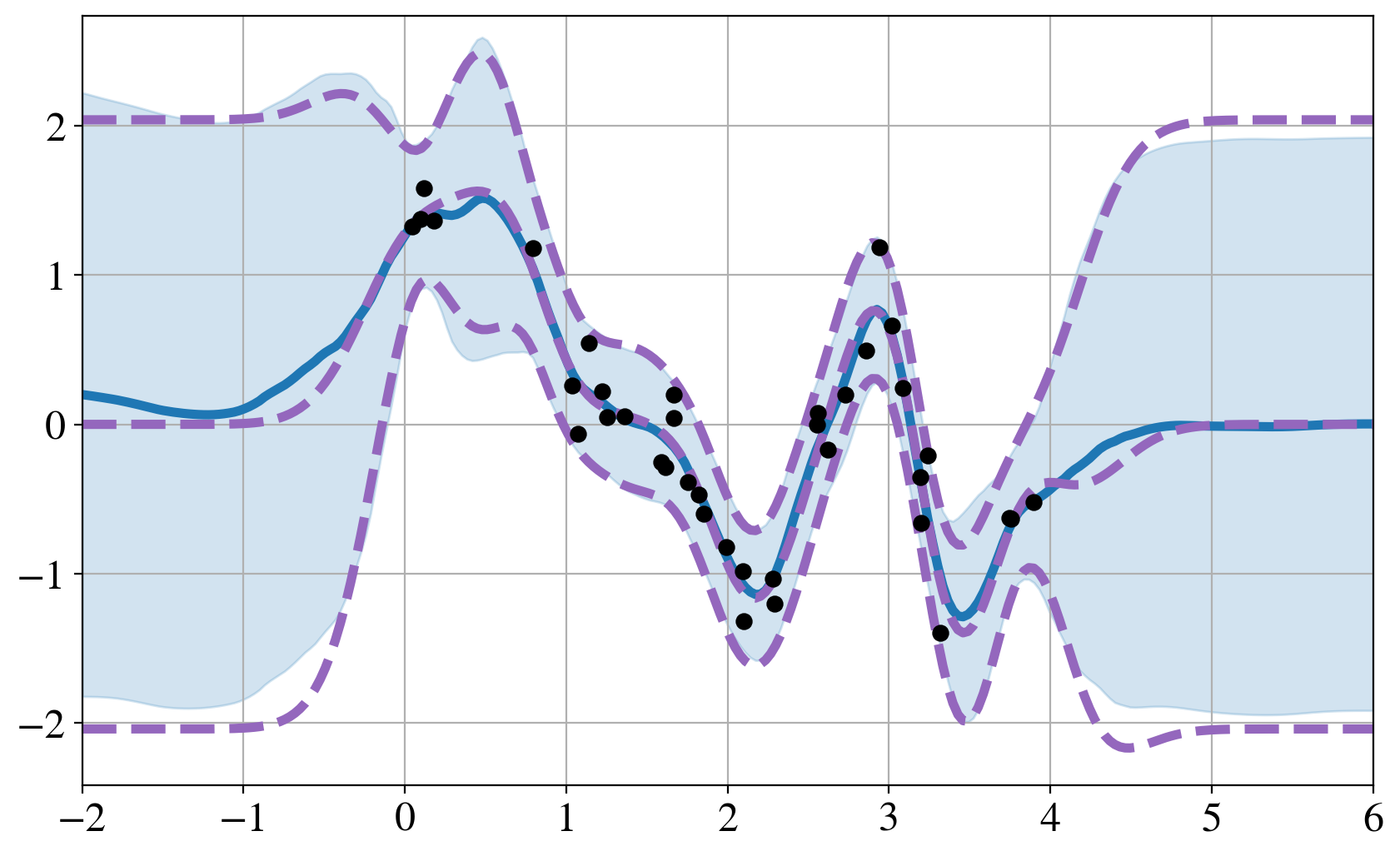}
        \caption{\gls{convcnp}.}
    \end{subfigure}
    \begin{subfigure}[t]{0.24\textwidth}
        \includegraphics[width=\textwidth,trim={10 10 10 10},clip]{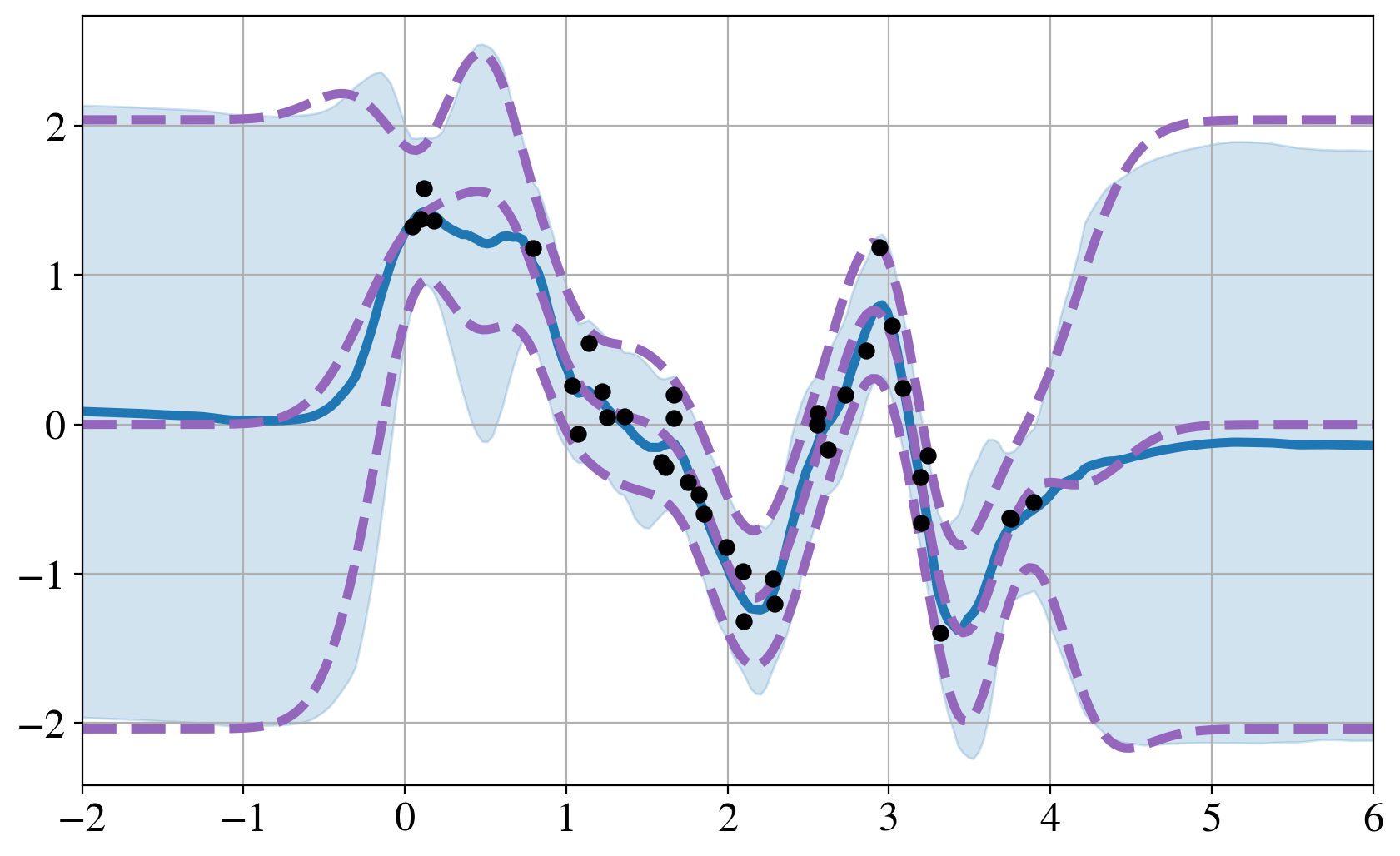}
        \caption{\gls{te-tnp}.}
    \end{subfigure}
    \caption{A comparison between the predictive distributions on a single synthetic-1D regression dataset of the \gls{cnp}, \gls{tnp}, \gls{te-tnp} and \gls{convcnp} when the data is shifted by amount $\Delta = 0$ (top) and $\Delta = 2$ (bottom). Observe that the \gls{te-tnp} and \gls{convcnp} models exhibit translation equivariance, whereas the \gls{cnp} and \gls{tnp} models do not. Context points are shown in black, and the ground-truth predictive mean and $\pm$ standard deviation are shown in dashed-purple.}
    \label{fig:synthetic-1d}
    \vspace{-1em}
\end{figure*}

\subsection{Synthetic 1-D Regression}

We construct a meta-dataset by sampling from Gaussian processes (GPs) with periodic, squared-exponential and Matern 5/2 kernels, all of which define stationary GPs, with randomly sampled kernel hyperparameters. The number of context and target points are sampled according to $N_c \sim U(1, 64)$ and $N_t = 128$, and the context and target inputs are sampled according to $x_c \sim U(-2, 2)$ and $x_t \sim U(-3, 3)$.
The range from which the context and target inputs are sampled from in the test set is shifted by amount $\Delta$. As $\Delta$ increases, so too does the importance of translation equivariance.
We compare the performance of \gls{te-tnp} and \gls{te-pt-tnp} with their non-translation-equivariant counterparts, a \gls{convcnp}, and a simple \gls{rcnp}. See \Cref{subapp:1d-regression} for a detailed description.


\Cref{fig:1d-regression-results} plots the mean test log-likelihood across the test datasets as $\Delta$ increases from 0 to 1. For $\Delta=0$, the input ranges for the test set and target set are equal and translation equivariance is not required to generalise. Nonetheless, we observe that both the \gls{te-tnp} and \gls{te-pt-tnp}-M32 outperform the \gls{tnp} and \gls{pt-tnp}-M32 models, recovering the performance of the \gls{convcnp} model. As $\Delta$ increases, the performance of the non-translation-equivariant models deteriorate as they are unable to generalise to inputs sampled outside the training range. In \Cref{fig:synthetic-1d}, we provide an example of this deterioration for the \gls{cnp} and \gls{tnp}, comparing their predictive posterior distributions to that of the \gls{convcnp} and \gls{te-tnp} for a single test dataset.

\subsection{Image Completion}
\label{subsec:image-completion}
We evaluate the \gls{te-pt-tnp} on image completion tasks, which can be interpreted as spatial regression of pixel values $\bfy_n \in \R^3$ given a 2-D pixel location $\bfx_n \in \R^2$. We consider experiments on MNIST \citep{lecun1998gradient} and CIFAR10. For both datasets, we randomly translate the images by a maximum of $W / 2$ horizontally and $H / 2$ vertically, where $W$ and $H$ denote the width and height of the original image. The number of context points are sampled according to $N_c \sim U\left(\frac{N}{100}, \frac{N}{2}\right)$ and the number of target points are $N_t = N$, where $N$ denotes the total number of pixels in the image. Due to the large number of context points, we do not evaluate the performance of the \gls{te-tnp}. See \Cref{subapp:image-completion} for full experimental details. 
We provide results for the average test log-likelihood for each model in \Cref{tab:image-completion}. Unlike the previous task, here it is possible to learn the required translation equivariance from data. The results of this task demonstrate the utility of equipping our models with translation equivariance when appropriate, even when it is not explicitly required to perform well at test time. A visual comparison between the predictive mean of the \gls{convcnp} and \gls{te-pt-tnp} is shown in \Cref{fig:mnist-translated}. Despite this task being well suited for the \gls{convcnp}, as the pixels fall on a grid, the \glspl{te-pt-tnp} perform competitively.

\begin{table}
    \centering
    \small
    \caption{Average log-likelihood (\textcolor{OliveGreen}{$\*\uparrow$}) on the test datasets for the image completion experiments. Standard errors are shown.}
    \vspace{2pt}
    \begin{tabular}{r c c}
    \toprule
         Model & T-MNIST & T-CIFAR10 \\
         \midrule
         \gls{cnp} & $0.64 \pm 0.00$ & $0.72 \pm 0.00$ \\
         \gls{rcnp} & $1.04 \pm 0.02$ & $1.19 \pm 0.02$ \\
         \gls{convcnp} & $\mathbf{1.20 \pm 0.02}$ & $1.44 \pm 0.02$\\
         \gls{pt-tnp}-M64 & $1.14 \pm 0.02$ & $1.43 \pm 0.01$ \\
         \gls{pt-tnp}-M128 & $1.14 \pm 0.02$ & $1.46 \pm 0.02$ \\
         \midrule
         \gls{te-pt-tnp}-M64 & $1.18 \pm 0.03$ & $\mathbf{1.50 \pm 0.02}$ \\
         \gls{te-pt-tnp}-M128 & $\mathbf{1.25 \pm 0.03}$ & $\mathbf{1.51 \pm 0.02}$\\
         \bottomrule
    \end{tabular}
    \label{tab:image-completion}
\end{table}



\subsection{Kolmogorov Flow}
\label{subsec:kolmogorov}
\begin{figure*}[t]
\begin{minipage}{0.35\linewidth}
    \centering
    \vspace{-10pt}
    \captionof{table}{Average test log-likelihood (\textcolor{OliveGreen}{$\*\uparrow$}) for Kolmogorov flow. Standard errors are shown.}
    \vspace{2pt}
    \small
    \begin{tabular}{r c}
    \toprule
        Model & LL \\
        \midrule
        Multi-task GP & $0.48 \pm 0.02$ \\
        CNP & $-0.97 \pm 0.01$ \\
        RCNP & $0.16 \pm 0.02$ \\
        \gls{pt-tnp}-M64 & $0.63 \pm 0.003$ \\
        \gls{pt-tnp}-M128 & $0.74 \pm 0.03$ \\
        \midrule
        \gls{te-pt-tnp}-M64 & $\mathbf{0.92 \pm 0.03}$ \\
        \gls{te-pt-tnp}-M128 & $\mathbf{0.90 \pm 0.03}$ \\
    \bottomrule
    \end{tabular}
    \label{tab:kolmogorov}
\end{minipage}
\hfill
\begin{minipage}{0.6\linewidth}
    \centering
    \begin{subfigure}{0.24\linewidth}
        \includegraphics[width=\linewidth]{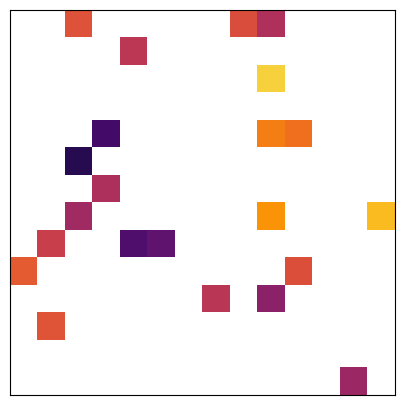}
        \caption{Context.}
    \end{subfigure}
    \begin{subfigure}{0.24\linewidth}
        \includegraphics[width=\linewidth]{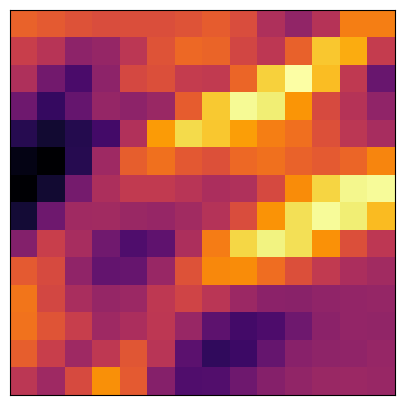}
        \caption{Ground truth.}
    \end{subfigure}
    \begin{subfigure}{0.24\linewidth}
        \includegraphics[width=\linewidth]{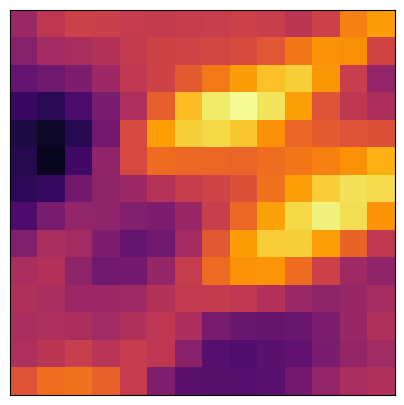}
        \caption{\gls{te-pt-tnp}.}
    \end{subfigure}
    \begin{subfigure}{0.24\linewidth}
        \includegraphics[width=\linewidth]{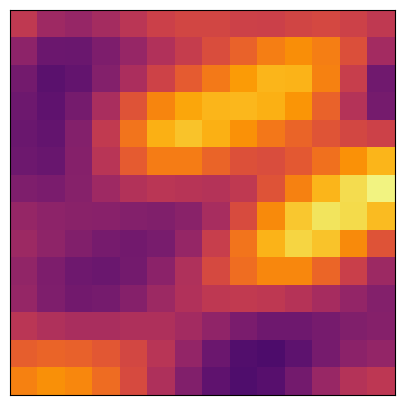}
        \caption{Multi-task GP.}
    \end{subfigure}
    \caption{A comparison between the vorticity at a single point in time, computed using the predicted velocities for a single test Kolmogorov flow dataset. Here, the proportion of datapoints in the context set is 10\%.}
    \label{fig:kolmogorov}
\end{minipage}
\end{figure*}
We consider a dataset generated by the 2-D Kolmogorov flow PDE used in \citep{lippe2023pde,kochkov2021machine,sun2023neural} (see \Cref{subapp:kolmogorov-flow}). 
%
%
The overall dataset consists of 921 simulations---we keep 819 for training and 102 for testing. Each simulation consists of a $64 \times 64 \times 64$ grid of 2-D observations. We sampled individual tasks by first sampling a $16 \times 16 \times 16$ region of a simulation. We then sample the number of context points $N_c \sim U(1, 500)$, with $N_t$ set to all remaining points. As the inputs are 3-D, it is difficult to evaluate the performance of the \gls{convcnp} due to the computational inefficiency of 3-D convolutions. Similarly, the large number of context points restricts our attention to \glspl{pt-tnp}. We compare the performance of the \gls{pt-tnp} and the \gls{te-pt-tnp} to the \gls{rcnp}, \gls{cnp} and a multi-task GP baseline with an SE kernel.\footnote{The multi-task GP baseline is susceptible to overfitting when $N_c$ is small. We therefore remove extreme values from the reported test log-likelihoods.} 
\Cref{tab:kolmogorov} compares the average test log-likelihoods. The \gls{te-pt-tnp} significantly outperforms all other models. In \Cref{fig:kolmogorov}, we visualise the vorticities computed using the predicted velocities of the \gls{te-pt-tnp} and multi-task GP models.

\subsection{Environmental Data}
\label{subsec:environmental-data}
We consider a real-world environmental dataset, derived from ERA5 \citep{cccs2020}, consisting of surface air temperatures across space and time. Measurements are collected at a latitudinal and longitudinal resolution of $0.5^{\circ}$, and temporal resolution of an hour ($\bfx_n \in \R^3$). We also provide the surface elevation at each coordinate as inputs. We consider measurements collected in 2019. Models are trained on measurements within the latitude / longitude range of $[42^{\circ},\ 53^{\circ}]$ / $[8^{\circ},\ 28^{\circ}]$ (roughly corresponding to central Europe), and evaluated on three non-overlapping regions: the training region, western Europe ($[42^{\circ},\ 53^{\circ}]$ / $[-4^{\circ},\ 8^{\circ}]$), and northern Europe ($[53^{\circ},\ 62^{\circ}]$ / $[8^{\circ},\ 28^{\circ}]$). 
During training, we sample datasets spanning 30 hours, sub-sampled to one every six hours, and $7.5^{\circ}$ across each axis. 
Each dataset consists of a maximum of $N = 1125$ datapoints, from which the number of context points are sampled according to $N_c \sim U(\frac{N}{100}, \frac{N}{3})$. As with the image completion experiments, the large number of datapoints present in each dataset render the \gls{tnp} and \gls{te-tnp} too computationally intensive to train and evaluate. Similarly, because the inputs are 4-D, the \gls{convcnp} cannot be implement due to insufficient support for 4-D convolutions. We therefore restrict our attention to the \gls{pt-tnp} and \gls{te-pt-tnp}, both with $M=128$ pseudo-tokens. We also evaluate the predictive performance of the \gls{cnp}, \gls{rcnp} and GPs with SE kernels. See \Cref{subapp:environmental-data} for full experimental details. \Cref{tab:cru-tair} compares the average test log-likelihood for each method on the different regions. The \gls{te-pt-tnp} model outperforms all baselines in all three regions. As noted by \citet{foong2020meta}, this is somewhat unsurprising given that: 
\begin{enumerate*}[label={(\arabic*)}]
    \item the GP is prone to overfitting on small context sizes; and
    \item the degree to which points influence each other is unlikely to be well explained by a SE kernel.
\end{enumerate*}

\begin{table}
\addtolength{\tabcolsep}{-0.2em}
    \centering
    \caption{Average log-likelihood (\textcolor{OliveGreen}{$\*\uparrow$}) for tasks sampled within the train range (C.\ Europe) and test ranges (W.\ Europe and N.\ Europe).}
    \vspace{2pt}
    \small
    \begin{tabular}{r c c c}
    \toprule
         Model & C.\ Europe & W.\ Europe & N.\ Europe \\
         \midrule
         GP & $0.90 \pm 0.01$ & $1.22 \pm 0.01$ & $1.02 \pm 0.01$  \\
        \gls{cnp} & $0.36 \pm 0.00$ & $-5.70 \pm 0.05$ & $-1.64 \pm 0.02$ \\
         \gls{rcnp} & $0.96 \pm 0.01$ & $-0.09 \pm 0.04$ & $0.34 \pm 0.01$  \\
         \gls{pt-tnp} & $0.73 \pm 0.01$ & $-0.40 \pm 0.01$ & $-0.19 \pm 0.00$ \\
         \midrule
         \gls{te-pt-tnp} & $\mathbf{1.35 \pm 0.01}$ & $\mathbf{1.27 \pm 0.01}$ & $\mathbf{1.44 \pm 0.01}$ \\
         \bottomrule
    \end{tabular}
    \label{tab:cru-tair}
\addtolength{\tabcolsep}{0.2em}
\end{table}

\begin{figure*}[ht]
    \centering
    \begin{subfigure}[t]{0.24\textwidth}
        \includegraphics[width=\textwidth]{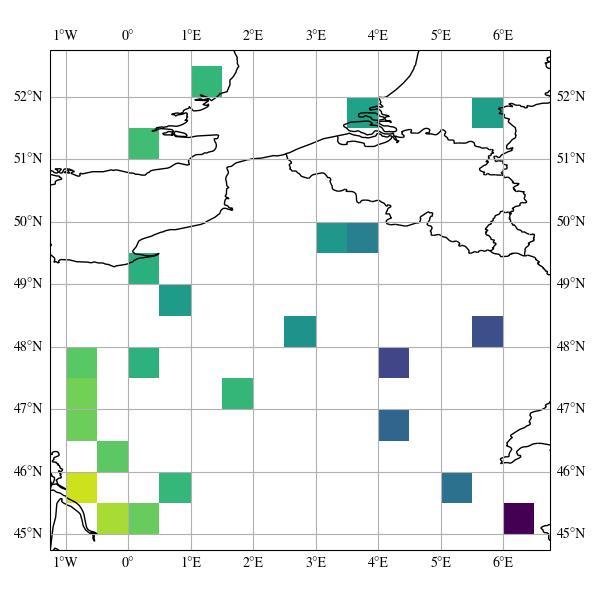}
        \caption{Context dataset.}
        \label{subfig:context}
    \end{subfigure}
    \begin{subfigure}[t]{0.24\textwidth}
        \includegraphics[width=\textwidth]{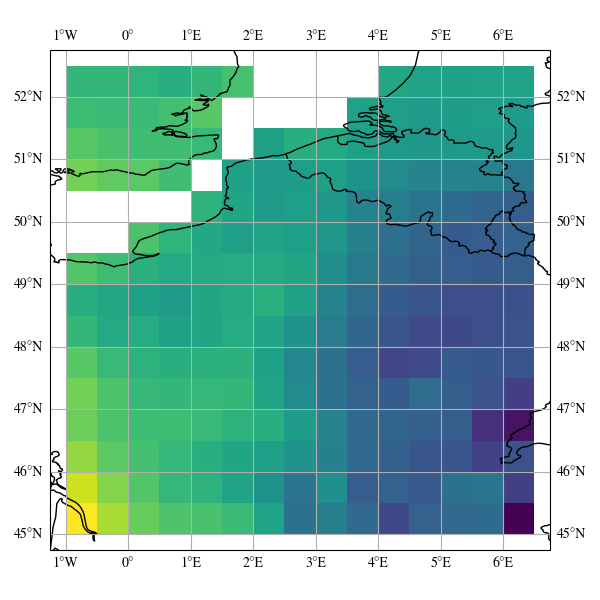}
        \caption{Ground truth data.}
        \label{subfig:ground-truth}
    \end{subfigure}
    \begin{subfigure}[t]{0.24\textwidth}
        \includegraphics[width=\textwidth]{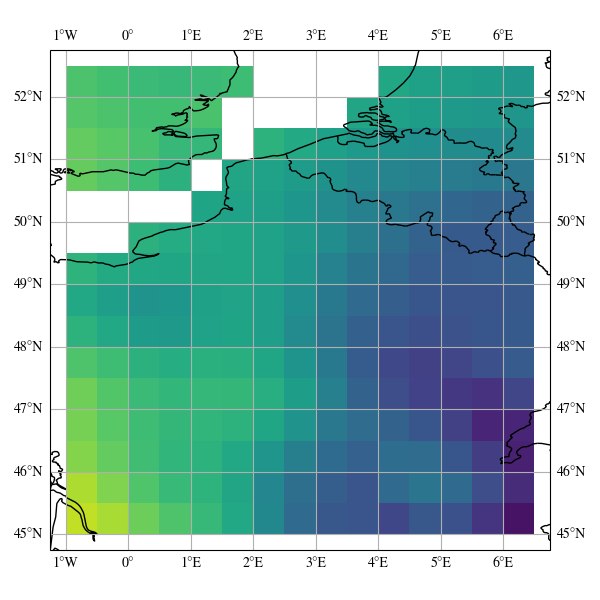}
        \caption{\gls{te-pt-tnp}.}
        \label{subfig:telbanp-m64-pred}
    \end{subfigure}
    \begin{subfigure}[t]{0.24\textwidth}
        \includegraphics[width=\textwidth]{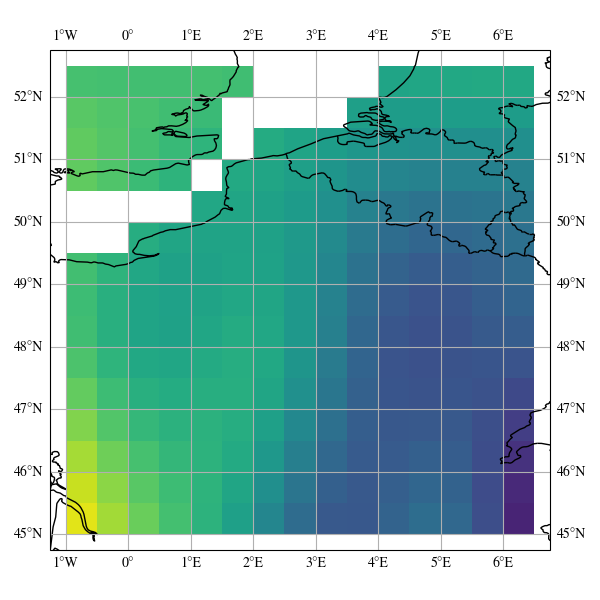}
        \caption{GP.}
        \label{subfig:gp-pred}
    \end{subfigure}
    \caption{A comparison between the predictive means of the \gls{te-pt-tnp} model and a GP for a single test dataset sampled from western Europe. The figures visualise a single slice through time of the (predictive) temperature, with the same colour scale being used throughout.}
    \label{fig:western-europe-2019}
\end{figure*}

\section{Related Work}
\paragraph{Neural processes}
The combination of transformers and \glspl{np} was first considered by \citet{kim2019attentive} who developed the \gls{anp}. The \gls{anp} is characterised by multiple MHSA layers operating on the context tokens, followed by a single multi-head attention (MHA) layer in which the queries are the target locations, the keys are the context locations, and the values are the context tokens. 
%
The \gls{tnp-d} architecture introduced by \citet{nguyen2022transformer} built upon the \gls{anp} with multiple MHSA layers operating on the concatenation of context and target tokens. This repeated transfer of information from the context tokens to the target tokens at each layer improves performance, but is computationally inefficient in comparison to the \gls{eq-tnp} \citep{feng2022latent} which replaces masked MHSA layers with MHSA and MHCA. Finally, the \gls{lbanp} \citep{feng2022latent} introduce the use of pseudo-tokens in a perceiver-style architecture to reduce the computational complexity further. To the best of our knowledge, there do not exist any \gls{tnp} models which incorporate translation equivariance. However, there have been other \gls{np} variants which seek to achieve this. The \gls{rcnp} \citep{huang2023practical} is similar to a single layer of the \gls{te-tnp}. However, as their choice of permutation aggregation function is linear (summation), they are not equivalent. \glspl{convcnp} \citep{gordon2019convolutional}, and more generally steerable CNPs \citep{holderrieth2021equivariant}, incorporate translation equivariance by obtaining context representations in function space, discretising, and then performing translation equivariant operations using a CNN. A key computational advantage of the \gls{convcnp} over the \gls{te-tnp} is the use of a CNN, which has computational complexity linear in the number of input points. However, the necessity of discretisation and convolutions restricts the \gls{convcnp} to low-dimensional input domains. Further, the \gls{convcnp} requires practioners to be much more judicious in their choice of model architecture, as careful consideration of discretisation the implied receptive field are required. We provide a summary of some important difference between the \gls{np} models discussed here in \Cref{tab:np-comparison}.
\begin{table}[htb]
    \centering
    \caption{A comparison between NP models. Complexity refers to computational complexity. FE denotes functional embedding: whether or not $e(\mcD_c, \bfx_t)$ depends on the input location $\bfx_t$. $D_x$-S denotes $D_x$ scalability, where $D_x$ is the input dimension. TE denotes translation equivariance.}
    \vspace{2pt}
    \small
    \begin{tabular}{l l c c c}
        \toprule
        Model & Complexity & $D_x$-S & FE & TE  \\
        \midrule
        \gls{cnp} & $\order{N_c + N_t}$ & \textcolor{OliveGreen}{\cmark} & \textcolor{BrickRed}{\xmark} & \textcolor{BrickRed}{\xmark} \\
        \gls{rcnp} & $\order{N_cN_t}$ & \textcolor{OliveGreen}{\cmark} & \textcolor{OliveGreen}{\cmark} & \textcolor{OliveGreen}{\cmark} \\
        \gls{convcnp} & $\order{N_cD_x^3 + N_tD_x}$ & \textcolor{BrickRed}{\xmark} & \textcolor{BrickRed}{\xmark} & \textcolor{OliveGreen}{\cmark} \\
        \gls{tnp} & $\order{N_c^2 + N_cN_t}$ & \textcolor{OliveGreen}{\cmark} & \textcolor{OliveGreen}{\cmark} & \textcolor{BrickRed}{\xmark} \\
        \gls{pt-tnp} & $\order{MN_c + MN_t}$ & \textcolor{OliveGreen}{\cmark} & \textcolor{OliveGreen}{\cmark} & \textcolor{BrickRed}{\xmark} \\
        \midrule
        \gls{te-tnp} & $\order{N_c^2 + N_cN_t}$ & \textcolor{OliveGreen}{\cmark} & \textcolor{OliveGreen}{\cmark} & \textcolor{OliveGreen}{\cmark} \\
        \gls{te-pt-tnp} & $\order{MN_c + MN_t}$ & \textcolor{OliveGreen}{\cmark} & \textcolor{OliveGreen}{\cmark} & \textcolor{OliveGreen}{\cmark} \\
        \bottomrule
    \end{tabular}
    \label{tab:np-comparison}
\end{table}

\paragraph{Equivariant GNNs and transformers}
When graph neural networks (GNNs) are applied to nodes in the euclidean domain, it is often beneficial to incorporate certain forms of euclidean equivariance. Doing so has been a topic of significant interest \citep{bronstein2021geometric}. 
\citet{satorras2021n} build E(n)-equivariant GNNs in the form of equivariant message passing, and bears close similarity to the approach taken in this paper. However, they tackle a different form of equivariance and their method does not closely resemble the attention-mechanism of transformers. 
\citet{fuchs2020se} build SE(3)-equivariant transformers, which our method shares similarities with. However, we avoid the complexities of irreducible representations by considering only translation equivariance. We choose a simpler, less memory intensive approach for incorporating translation equivariance that is effective in practice.
The LieTransformer \citep{hutchinson2021lietransformer} is a generalisation of translation equivariant transformers to Lie groups. Indeed, for certain choices of content-based and location-based attention mechanisms, our method can be recovered from the LieSelfAttention operation (see \Cref{app:lietransformer}). Nonetheless, \citeauthor{hutchinson2021lietransformer} focus on SE-equivariance in their experiments, and do not consider integration into \glspl{tnp}.

\paragraph{Data augmentation} An alternative strategy to directly incorporating inductive biases, such as translation equivariance, into models is to use data augmentation during training. However, several works have shown that this approach has worse sample complexity and generalisation guarantees~\citep{pmlr-v134-mei21a, wang2022data, holderrieth2021equivariant}. We found this to be empirically true in preliminary experiments, hence exclude comparisons in our experiments.

\section{Conclusion}
\label{sec:discussion}
We have introduced the \gls{te-tnp} and \gls{te-pt-tnp}, expanding the family of \glspl{tnp} to include their translation equivariant equivalents through the development of \gls{te-mhsa} and \gls{te-mhca} operations. 
An extensive range of empirical results demonstrate that the \gls{te-tnp} and \gls{te-pt-tnp} perform on par or better than state-of-the-art \glspl{np}, such as the \gls{convcnp}, whilst being versatile in their applicability.
These models are not without their drawbacks, the two most significant being: 
\begin{enumerate*}[label={(\arabic*)}]
    \item although not affecting the asymptotic behaviour with respect to the number of datapoints, in practice the need to pass pairwise computations through MLPs scales the computational complexity by a large factor; and
    \item the need for the number of pseudo-tokens in \glspl{pt-tnp} to scale with the number of datapoints.
\end{enumerate*}
We seek to address both of these in future work.
%

\section*{Acknowledgements}
CD is supported by the Cambridge Trust Scholarship. SM is supported by the Vice Chancellor's and Marie and George Vergottis Scholarship, and the Qualcomm Innovation Fellowship. JM acknowledges funding from the Data Sciences Institute at the University of Toronto and the Vector Institute. RET is supported by gifts from Google, Amazon, ARM, Improbable and EPSRC grant EP/T005386/1. We thank Emile Mathieu for their valuable feedback.

\section*{Impact Statement}
This paper presents work whose goal is to advance the field of Machine Learning. There are many potential societal consequences of our work, none which we feel must be specifically highlighted here.



\bibliography{bibliography}

\begin{thebibliography}{40}
\providecommand{\natexlab}[1]{#1}
\providecommand{\url}[1]{\texttt{#1}}
\expandafter\ifx\csname urlstyle\endcsname\relax
  \providecommand{\doi}[1]{doi: #1}\else
  \providecommand{\doi}{doi: \begingroup \urlstyle{rm}\Url}\fi

\bibitem[Achiam et~al.(2023)Achiam, Adler, Agarwal, Ahmad, Akkaya, Aleman, Almeida, Altenschmidt, Altman, Anadkat, et~al.]{achiam2023gpt}
Achiam, J., Adler, S., Agarwal, S., Ahmad, L., Akkaya, I., Aleman, F.~L., Almeida, D., Altenschmidt, J., Altman, S., Anadkat, S., et~al.
\newblock Gpt-4 technical report.
\newblock \emph{arXiv preprint arXiv:2303.08774}, 2023.

\bibitem[Betker et~al.(2023)Betker, Goh, Jing, Brooks, Wang, Li, Ouyang, Zhuang, Lee, Guo, et~al.]{betker2023improving}
Betker, J., Goh, G., Jing, L., Brooks, T., Wang, J., Li, L., Ouyang, L., Zhuang, J., Lee, J., Guo, Y., et~al.
\newblock Improving image generation with better captions.
\newblock \emph{Computer Science. https://cdn. openai. com/papers/dall-e-3. pdf}, 2\penalty0 (3), 2023.

\bibitem[Bronstein et~al.(2021)Bronstein, Bruna, Cohen, and Veli{\v{c}}kovi{\'c}]{bronstein2021geometric}
Bronstein, M.~M., Bruna, J., Cohen, T., and Veli{\v{c}}kovi{\'c}, P.
\newblock Geometric deep learning: Grids, groups, graphs, geodesics, and gauges.
\newblock \emph{arXiv preprint arXiv:2104.13478}, 2021.

\bibitem[Bruinsma et~al.(2021)Bruinsma, Requeima, Foong, Gordon, and Turner]{Bruinsma:2021:The_Gaussian_Neural_Process}
Bruinsma, W.~P., Requeima, J., Foong, A. Y.~K., Gordon, J., and Turner, R.~E.
\newblock The {Gaussian} neural process.
\newblock In \emph{Proceedings of the 3rd Symposium on Advances in Approximate {Bayesian} Inference}, 2021.

\bibitem[{Copernicus Climate Change Service}(2020)]{cccs2020}
{Copernicus Climate Change Service}.
\newblock Near surface meteorological variables from 1979 to 2018 derived from bias-corrected reanalysis, 2020.

\bibitem[Dosovitskiy et~al.(2020)Dosovitskiy, Beyer, Kolesnikov, Weissenborn, Zhai, Unterthiner, Dehghani, Minderer, Heigold, Gelly, et~al.]{dosovitskiy2020image}
Dosovitskiy, A., Beyer, L., Kolesnikov, A., Weissenborn, D., Zhai, X., Unterthiner, T., Dehghani, M., Minderer, M., Heigold, G., Gelly, S., et~al.
\newblock An image is worth 16x16 words: Transformers for image recognition at scale.
\newblock \emph{arXiv preprint arXiv:2010.11929}, 2020.

\bibitem[Feng et~al.(2022)Feng, Hajimirsadeghi, Bengio, and Ahmed]{feng2022latent}
Feng, L., Hajimirsadeghi, H., Bengio, Y., and Ahmed, M.~O.
\newblock Latent bottlenecked attentive neural processes.
\newblock \emph{arXiv preprint arXiv:2211.08458}, 2022.

\bibitem[Foong et~al.(2020)Foong, Bruinsma, Gordon, Dubois, Requeima, and Turner]{foong2020meta}
Foong, A., Bruinsma, W., Gordon, J., Dubois, Y., Requeima, J., and Turner, R.
\newblock Meta-learning stationary stochastic process prediction with convolutional neural processes.
\newblock \emph{Advances in Neural Information Processing Systems}, 33:\penalty0 8284--8295, 2020.

\bibitem[Fuchs et~al.(2020)Fuchs, Worrall, Fischer, and Welling]{fuchs2020se}
Fuchs, F., Worrall, D., Fischer, V., and Welling, M.
\newblock Se (3)-transformers: 3d roto-translation equivariant attention networks.
\newblock \emph{Advances in neural information processing systems}, 33:\penalty0 1970--1981, 2020.

\bibitem[Gardner et~al.(2018)Gardner, Pleiss, Weinberger, Bindel, and Wilson]{gardner2018gpytorch}
Gardner, J., Pleiss, G., Weinberger, K.~Q., Bindel, D., and Wilson, A.~G.
\newblock Gpytorch: Blackbox matrix-matrix gaussian process inference with gpu acceleration.
\newblock \emph{Advances in neural information processing systems}, 31, 2018.

\bibitem[Garnelo \& Czarnecki(2023)Garnelo and Czarnecki]{garnelo2023exploring}
Garnelo, M. and Czarnecki, W.~M.
\newblock Exploring the space of key-value-query models with intention.
\newblock \emph{arXiv preprint arXiv:2305.10203}, 2023.

\bibitem[Garnelo et~al.(2018{\natexlab{a}})Garnelo, Rosenbaum, Maddison, Ramalho, Saxton, Shanahan, Teh, Rezende, and Eslami]{garnelo2018conditional}
Garnelo, M., Rosenbaum, D., Maddison, C., Ramalho, T., Saxton, D., Shanahan, M., Teh, Y.~W., Rezende, D., and Eslami, S.~A.
\newblock Conditional neural processes.
\newblock In \emph{International conference on machine learning}, pp.\  1704--1713. PMLR, 2018{\natexlab{a}}.

\bibitem[Garnelo et~al.(2018{\natexlab{b}})Garnelo, Schwarz, Rosenbaum, Viola, Rezende, Eslami, and Teh]{garnelo2018neural}
Garnelo, M., Schwarz, J., Rosenbaum, D., Viola, F., Rezende, D.~J., Eslami, S., and Teh, Y.~W.
\newblock Neural processes.
\newblock \emph{arXiv preprint arXiv:1807.01622}, 2018{\natexlab{b}}.

\bibitem[Gordon et~al.(2019)Gordon, Bruinsma, Foong, Requeima, Dubois, and Turner]{gordon2019convolutional}
Gordon, J., Bruinsma, W.~P., Foong, A.~Y., Requeima, J., Dubois, Y., and Turner, R.~E.
\newblock Convolutional conditional neural processes.
\newblock \emph{arXiv preprint arXiv:1910.13556}, 2019.

\bibitem[Holderrieth et~al.(2021)Holderrieth, Hutchinson, and Teh]{holderrieth2021equivariant}
Holderrieth, P., Hutchinson, M.~J., and Teh, Y.~W.
\newblock Equivariant learning of stochastic fields: Gaussian processes and steerable conditional neural processes.
\newblock In \emph{International Conference on Machine Learning}, pp.\  4297--4307. PMLR, 2021.

\bibitem[Huang et~al.(2023)Huang, Haussmann, Remes, John, Clart{\'e}, Luck, Kaski, and Acerbi]{huang2023practical}
Huang, D., Haussmann, M., Remes, U., John, S., Clart{\'e}, G., Luck, K.~S., Kaski, S., and Acerbi, L.
\newblock Practical equivariances via relational conditional neural processes.
\newblock \emph{arXiv preprint arXiv:2306.10915}, 2023.

\bibitem[Hutchinson et~al.(2021)Hutchinson, Le~Lan, Zaidi, Dupont, Teh, and Kim]{hutchinson2021lietransformer}
Hutchinson, M.~J., Le~Lan, C., Zaidi, S., Dupont, E., Teh, Y.~W., and Kim, H.
\newblock Lietransformer: Equivariant self-attention for lie groups.
\newblock In \emph{International Conference on Machine Learning}, pp.\  4533--4543. PMLR, 2021.

\bibitem[Jaegle et~al.(2021)Jaegle, Gimeno, Brock, Vinyals, Zisserman, and Carreira]{jaegle2021perceiver}
Jaegle, A., Gimeno, F., Brock, A., Vinyals, O., Zisserman, A., and Carreira, J.
\newblock Perceiver: General perception with iterative attention.
\newblock In \emph{International conference on machine learning}, pp.\  4651--4664. PMLR, 2021.

\bibitem[Jha et~al.(2022)Jha, Gong, Wang, Turner, and Yao]{jha2022neural}
Jha, S., Gong, D., Wang, X., Turner, R.~E., and Yao, L.
\newblock The neural process family: Survey, applications and perspectives.
\newblock \emph{arXiv preprint arXiv:2209.00517}, 2022.

\bibitem[Kim et~al.(2019)Kim, Mnih, Schwarz, Garnelo, Eslami, Rosenbaum, Vinyals, and Teh]{kim2019attentive}
Kim, H., Mnih, A., Schwarz, J., Garnelo, M., Eslami, A., Rosenbaum, D., Vinyals, O., and Teh, Y.~W.
\newblock Attentive neural processes.
\newblock \emph{arXiv preprint arXiv:1901.05761}, 2019.

\bibitem[Kingma \& Ba(2014)Kingma and Ba]{kingma2014adam}
Kingma, D.~P. and Ba, J.
\newblock Adam: A method for stochastic optimization.
\newblock \emph{arXiv preprint arXiv:1412.6980}, 2014.

\bibitem[Kochkov et~al.(2021)Kochkov, Smith, Alieva, Wang, Brenner, and Hoyer]{kochkov2021machine}
Kochkov, D., Smith, J.~A., Alieva, A., Wang, Q., Brenner, M.~P., and Hoyer, S.
\newblock Machine learning--accelerated computational fluid dynamics.
\newblock \emph{Proceedings of the National Academy of Sciences}, 118\penalty0 (21):\penalty0 e2101784118, 2021.

\bibitem[LeCun et~al.(1998)LeCun, Bottou, Bengio, and Haffner]{lecun1998gradient}
LeCun, Y., Bottou, L., Bengio, Y., and Haffner, P.
\newblock Gradient-based learning applied to document recognition.
\newblock \emph{Proceedings of the IEEE}, 86\penalty0 (11):\penalty0 2278--2324, 1998.

\bibitem[Lee et~al.(2019)Lee, Lee, Kim, Kosiorek, Choi, and Teh]{lee2019set}
Lee, J., Lee, Y., Kim, J., Kosiorek, A., Choi, S., and Teh, Y.~W.
\newblock Set transformer: A framework for attention-based permutation-invariant neural networks.
\newblock In \emph{International conference on machine learning}, pp.\  3744--3753. PMLR, 2019.

\bibitem[Lippe et~al.(2023)Lippe, Veeling, Perdikaris, Turner, and Brandstetter]{lippe2023pde}
Lippe, P., Veeling, B.~S., Perdikaris, P., Turner, R.~E., and Brandstetter, J.
\newblock Pde-refiner: Achieving accurate long rollouts with neural pde solvers.
\newblock \emph{arXiv preprint arXiv:2308.05732}, 2023.

\bibitem[Loshchilov \& Hutter(2017)Loshchilov and Hutter]{loshchilov2017decoupled}
Loshchilov, I. and Hutter, F.
\newblock Decoupled weight decay regularization.
\newblock \emph{arXiv preprint arXiv:1711.05101}, 2017.

\bibitem[Markou et~al.(2022)Markou, Requeima, Bruinsma, Vaughan, and Turner]{Markou:2022:Practical_Conditional_Neural_Processes_for_Tractable}
Markou, S., Requeima, J., Bruinsma, W.~P., Vaughan, A., and Turner, R.~E.
\newblock Practical conditional neural processes via tractable dependent predictions.
\newblock In \emph{Proceedings of the 10th International Conference on Learning Representations}, 2022.

\bibitem[Mei et~al.(2021)Mei, Misiakiewicz, and Montanari]{pmlr-v134-mei21a}
Mei, S., Misiakiewicz, T., and Montanari, A.
\newblock Learning with invariances in random features and kernel models.
\newblock In Belkin, M. and Kpotufe, S. (eds.), \emph{Proceedings of Thirty Fourth Conference on Learning Theory}, volume 134 of \emph{Proceedings of Machine Learning Research}, pp.\  3351--3418. PMLR, 15--19 Aug 2021.
\newblock URL \url{https://proceedings.mlr.press/v134/mei21a.html}.

\bibitem[M{\"u}ller et~al.(2021)M{\"u}ller, Hollmann, Arango, Grabocka, and Hutter]{muller2021transformers}
M{\"u}ller, S., Hollmann, N., Arango, S.~P., Grabocka, J., and Hutter, F.
\newblock Transformers can do bayesian inference.
\newblock \emph{arXiv preprint arXiv:2112.10510}, 2021.

\bibitem[Nguyen \& Grover(2022)Nguyen and Grover]{nguyen2022transformer}
Nguyen, T. and Grover, A.
\newblock Transformer neural processes: Uncertainty-aware meta learning via sequence modeling.
\newblock \emph{arXiv preprint arXiv:2207.04179}, 2022.

\bibitem[Ronneberger et~al.(2015)Ronneberger, Fischer, and Brox]{ronneberger2015u}
Ronneberger, O., Fischer, P., and Brox, T.
\newblock U-net: Convolutional networks for biomedical image segmentation.
\newblock In \emph{Medical Image Computing and Computer-Assisted Intervention--MICCAI 2015: 18th International Conference, Munich, Germany, October 5-9, 2015, Proceedings, Part III 18}, pp.\  234--241. Springer, 2015.

\bibitem[Rozet \& Louppe(2023)Rozet and Louppe]{rozet2023score}
Rozet, F. and Louppe, G.
\newblock Score-based data assimilation.
\newblock \emph{arXiv preprint arXiv:2306.10574}, 2023.

\bibitem[Satorras et~al.(2021)Satorras, Hoogeboom, and Welling]{satorras2021n}
Satorras, V.~G., Hoogeboom, E., and Welling, M.
\newblock E (n) equivariant graph neural networks.
\newblock In \emph{International conference on machine learning}, pp.\  9323--9332. PMLR, 2021.

\bibitem[Schilling(2005)]{Schilling:2005:Measures_Integrals_and_Martingales}
Schilling, R.~L.
\newblock \emph{Measures, Integrals and Martingales}.
\newblock Cambridge University Press, 2005.
\newblock \doi{10.1017/CBO9780511810886}.

\bibitem[Su et~al.(2024)Su, Ahmed, Lu, Pan, Bo, and Liu]{su2024roformer}
Su, J., Ahmed, M., Lu, Y., Pan, S., Bo, W., and Liu, Y.
\newblock Roformer: Enhanced transformer with rotary position embedding.
\newblock \emph{Neurocomputing}, 568:\penalty0 127063, 2024.

\bibitem[Sun et~al.(2023)Sun, Yang, and Yoo]{sun2023neural}
Sun, Z., Yang, Y., and Yoo, S.
\newblock A neural pde solver with temporal stencil modeling.
\newblock \emph{arXiv preprint arXiv:2302.08105}, 2023.

\bibitem[Vaswani et~al.(2017)Vaswani, Shazeer, Parmar, Uszkoreit, Jones, Gomez, Kaiser, and Polosukhin]{vaswani2017attention}
Vaswani, A., Shazeer, N., Parmar, N., Uszkoreit, J., Jones, L., Gomez, A.~N., Kaiser, {\L}., and Polosukhin, I.
\newblock Attention is all you need.
\newblock \emph{Advances in neural information processing systems}, 30, 2017.

\bibitem[Wagstaff et~al.(2022)Wagstaff, Fuchs, Engelcke, Osborne, and Posner]{wagstaff2022universal}
Wagstaff, E., Fuchs, F.~B., Engelcke, M., Osborne, M.~A., and Posner, I.
\newblock Universal approximation of functions on sets.
\newblock \emph{The Journal of Machine Learning Research}, 23\penalty0 (1):\penalty0 6762--6817, 2022.

\bibitem[Wang et~al.(2022)Wang, Walters, and Yu]{wang2022data}
Wang, R., Walters, R., and Yu, R.
\newblock Data augmentation vs. equivariant networks: A theory of generalization on dynamics forecasting.
\newblock \emph{arXiv preprint arXiv:2206.09450}, 2022.

\bibitem[Zaheer et~al.(2017)Zaheer, Kottur, Ravanbakhsh, Poczos, Salakhutdinov, and Smola]{zaheer2017deep}
Zaheer, M., Kottur, S., Ravanbakhsh, S., Poczos, B., Salakhutdinov, R.~R., and Smola, A.~J.
\newblock Deep sets.
\newblock \emph{Advances in neural information processing systems}, 30, 2017.

\end{thebibliography}
\bibliographystyle{icml2024}

\newpage
\appendix
\onecolumn



\section{Relationship to LieTransformer}
\label{app:lietransformer}
At the core of the LieTransformer is the LieSelfAttention operation. Let group $G$ denote the set of symmetries we wish to be equivariant to, $\bfx_n \in \R^{d_x}$ denote the spatial coordinate of a point and $\bfz_n \in \R^{d_z}$ the token corresponding to this spatial coordinate. Each $\bfx_n$ corresponds to the coset $s(\bfx)H = \{s(\bfx)h | h\in H\}$, where the subgroup $H = \{g \in G | g\bfx_0 = \bfx_0\}$ is called the stabiliser of origin $\bfx_0$ and $s(x)\in G$ is a group element that maps $\bfx_0$ to $\bfx$. In the case of the group of translations, $s(\bfx)$ is just the translation that maps from the origin to $\bfx$. Thus, each spatial coordinate can be mapped to group elements in $G$. This can be thought of as lifting the feature map $f_{\mcX}: \bfx_n \rightarrow \bfz_n$ defined on $\mcX$ (i.e.\ $\R^{d_x}$) to a feature map $\mcL[f_{\mcX}]: g\rightarrow \bfz_n$ defined on $G$ (also $\R^{d_x}$ in the case of translations).

Let $G_f = \cup_{n=1}^N s(\bfx_n)H$. The LieSelfAttention operations updates the feature values $\bfz_n$ at spatial coordinates $\bfx_n$ (corresponding to group element $g_n$) as
\begin{equation}
    \tilde{\bfz}_n = \int \alpha_f(\bfz_n, \bfz_m, g_n, g_m) \bfW_V \bfz_m dg_m
\end{equation}
with the attention-weights given by
\begin{equation}
    \alpha_f(\bfz_n, \bfz_m, g_n, g_m) = \operatorname{softmax}\left(\phi\left(k_c\left(\bfz_n, \bfz_m\right), k_l\left(g_n^{-1}g_m\right)\right)\right).
\end{equation}

In the case of the group of translations, this corresponds to
\begin{equation}
    \tilde{\bfz}_n = \sum_{m=1}^M \alpha_f(\bfz_n, \bfz_m, g_n, g_m) \bfW_V\bfz_m.
\end{equation}
Choosing $k_c(\bfz_n, \bfz_m) = \bfz_n^T \bfW_{Q, h}\bfW_{K, h}^T\bfz_m$ and $k_l(\bfx_n, \bfx_m) = \bfx_n - \bfx_m$ recovers the \gls{te-mhsa} layer with $H=1$.

\section{Efficient Masked Attention}
\label{app:masked-attention}
Often, conditional independencies amongst the set of tokens---in the sense that the set $\{\bfz^{\ell}_n\}^{\ell=L}_{\ell=1}$ do not depend on the set $\{\bfz^{\ell}_m\}^{\ell=L}_{\ell=0}$ for some $n,\ m \in \{1, \ldots, N\}$---are desirable. This is typically achieved through masking, such that the pre-softmax activations are replaced by $\tilde{\alpha}^{\ell}_h$, where 
\begin{equation}
    \tilde{\alpha}^{\ell}_h(\bfz^{\ell}_n, \bfz^{\ell}_m) = \begin{cases}
        -\infty, & \text{$m\in A(n)$.} \\
        {\bfz^{\ell}_n}^T\bfW^{\ell}_{Q, h}\left[\bfW^{\ell}_{K, h}\right]^T\bfz^{\ell}_m, & \text{otherwise.}
    \end{cases}
\end{equation}
Here, $A(n) \subseteq \N_{\leq N}$ indexes the set of tokens we wish to make the update for token $\bfz^{\ell}_n$ independent of. If $A(n) = A$ (i.e.\ the same set of tokens are conditioned on for every $n$) then in practice it is more computationally efficient to use MHCA operations together with MHSA operations than it is to directly compute \Cref{eq:self-attention}. An MHCA operation uses the subset of tokens $\{\bfz^{\ell}_m | m\in A\}$ to update the complementary set of tokens $\{\bfz^{\ell}_n | n\in A^c\}$ in a computationally efficient manner. For $N$ tokens that solely depend on a subset of $N_1$ tokens, the computational complexity is reduced from $\order{N^2}$ using masked-\gls{mhsa} operations to $\order{NN_1}$ using \gls{mhsa} and \gls{mhca} operations.

In the context of \gls{tnp}, the tokens in both the context and target set are conditioned only on tokens in the context set. Thus, replacing masked-\gls{mhsa} operations reduces the computational complexity from $\order{(N_c + N_t)^2}$ to $\order{N_c^2 + N_cN_t}$. If the tokens in the target set are conditioned on context set tokens \emph{and} themselves, then we can easily modify the standard \gls{mhca} operation to include the individual target tokens.

\section{Pseudo-Token-Based Transformers}
\label{app:pseudo-token-transformers}
We illustrate the two types of pseudo-token-based transformers, the IST-style and perceiver-style, in \Cref{fig:perceiver} and \Cref{fig:ist}.

\begin{figure}[ht]
    \centering
    \begin{subfigure}[t]{0.28\textwidth}
        \centering
        \resizebox{\textwidth}{!}{
        \begin{tikzpicture}[rotate=90, transform shape]
        \node[align=center] (latent) {\rotatebox{-90}{$\bfU \in \R^{M\times D}$}};
        
        \node[mhca, right=0.5cm of latent] (mhca1) {\rotatebox{-90}{$\operatorname{MHCA}(\bfU^0$, $\bfZ)$}};
        \node[above=0.5cm of mhca1] (xc1) {\rotatebox{-90}{$\bfZ \in \R^{N\times D}$}};
        \node[mhsa, right=0.5cm of mhca1] (mhsa1) {\rotatebox{-90}{$\operatorname{MHSA}(\tilde{\bfU}^0)$}};
        \node[mhca, right=0.5cm of mhsa1] (mhca2) {\rotatebox{-90}{$\operatorname{MHCA}(\bfU^1$, $\bfZ)$}};
        \node[above=0.5cm of mhca2] (xc2) {\rotatebox{-90}{$\bfZ \in \R^{N\times D}$}};
        \node[mhsa, right=0.5cm of mhca2] (mhsa2) {\rotatebox{-90}{$\operatorname{MHSA}(\tilde{\bfU}^1)$}};
        \node[right=0.5cm of mhsa2] (elipse) {...};
        \node[mhsa, right=0.5cm of elipse] (mhsa3) {\rotatebox{-90}{$\operatorname{MHSA}(\tilde{\bfU}^{L-1})$}};
        
        \node[align=center, right=0.5cm of mhsa3] (output) {\rotatebox{-90}{$\bfU^L \in \R^{M\times D}$}};
        
        \draw[arrow] (latent) -- (mhca1);
        \draw[arrow](mhsa3) -- (output);
        \foreach \i [evaluate={\next=int(\i+1)}] in {1,2}
        {
            \draw[arrow] (mhca\i) -- (mhsa\i);
            \draw[arrow] (xc\i) -- (mhca\i);
            \ifnum\next<3
                \draw[arrow] (mhsa\i) -- (mhca\next);
            \fi
        }
        \draw[arrow] (mhsa2) -- (elipse);
        \draw[arrow] (elipse) -- (mhsa3);
        
        \end{tikzpicture}
        }
        \caption{Perceiver-style architecture.}
        \label{fig:perceiver}
    \end{subfigure}
    \hspace{20pt}
    \begin{subfigure}[t]{0.39\textwidth}
        \centering
        \resizebox{\textwidth}{!}{
        \begin{tikzpicture}[rotate=90, transform shape]
        \node[align=center] (input) {\rotatebox{-90}{$\bfZ \in \R^{N\times D}$}};
        
        %
        \node[mhca, right=0.5cm of input] (mhca1) {\rotatebox{-90}{$\operatorname{MHCA}(\bfZ^0$, $\bfU^1)$}};
        \node[mhca, below=1cm of mhca1] (mhca_latent1) {\rotatebox{-90}{$\operatorname{MHCA}(\bfU^0$, $\bfZ^0)$}};
        \node[mhca, right=1cm of mhca_latent1] (mhca_latent2) {\rotatebox{-90}{$\operatorname{MHCA}(\bfU^1$, $\bfZ^1)$}};
        \node[mhca, right=1cm of mhca1] (mhca2) {\rotatebox{-90}{$\operatorname{MHCA}(\bfZ^1$, $\bfU^2)$}};
        \node[right=0.5cm of mhca_latent2] (elipse_latent) {...};
        \node[right=0.5cm of mhca2] (elipse) {...};
        \node[mhca, right=0.5cm of elipse_latent] (mhca_latent3) {\rotatebox{-90}{$\operatorname{MHCA}(\bfU^{L-1}$, $\bfZ^{L-1})$}};
        \node[mhca, right=0.5cm of elipse] (mhca3) {\rotatebox{-90}{$\operatorname{MHCA}(\bfZ^{L-1}$, $\bfU^L)$}};

        \node[align=center, left=0.5cm of mhca_latent1] (latent) {\rotatebox{-90}{$\bfU \in \R^{M\times D}$}};

        \node[right=0.5cm of mhca3] (output) {\rotatebox{-90}{$\bfZ^L \in \R^{N \times D}$}};
        \node[right=0.5cm of mhca_latent3] (latent_output) {\rotatebox{-90}{$\bfU^L \in \R^{M \times D}$}};

        \draw[arrow] (input) -- (mhca1);
        \draw[arrow] (latent) -- (mhca_latent1);
        \foreach \i [evaluate={\next=int(\i+1)}] in {1, 2, 3}
        {
            \draw[arrow] (mhca_latent\i) -- (mhca\i);

            \ifnum\next<3
                \draw[arrow] (mhca\i) -- (mhca_latent\next);
                \draw[arrow] (mhca\i) -- (mhca\next);
                \draw[arrow] (mhca_latent\i) -- (mhca_latent\next);
            \fi
        }

        \draw[arrow] (mhca2) -- (elipse);
        \draw[arrow] (elipse) -- (mhca3);
        \draw[arrow] (mhca_latent2) -- (elipse_latent);
        \draw[arrow] (elipse_latent) -- (mhca_latent3);
        \draw[arrow] (mhca3) -- (output);
        \draw[arrow] (mhca_latent3) -- (latent_output);
        \draw[arrow] (input) -- (mhca_latent1);
        
        \end{tikzpicture}
        }
        \caption{IST-style architecture.}
        \label{fig:ist}
    \end{subfigure}
    \caption{Block diagrams the two pseudo-token-based transformer architectures.}
\end{figure}
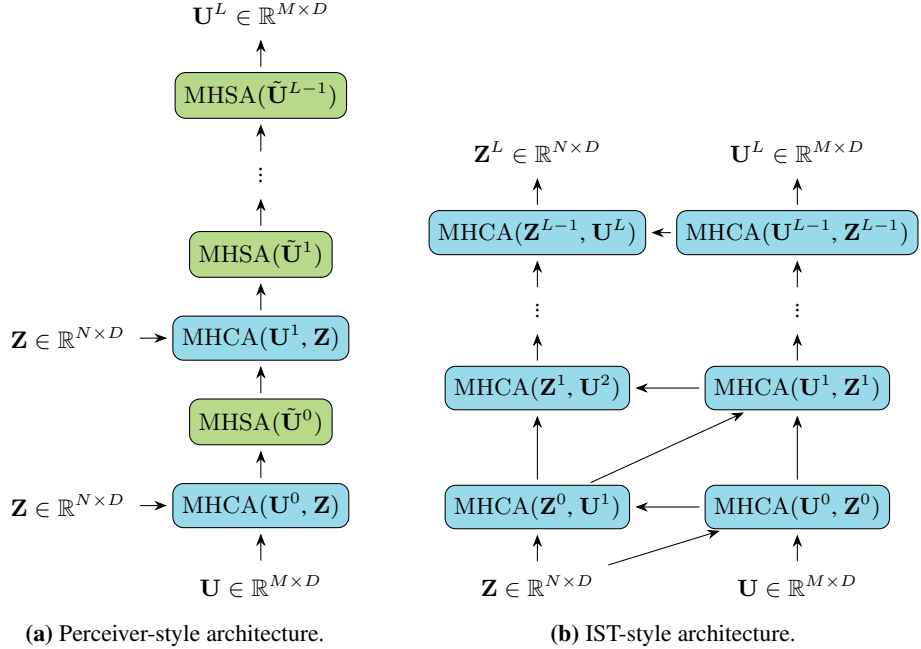

\section{General Form for Translation Equivariant and Permutation Invariant Functions}
\label{app:te-pe-functions}
Let $f\colon (\R^D)^N \to \R^D$ be a continuous function which is (1) permutation equivariant and (2) translation equivariant.
That $f$ is permutation equivariant means that, for all permutations $\sigma \in \mathbb{S}^{N}$ of $N$ elements,
\begin{equation}
    f(\bfx_1, \ldots, \bfx_N)
    = f(\bfx_{\sigma(1)}, \ldots, \bfx_{\sigma(N)});
\end{equation}
and that $f$ is translation equivariant means that, for all translations $\*\tau \in \R^D$,
\begin{equation}
    f(\bfx_1 + \*\tau, \ldots, \bfx_N + \*\tau)
    = f(\bfx_1, \ldots, \bfx_N) + \*\tau.
\end{equation}
Let $\alpha_1, \ldots, \alpha_N \in \R$ be \emph{any} set of weights, possibly negative, such that $\sum_{i=1}^N \alpha_i = 1$.
Then, using translation equivariance of $f$,
\begin{equation}
    f(\bfx_1, \ldots, \bfx_N)
    = \sum_{i=1}^N \alpha_i f(\bfx_1, \ldots, \bfx_N)
    = \sum_{i=1}^N \alpha_i f(\bfx_1 - \bfx_i, \ldots, \bfx_N - \bfx_i)
    + \sum_{i=1}^N \alpha_i \bfx_i.
\end{equation}
Since $f$ is also permutation equivariant, it can be written in the following way \citep{zaheer2017deep}:
\begin{equation}
    f(\bfx_1, \ldots, \bfx_N) = \rho\Big(\sum_{j=1}^N \phi(\bfx_j)\Big)
\end{equation}
for some continuous functions $\rho$ and $\phi$.
Therefore
\begin{equation} \label{eq:f_decomp}
    f(\bfx_1, \ldots, \bfx_N)
    = \sum_{i=1}^N \alpha_i \rho \Big(\sum_{j=1}^M \phi(\bfx_j - \bfx_i)\Big)
    + \sum_{i=1}^N \alpha_i \bfx_i.
\end{equation}
Note that this decomposes $f$ into a \emph{translation invariant} component and  \emph{translation equivariant} component.
We emphasise that this holds for \emph{any} set of weights $\alpha_1, \ldots, \alpha_N$.
In particular, these weights may depend on $\bfx_1,\ldots, \bfx_N$.
Finally, we can rewrite \eqref{eq:f_decomp} into residual form.
Let $K \in \mathbb{N}$, $1 \le K \le N$.
Then
\begin{equation}
    f(\bfx_1, \ldots, \bfx_N)
    = \bfx_K + \sum_{i=1}^N \alpha_i \rho \Big(\sum_{j=1}^M \phi(\bfx_j - \bfx_i)\Big)
    + \sum_{i=1}^N \alpha_i (\bfx_i - \bfx_K).
\end{equation}

\section{Proofs for Subsection \ref{subsec:translation-equivariance}}
\label{sec:spatial-generalisation}

In the following, we use the notation and definitions from \Cref{subsec:translation-equivariance}.
Recall that the collection of all data sets $\mathcal{S}$ includes the empty set $\varnothing$, which is the data set containing no data points.

\begin{proof}[Proof of \Cref{thm:te_iff_stat}]
    Suppose that the ground-truth stochastic process $f \sim P$ is stationary: $f \overset{\text{d}}{=} \mathsf{T}_{\*\tau} f$ for all $\*\tau \in \mcX$.
    Also suppose that $\pi'_P$ is translation invariant.
    Let $\mcD \in \mcS$ and $\*\tau \in \mcX$.
    Let $B$ be a cylinder set.
    Then, by the changes-of-variables formula for pushforward measures \citep[Theorem 14.1;][]{Schilling:2005:Measures_Integrals_and_Martingales},%
    \begin{align}
        \int_B \pi'_P(\mathsf{T}_{\*\tau} \mcD) \,\mathrm{d} P
        &= \int_B \pi'_P(\mcD) \circ \mathsf{T}^{-1}_{\*\tau} \,\mathrm{d} P &&\text{($\pi'_P$ is translation invariant)} \\
        &= \int_{\mathsf{T}^{-1}_{{\*\tau}}(B)} \pi'_P(\mcD) \,\mathrm{d} \mathsf{T}^{-1}_{{\*\tau}}(P) &&\text{(change of variables)}\\
        &= \int_{\mathsf{T}^{-1}_{{\*\tau}}(B)} \pi'_P(\mcD) \,\mathrm{d} P. && \text{($f$ is stationary)}
    \end{align}
    We conclude that $\pi_P$ is translation equivariant.
    
    Conversely, if $\pi_P$ is translation equivariant, then, considering the data set containing no data points $\varnothing$,
    \begin{equation}
        \mathsf{T}_{\*\tau} \pi_P(\varnothing) = \pi_P(\mathsf{T}_{\*\tau} \varnothing) = \pi_P(\varnothing)
        \quad\text{ for all ${\*\tau} \in \mcX$}.
    \end{equation}
    Since $\pi_P(\varnothing) = P$, this means that $f$ is stationary.
    Moreover, let $B$ be a cylinder set.
    Then
    \begin{align}
        \int_B \pi'_P(\mathsf{T}_{\*\tau} \mcD) \,\mathrm{d} P
        &= \int_{\mathsf{T}^{-1}_{{\*\tau}}(B)} \pi'_P(\mcD) \,\mathrm{d} P  && \text{($\pi_P$ is translation equivariant)} \\
        &= \int_{\mathsf{T}^{-1}_{{\*\tau}}(B)} \pi'_P(\mcD) \,\mathrm{d} \mathsf{T}^{-1}_{{\*\tau}}(P) &&\text{($f$ is stationary)}\\
        &= \int_B \pi'_P(\mcD) \circ \mathsf{T}_{-\*\tau} \,\mathrm{d} P. &&\text{(change of variables)}
    \end{align}
    Since this holds for all cylinder sets, $\pi'_P(\mathsf{T}_{\*\tau} \mcD) = \pi'_P(\mcD) \circ \mathsf{T}_{\*\tau}$ $P$--almost surely, so $\pi'_P$ is translation invariant.
\end{proof}

The proof of \Cref{thm:generalisation} follows the idea illustrated in \Cref{fig:generalisation}.
Let $\bfx_1 \oplus \bfx_2$ denote the concatenation of two vectors $\bfx_1$ and $\bfx_2$.

\begin{proof}[Proof of \Cref{thm:generalisation}]
    Let $M > 0$, $n \in \set{1, \ldots, N}$, $\vx \in [0, M]^n$, and $\mcD \in \mcS \cap \union_{n=0}^\infty ([0, M] \times \R)^n$.
    Sort and put the $n$ inputs $\vx$ into $B = \ceil*{M/\tfrac12 R}$ buckets $(B_i)_{i=1}^B$ such that $x_j \in [(i - 1)\cdot \tfrac12 R, i\cdot\tfrac12R]$ for all $j \in B_i$.
    More concisely written, $\vx_{B_i} \in [(i -1)\cdot \tfrac12 R, i\cdot \tfrac12 R]^{\abs{B_i}}$.
    Write $C_i = \union_{j=1}^{i - 1} B_i$.
    Let $\mcD_i$ be the sub--data set of $\mcD$ with inputs in $[\min(\vx_{B_{i-2}}), \max(\vx_{B_{i+1}})]$.
    
    If $\vy_1 \oplus \vy_2 \sim P_{\vx_1 \oplus \vx_2} \pi(\mcD)$, then denote the distribution of $\vy_1 \cond \vy_2$ by $P_{\vx_1 \cond \vx_2} \pi(\mcD)$.
    Use the chain rule for the KL divergence to decompose
    \begin{align}
        \KL{P_\vx \pi_1(\mcD)}{P_\vx \pi_2(\mcD)} 
        = \sum_{i=1}^B
        \E_{P_{\vx_{C_i}} \pi_1(\mcD)}[
                \KL{
                    P_{\vx_{B_i} \cond \vx_{C_i}} \pi_1(\mcD)
                }{
                    P_{\vx_{B_i} \cond \vx_{C_i}} \pi_2(\mcD)
                }
            ].
    \end{align}
    We focus on the $i$\textsuperscript{th} term in the sum.
    Using that $\pi_1(\mcD)$ and $\pi_2(\mcD)$ have receptive field $R$, we may drop the dependency on $B_1, \ldots, B_{i-2}$:
    \begin{align}
        \KL{
            P_{\vx_{B_i} \cond \vx_{C_i}} \pi_1(\mcD)
        }{
            P_{\vx_{B_i} \cond \vx_{C_i}} \pi_2(\mcD)
        }
        = \KL{
            P_{\vx_{B_i} \cond \vx_{B_{i-1}}} \pi_1(\mcD)
        }{
            P_{\vx_{B_i} \cond \vx_{B_{i-1}}} \pi_2(\mcD)
        }.
    \end{align}
    Therefore,
    \begin{align}
        \E_{P_{\vx_{C_i}} \pi_1(\mcD)}[ \KL{
            P_{\vx_{B_i} \cond \vx_{C_i}} \pi_1(\mcD)
        }{
            P_{\vx_{B_i} \cond \vx_{C_i}} \pi_2(\mcD)
        }]
        &= \E_{P_{\vx_{B_{i-1}}}\pi_1(\mcD)}[\KL{
            P_{\vx_{B_i} \cond \vx_{B_{i-1}}} \pi_1(\mcD)
            }{
            P_{\vx_{B_i} \cond \vx_{B_{i-1}}} \pi_2(\mcD)
            }] \\
        &\le \KL{
            P_{\vx_{B_i \cup B_{i-1}}} \pi_1(\mcD)
        }{
            P_{\vx_{B_i \cup B_{i-1}}} \pi_2(\mcD)
        }.
    \end{align}
    Next, we use that $\pi_1$ and $\pi_2$ also have receptive field $R$, allowing us to replace $\mcD$ with $\mcD_i$:
    \begin{align}
        \KL{
            P_{\vx_{B_i \cup B_{i-1}}} \pi_1(\mcD)
        }{
            P_{\vx_{B_i \cup B_{i-1}}} \pi_2(\mcD)
        }
        = \KL{
            P_{\vx_{B_i \cup B_{i-1}}} \pi_1(\mcD_i)
        }{
            P_{\vx_{B_i \cup B_{i-1}}} \pi_2(\mcD_i)
        }.
    \end{align}
    Finally, we use translation equivariance to shift everything back to the origin.
    Let $\tau_i$ be $\min(\vx_{B_{i-2}})$.
    By translation equivariance of $\pi_1$,
    \begin{equation}
        P_{\vx_{B_i \cup B_{i-1}}} \pi_1(\mcD_i)
        = P_{\vx_{B_i \cup B_{i-1}}} \T_{\tau_i} \pi_1(\T_{-\tau_i} \mcD_i)
        = P_{\vx_{B_i \cup B_{i-1}} - \tau_i} \pi_1(\T_{-\tau_i} \mcD_i)
    \end{equation}
    where by $\vx_{B_i \cup B_{i-1}} - \tau_i$ we mean subtraction elementwise.
    Crucially, note that all elements of $\vx_{B_i \cup B_{i-1}} - \tau_i$ and all inputs of $\T_{-\tau_i} \mcD_i$ lie in $[0, 4 \cdot \tfrac12 R]$.
    We have a similar equality for $\pi_2$.
    Therefore, putting everything together,
    \begin{align}
        \E_{P_{\vx_{C_i}}\pi_1(\mcD)}[ \KL{
            P_{\vx_{B_i} \cond \vx_{C_i}} \pi_1(\mcD)
        }{
            P_{\vx_{B_i} \cond \vx_{C_i}} \pi_2(\mcD)
        }]
        &\le \KL{
        P_{\vx_{B_i \cup B_{i-1}} - \tau_i} \pi_1(\T_{-\tau_i}\mcD_i)
        }{
        P_{\vx_{B_i \cup B_{i-1}} - \tau_i} \pi_2(\T_{-\tau_i}\mcD_i)
        },
    \end{align}
    which is less than $\e$ by the assumption of the theorem.
    The conclusion now follows.
\end{proof}
\endgroup


\section{Experimental Details and Additional Results}
\label{app:experimental-details}
\subsection{Synthetic 1-D Regression}
\label{subapp:1d-regression}
For each dataset, we first sample a kernel $k \sim U([k_{\text{se}}, k_{\text{pe}}, k_{\text{ma}-2.5}])$, where
\begin{equation}
    \begin{aligned}
        k_{\text{se}} &= \exp\left(-\frac{\left(x - x'\right)^2}{2\ell^2}\right) \\
        k_{\text{pe}} &= \exp\left(-2\sin^2\left(\frac{\pi}{\ell} \left(x - x'\right)\right)^2\right) \\
        k_{\text{ma}-2.5} &= \frac{2^{-1.5}}{\Gamma(2.5)}\left(\frac{\sqrt{5}\left(x - x'\right)}{\ell^2}\right)^{2.5} K_{2.5}\left(\frac{\sqrt{5}\left(x - x'\right)}{\ell^2}\right).
    \end{aligned}
\end{equation}
We sample $\ell \sim U(\log 0.25, \log 4)$, the number of context points $N_c \sim U(1, 64)$, the context inputs $x_{c, n} \sim U(-2, 2)$, and target inputs $x_{t, n} \sim U(-3, 3)$. All tasks use the same number of target points $N_t = 128$. The observations for each task are drawn from a GP with kernel
\begin{equation}
    k_{\text{obs}} = k + \sigma_n^2 \delta(x - x')
\end{equation}
where the observation noise $\sigma_n = 0.2$. 

For all models, we use an embedding / token size of $D_z = 128$, and decoder consisting of an MLP with two hidden layers of dimension $D_z$. The decoder parameterises the mean and pre-softplus variance of a Gaussian likelihood with heterogeneous noise. Model specific architectures are as follows:

\paragraph{\gls{tnp}}
The initial context tokens are obtained by passing the concatenation $[x, y, 1]$ through an MLP with two hidden layers of dimension $D_z$. The initial target tokens are obtained by passing the concatenation $[x, 0, 0]$ through the same MLP. The final dimension of the input acts as a `density' channel to indicate whether or not an observation is present. The \gls{tnp} encoder consists of five layers of self-attention and cross-attention blocks, each with $H=8$ attention heads with dimensions $D_V = D_{QK} = 16$. In each of the attention blocks, we apply a residual connection consisting of layer-normalisation to the input tokens followed by the attention mechanism. Following this there is another residual connection consisting of a layer-normalisation followed by a pointwise MLP with two hidden layers of dimension $D_z$.

\paragraph{\gls{pt-tnp}}
For the \gls{pt-tnp} models we use the same architecture dimensions as the \gls{tnp}. The initial pseudo-token values are sampled from a standard normal distribution. 

\paragraph{\gls{rcnp}}
We implement the simple \gls{rcnp} from \citet{huang2023practical}, as the memory requirements of the full \gls{rcnp} exceed the limits of our hardware for all but the simplest architectures. The simple \gls{rcnp} encoder is implemented as
\begin{equation}
    e(\bfx, \mcD_c) = \oplus_{n=1}^{N_c} \phi\left(\bfx - \bfx_{c, n}, \bfy_{c, n}\right)
\end{equation}
where $\oplus$ denotes a permutation invariant aggregation, for which we use the mean operation.\footnote{We found that for large datasets, the summation operation resulted in inputs to the decoder that were very large, resulting in numerical instabilities.} We implement the relational encoder $\phi: \R \times \R \rightarrow \R^{D_z}$ as an MLP with five hidden layers of dimension $D_z$.

\paragraph{\gls{cnp}}
The \gls{cnp} encoder is implemented as
\begin{equation}
    e(\bfx, \mcD_c) = \oplus_{n=1}^{N_c} \phi\left(\bfx_{c, n}, \bfy_{c, n}\right)
\end{equation}
where $\oplus$ denotes a permutation invariant aggregation, for which we use the mean operation. We implement $\phi: \R \times \R \rightarrow \R^{D_z}$ as an MLP with five hidden layers of dimension $D_z$.

\paragraph{\gls{convcnp}}
For the \gls{convcnp} model, we use a UNet \citep{ronneberger2015u} architecture for the CNN with 11 layers. We use $C=128$ input / output channels for the downwards layers, between which we apply pooling with size two. For the upwards layers, we use $2C$ input channels and $C$ output channels, as the input channels are formed from the output of the previous layer concatenated with the output of the corresponding downwards layer. Between the upwards layers we apply linear up-sampling to match the dimensions of the downwards layer. We use a kernel size of nine with a stride of one. The input domain is discretised with 64 points per units.

\paragraph{\gls{te-tnp}}
The \gls{te-tnp} model share a similar architecture with the \gls{tnp} model, with the attention blocks replaced with their translation equivariant counterparts. For the translation equivariant attention mechanisms, we implement $\rho^{\ell}: \R^H \times \R \rightarrow \R^H$ as an MLP with two hidden layers of dimension $D_z$. We implement $\phi^{\ell}: \R^H \rightarrow \R^H$ as an MLP with two hidden layers of dimension $D_z$. The initial context token embeddings are obtained by passing the context observations through an MLP with two hidden layers of dimension $D_z$. The initial target token embeddings are sampled from a standard normal.

\paragraph{\gls{te-pt-tnp}}
The \gls{te-pt-tnp} models adopt the same architecture choices as the \gls{te-tnp}. The initial pseudo-tokens and pseudo-input-locations are sampled from a standard normal.

\paragraph{Training Details}
For all models, we optimise the model parameters using AdamW \citep{loshchilov2017decoupled} with a learning rate of $5\times 10^{-4}$ and batch size of 16. Gradient value magnitudes are clipped at 0.5. We train for a maximum of 500 epochs, with each epoch consisting of 16,000 datasets (10,000 iterations per epoch). We evaluate the performance of each model on test 80,000 datasets.

\begin{table*}[ht]
    \centering
    \small
    \caption{Average log-likelihood (\textcolor{OliveGreen}{$\*\uparrow$}) on the test datasets for the synthetic 1-D regression experiment. $\Delta$ denotes the amount by which the range from which the context and target inputs and sampled from is shifted to the right at test time.}
    \addtolength{\tabcolsep}{-4.5pt} 
    \begin{tabular}{rcccccccccc}
        \toprule
          & \multicolumn{6}{c}{$\Delta$} \\
         \cmidrule{2-7}
         Model & $0.0$ & $0.2$ & $0.4$ & $0.6$ & $0.8$ & $1.0$ \\
         \midrule
         \gls{tnp} & $-0.48 \pm 0.00$ & $-0.48 \pm 0.01$ & $-0.49 \pm 0.01$ & $-0.50 \pm 0.01$ & $-0.52 \pm 0.01$ & $-0.57 \pm 0.01$ \\
         \gls{pt-tnp}-M8  & $-0.52 \pm 0.00$ & $-0.52 \pm 0.01$ & $-0.56 \pm 0.01$ & $-0.73 \pm 0.01$ & $-0.76 \pm 0.01$ & $-0.71 \pm 0.01$ \\
         \gls{pt-tnp}-M16 & $-0.49 \pm 0.00$ & $-0.50 \pm 0.01$ & $-0.53 \pm 0.01$ & $-0.57 \pm 0.01$ & $-0.60 \pm 0.01$ & $-0.64 \pm 0.01$ \\
         \gls{pt-tnp}-M32 & $-0.48 \pm 0.00$ & $-0.49 \pm 0.01$ & $-0.52 \pm 0.01$ & $-0.55 \pm 0.01$ & $-0.59 \pm 0.01$ & $-0.63 \pm 0.01$ \\
        \gls{cnp} & $-0.69 \pm 0.01$ & $-0.71 \pm 0.01$ & $-0.77 \pm 0.005$ & $-0.86 \pm 0.01$ & $-0.98 \pm 0.00$ & $-1.08 \pm 0.00$ \\
         \gls{rcnp} & $-0.58 \pm 0.01$ & $-0.58 \pm 0.01$ & $-0.58 \pm 0.01$ & $-0.58 \pm 0.01$ & $-0.58 \pm 0.01$ & $-0.58 \pm 0.01$ \\
         \gls{convcnp} & $\mathbf{-0.46 \pm 0.01}$ & $\mathbf{-0.46 \pm 0.01}$ & $\mathbf{-0.46 \pm 0.01}$ & $\mathbf{-0.46 \pm 0.01}$ & $\mathbf{-0.46 \pm 0.01}$ & $\mathbf{-0.46 \pm 0.01}$ \\
         \midrule
         \gls{te-tnp} & $\mathbf{-0.47 \pm 0.01}$ & $\mathbf{-0.47 \pm 0.01}$ & $\mathbf{-0.47 \pm 0.01}$ & $\mathbf{-0.47 \pm 0.01}$ & $\mathbf{-0.47 \pm 0.01}$ & $\mathbf{-0.47 \pm 0.01}$ \\
         \gls{te-pt-tnp}-M8 & $-0.50 \pm 0.00$ & $-0.50 \pm 0.00$ & $-0.50 \pm 0.00$ & $-0.50 \pm 0.00$ & $-0.50 \pm 0.00$ & $-0.50 \pm 0.00$ \\
         \gls{te-pt-tnp}-M16 & $-0.50 \pm 0.00$ & $-0.50 \pm 0.00$ & $-0.50 \pm 0.00$ & $-0.50 \pm 0.00$ & $-0.50 \pm 0.00$ & $-0.50 \pm 0.00$ \\
         \gls{te-pt-tnp}-M32 & $\mathbf{-0.46 \pm 0.00}$ & $\mathbf{-0.46 \pm 0.00}$ & $\mathbf{-0.46 \pm 0.00}$ & $\mathbf{-0.46 \pm 0.00}$ & $\mathbf{-0.46 \pm 0.00}$ & $\mathbf{-0.46 \pm 0.00}$ \\
         \bottomrule
    \end{tabular}
    \addtolength{\tabcolsep}{4.5pt}  
    \label{tab:1d-regression-results}
\end{table*}

\begin{table}[htb]
    \centering
    \caption{Time taken to complete the first 200 training epochs.}
    \begin{tabular}{r c}
    \toprule
         Model & Training time (hours) \\
         \midrule
         \gls{tnp} & $3.249$ \\
         \gls{pt-tnp}-M32 & $3.565$ \\
         \gls{cnp} & $1.672$ \\
         \gls{rcnp} & $2.326$ \\
         \gls{convcnp} & $3.466$ \\
         \midrule
         \gls{te-tnp} & $7.767$ \\
         \gls{te-pt-tnp}-M32 & $5.502$ \\
         \bottomrule
    \end{tabular}
    \label{tab:1d-training-times}
\end{table}

\subsection{Image Completion}
\label{subapp:image-completion}
The image completion dataset uses MNIST \citep{lecun1998gradient} and CIFAR-10. Each MNIST dataset consists of a $28\times 28$ black and white image, and each CIFAR-10 dataset consists of a $32\times 32$ RGB image. To construct each randomly translated dataset from the original images, we sample a horizontal and vertical translation, $h, v \sim U([-14, 14])$ for MNIST and $h, v \sim U([-16, 16])$. The original image canvas is expanded to $56 \times 56$ for MNIST and $64 \times 64$ for CIFAR-10, with the new pixels given output values of zero, and the translated image is inserted. The training set consists of 60,000 images for T-MNIST and 50,000 images for T-CIFAR-10 (i.e.\ a single random translation applied to each image in the original training sets). The test set consists of 10,000 images for both T-MNIST and T-CIFAR-10. For each dataset, we sample the $N_c \sim U\left(\frac{N}{100}, \frac{N}{3}\right)$ and set $N_t$ to the remaining pixels.

For all models, we use an embedding / token size of $D_z = 32$. This was chosen to be relatively small due to the limitations of the hardware available (both the \gls{tnp} and \gls{rcnp} models have a memory complexity that scales with $\order{N_cN_T}$). For both the T-MNIST and T-CIFAR-10, we use a Gaussian likelihood with homogeneous noise. Similar to \citet{foong2020meta}, which found that this significantly improve training stability. Model specific architectures are as follows:

\paragraph{\gls{pt-tnp}}
Same as \Cref{subapp:1d-regression}. 

\paragraph{\gls{rcnp}}
Same as \Cref{subapp:1d-regression}.

\paragraph{\gls{cnp}}
Same as \Cref{subapp:1d-regression}.

\paragraph{\gls{convcnp}}
For the \gls{convcnp} model, we use a standard CNN 5 layers. We use $C=64$ input / output channels for each layer. As the input domain is already discretised, discretisation is not needed.

\paragraph{\gls{te-pt-tnp}}
Same as \Cref{subapp:1d-regression}.

\paragraph{Training Details}
For all models, we optimise the model parameters using AdamW \citep{loshchilov2017decoupled} with a learning rate of $5\times 10^{-4}$ and batch size of 16 (8 for the \gls{te-pt-tnp} and \gls{rcnp}). Gradient value magnitudes are clipped at 0.5. We train for a maximum of 500 epochs, with each epoch consisting of 1,000 iterations. We evaluate the performance of each model on the entire test set.

\begin{figure*}[ht]
    \centering
    \begin{subfigure}[t]{0.19\textwidth}
        \includegraphics[width=\textwidth,trim={10, 10, 10, 10},clip]{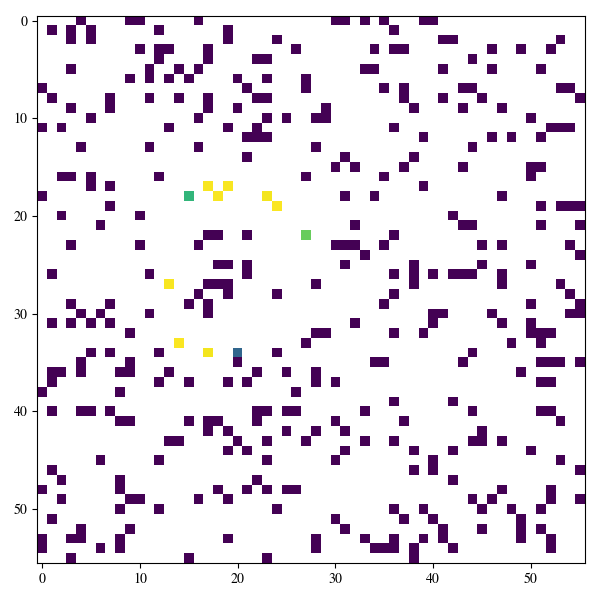}
    \end{subfigure}
    \begin{subfigure}[t]{0.19\textwidth}
        \includegraphics[width=\textwidth,trim={10, 10, 10, 10},clip]{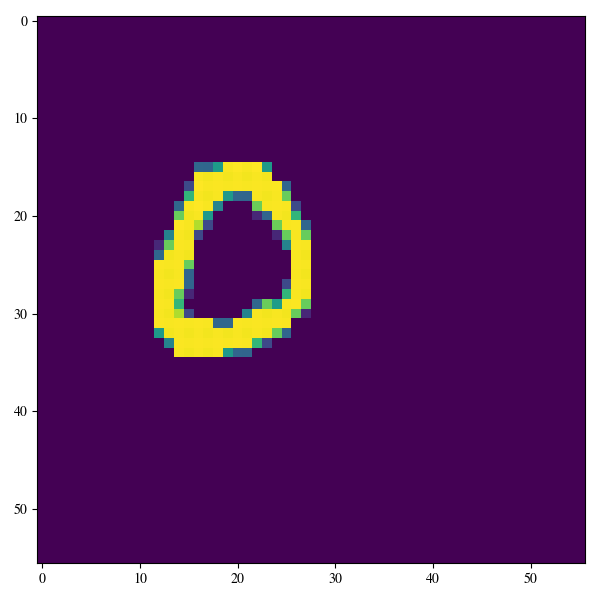}
    \end{subfigure}
    \begin{subfigure}[t]{0.19\textwidth}
        \includegraphics[width=\textwidth,trim={10, 10, 10, 10},clip]{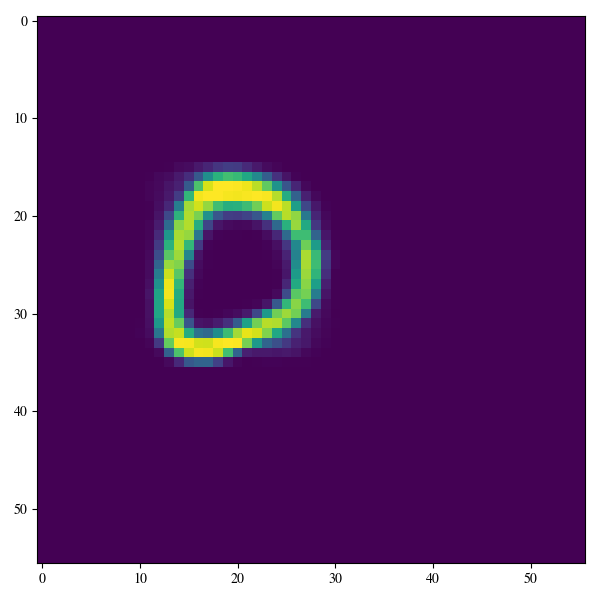}
    \end{subfigure}
    \begin{subfigure}[t]{0.19\textwidth}
        \includegraphics[width=\textwidth,trim={10, 10, 10, 10},clip]{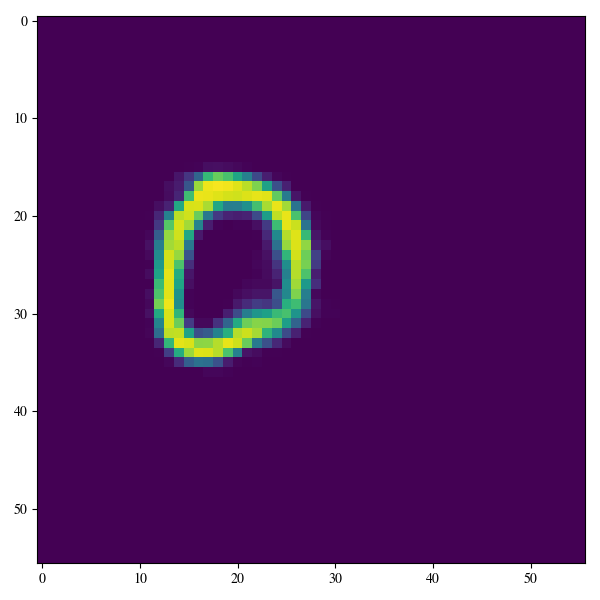}
    \end{subfigure}
    \begin{subfigure}[t]{0.19\textwidth}
        \includegraphics[width=\textwidth,trim={10, 10, 10, 10},clip]{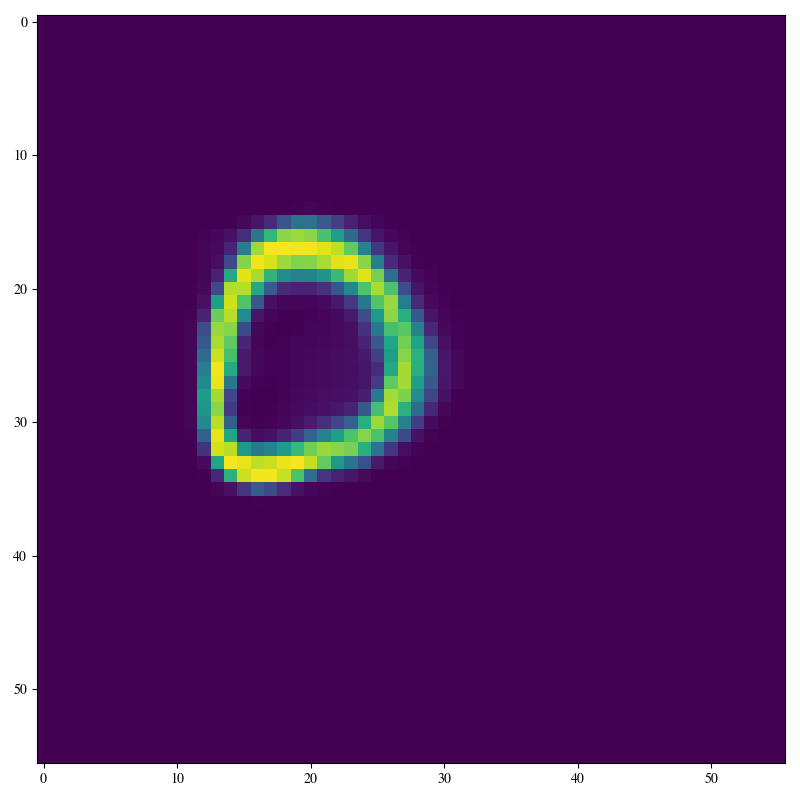}
    \end{subfigure}
    \begin{subfigure}[t]{0.19\textwidth}
        \includegraphics[width=\textwidth,trim={10, 10, 10, 10},clip]{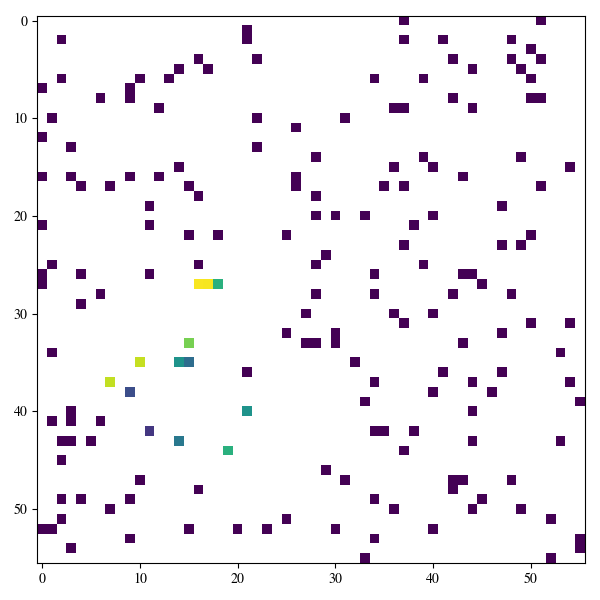}
    \end{subfigure}
    \begin{subfigure}[t]{0.19\textwidth}
        \includegraphics[width=\textwidth,trim={10, 10, 10, 10},clip]{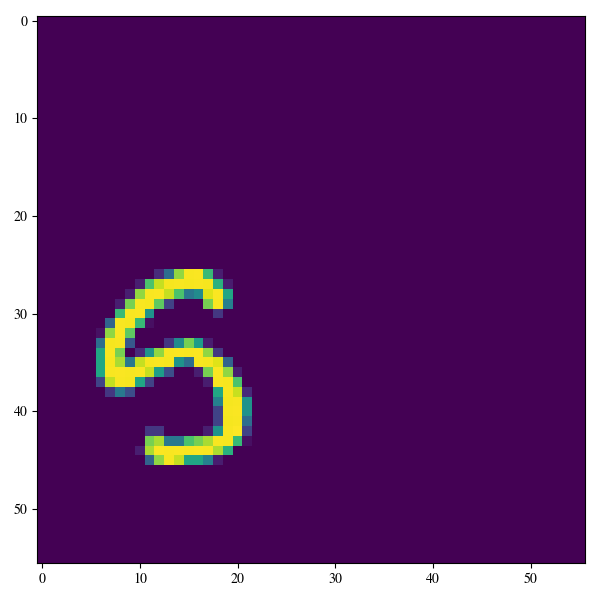}
    \end{subfigure}
    \begin{subfigure}[t]{0.19\textwidth}
        \includegraphics[width=\textwidth,trim={10, 10, 10, 10},clip]{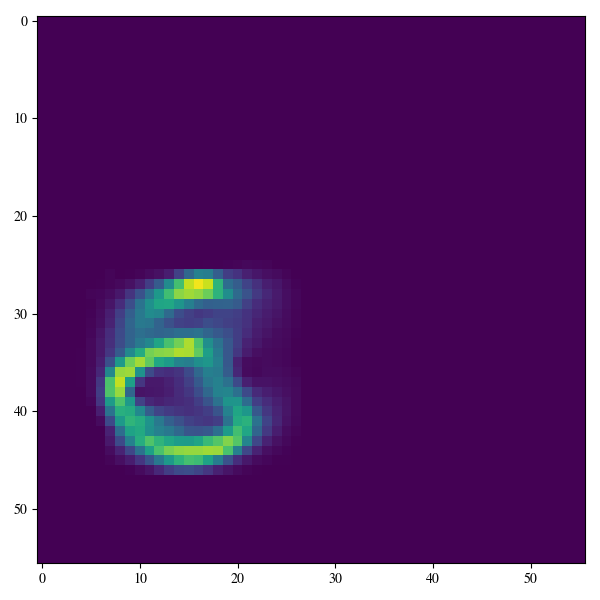}
    \end{subfigure}
    \begin{subfigure}[t]{0.19\textwidth}
        \includegraphics[width=\textwidth,trim={10, 10, 10, 10},clip]{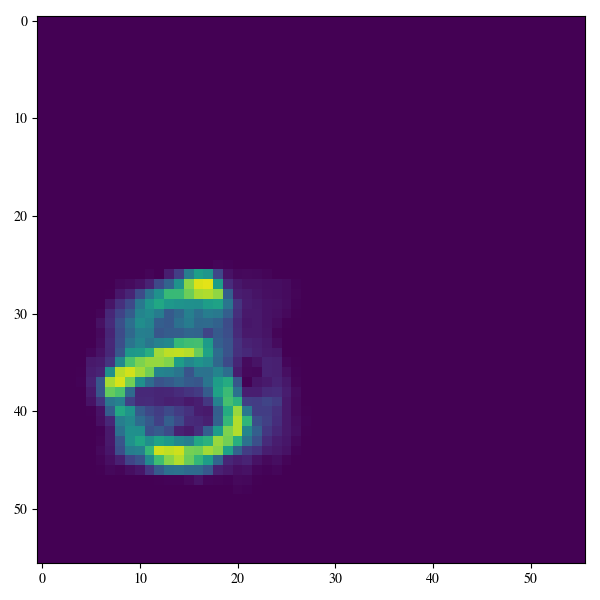}
    \end{subfigure}
    \begin{subfigure}[t]{0.19\textwidth}
        \includegraphics[width=\textwidth,trim={10, 10, 10, 10},clip]{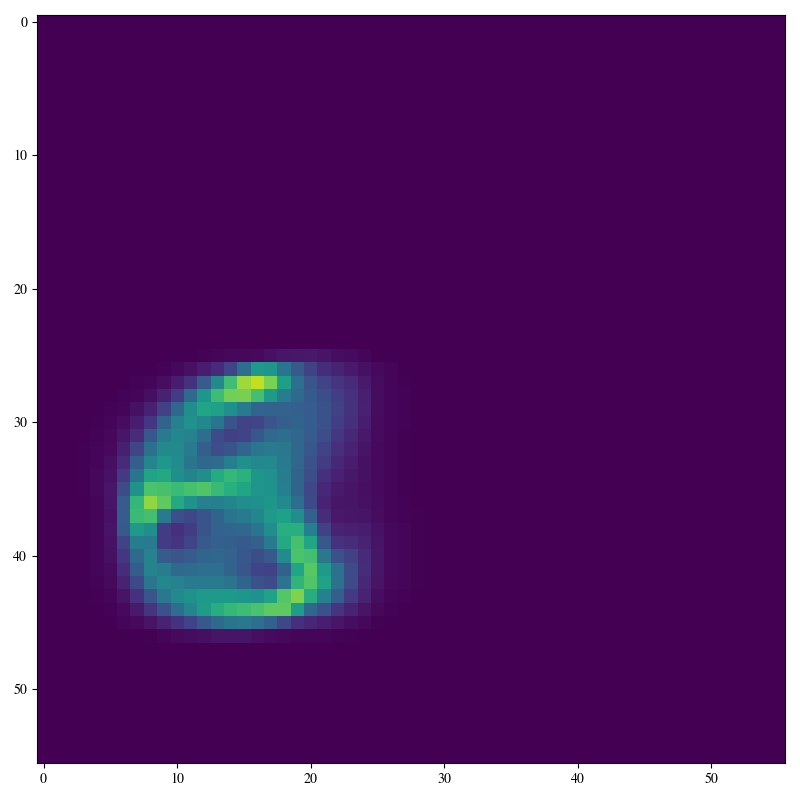}
    \end{subfigure}
    \begin{subfigure}[t]{0.19\textwidth}
        \includegraphics[width=\textwidth,trim={10, 10, 10, 10},clip]{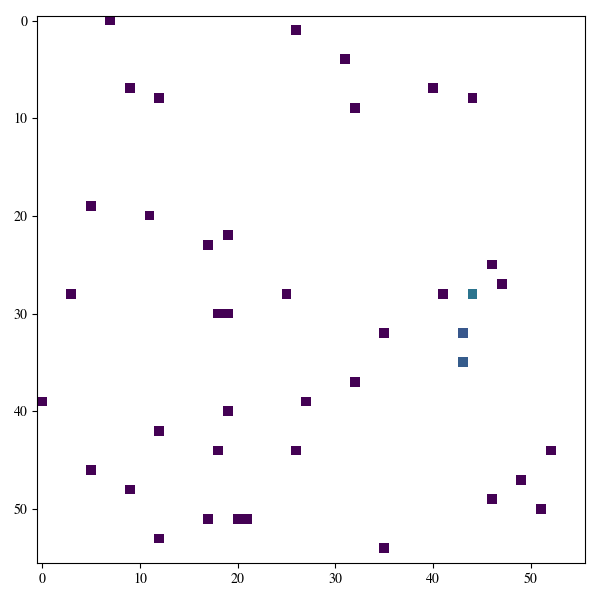}
    \end{subfigure}
    \begin{subfigure}[t]{0.19\textwidth}
        \includegraphics[width=\textwidth,trim={10, 10, 10, 10},clip]{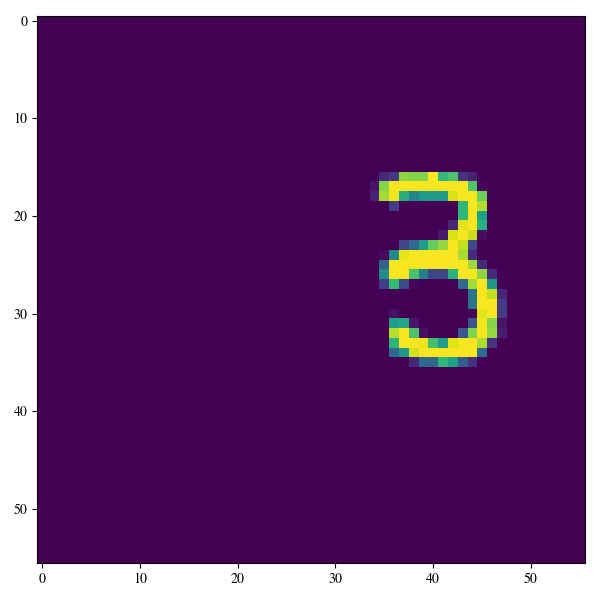}
    \end{subfigure}
    \begin{subfigure}[t]{0.19\textwidth}
        \includegraphics[width=\textwidth,trim={10, 10, 10, 10},clip]{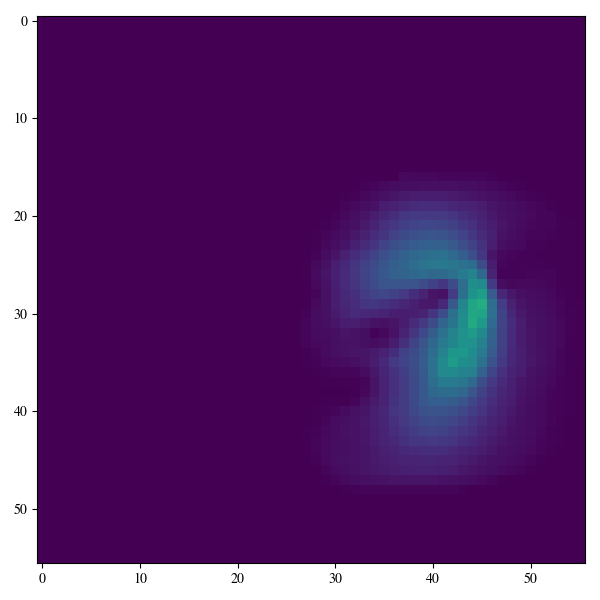}
    \end{subfigure}
    \begin{subfigure}[t]{0.19\textwidth}
        \includegraphics[width=\textwidth,trim={10, 10, 10, 10},clip]{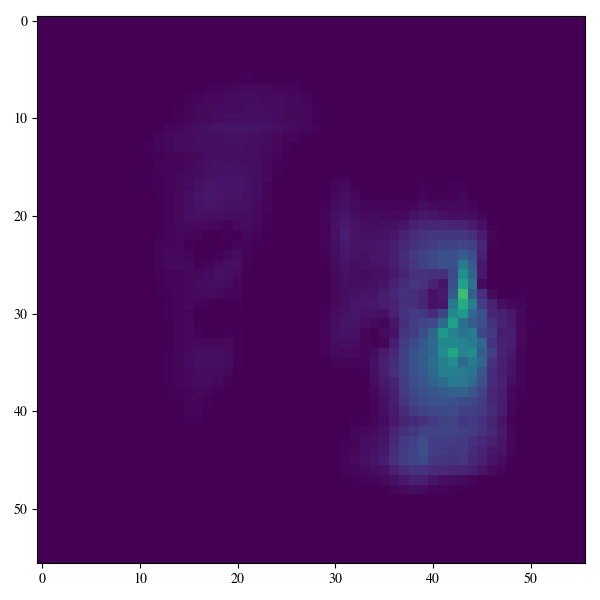}
    \end{subfigure}
    \begin{subfigure}[t]{0.19\textwidth}
        \includegraphics[width=\textwidth,trim={10, 10, 10, 10},clip]{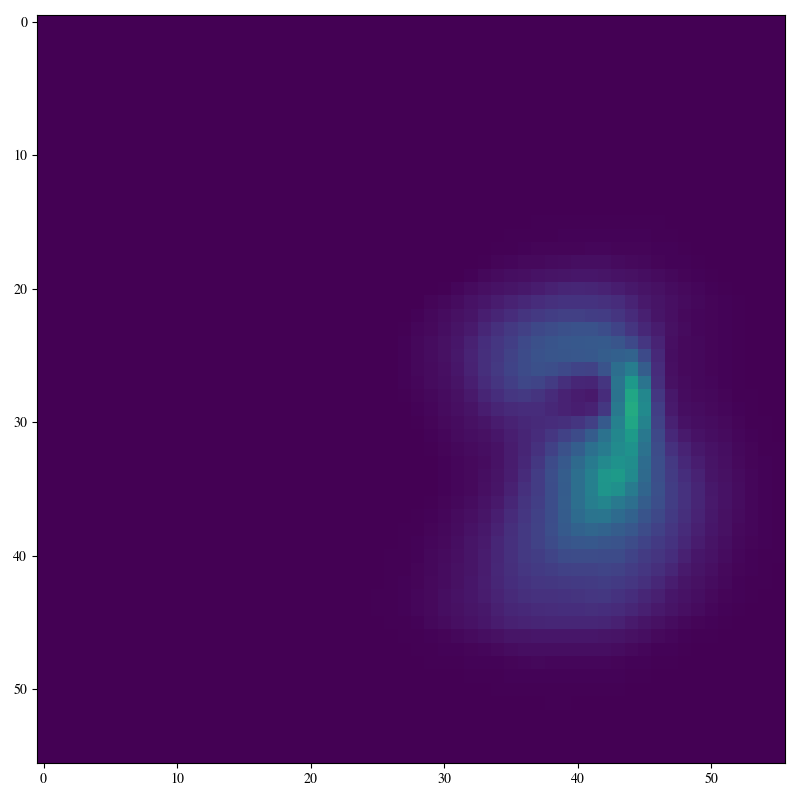}
    \end{subfigure}
    \begin{subfigure}[t]{0.19\textwidth}
        \includegraphics[width=\textwidth,trim={10, 10, 10, 10},clip]{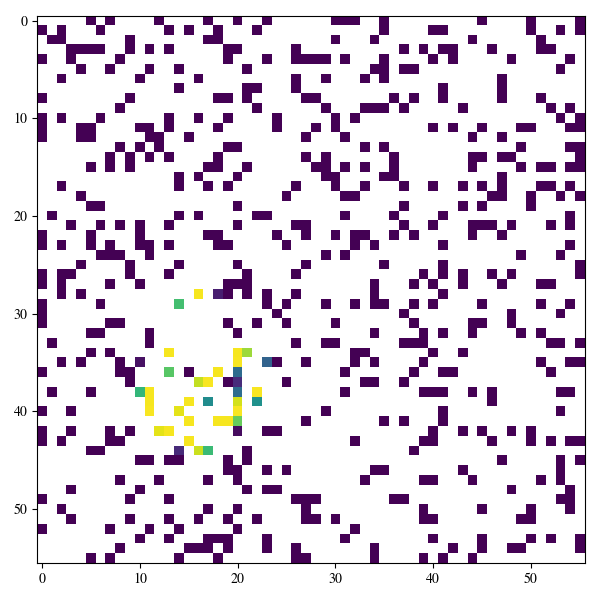}
    \end{subfigure}
    \begin{subfigure}[t]{0.19\textwidth}
        \includegraphics[width=\textwidth,trim={10, 10, 10, 10},clip]{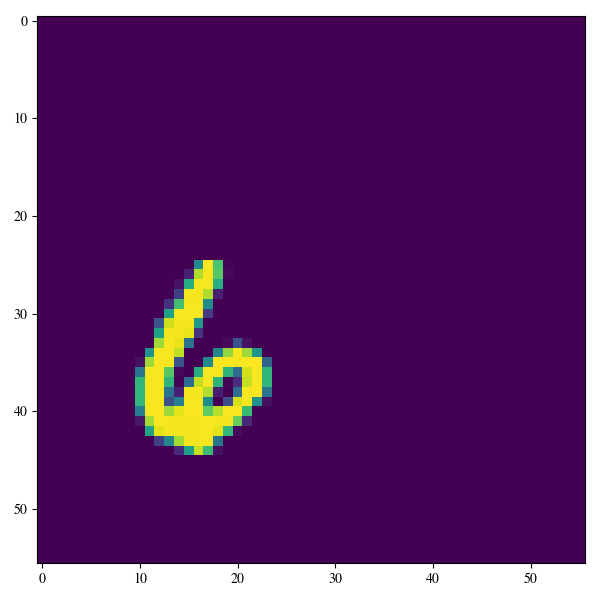}
    \end{subfigure}
    \begin{subfigure}[t]{0.19\textwidth}
        \includegraphics[width=\textwidth,trim={10, 10, 10, 10},clip]{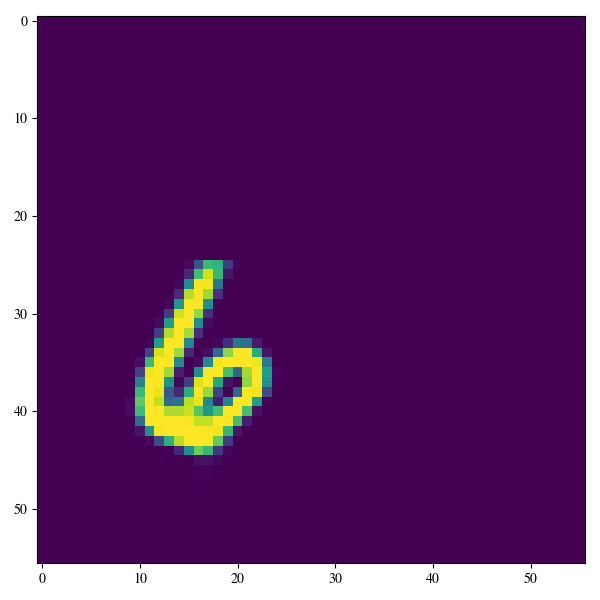}
    \end{subfigure}
    \begin{subfigure}[t]{0.19\textwidth}
        \includegraphics[width=\textwidth,trim={10, 10, 10, 10},clip]{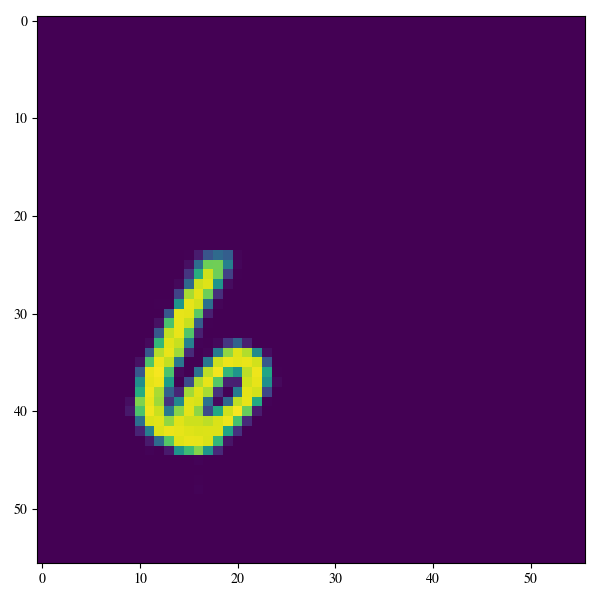}
    \end{subfigure}
    \begin{subfigure}[t]{0.19\textwidth}
        \includegraphics[width=\textwidth,trim={10, 10, 10, 10},clip]{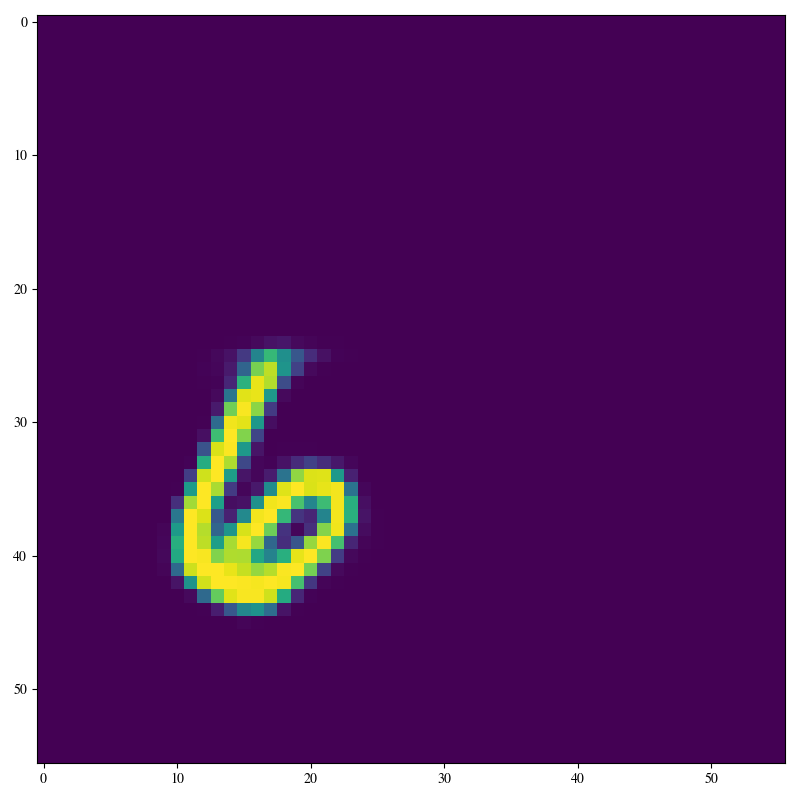}
    \end{subfigure}
    \begin{subfigure}[t]{0.19\textwidth}
        \includegraphics[width=\textwidth,trim={10, 10, 10, 10},clip]{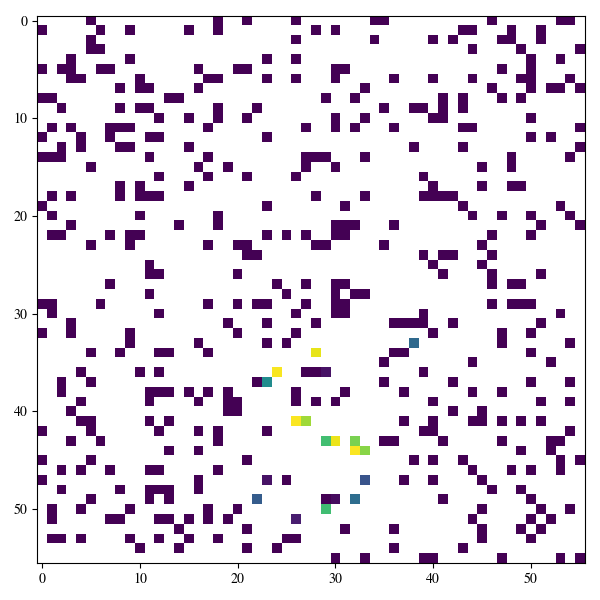}
        \caption{Context dataset.}
    \end{subfigure}
    \begin{subfigure}[t]{0.19\textwidth}
        \includegraphics[width=\textwidth,trim={10, 10, 10, 10},clip]{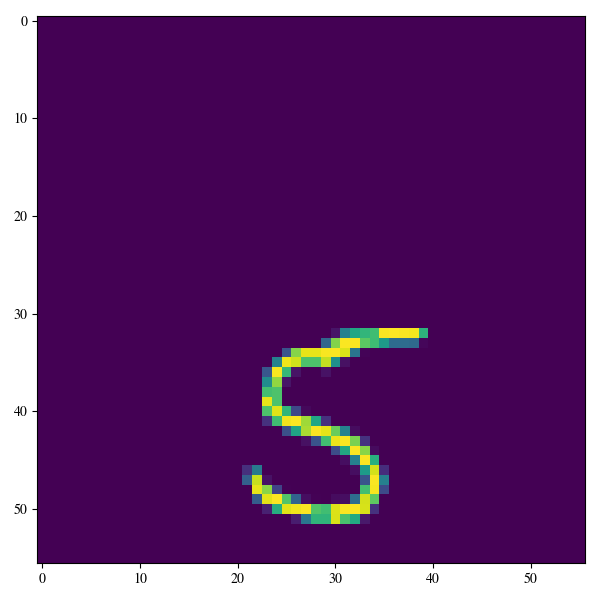}
        \caption{Ground truth data.}
    \end{subfigure}
    \begin{subfigure}[t]{0.19\textwidth}
        \includegraphics[width=\textwidth,trim={10, 10, 10, 10},clip]{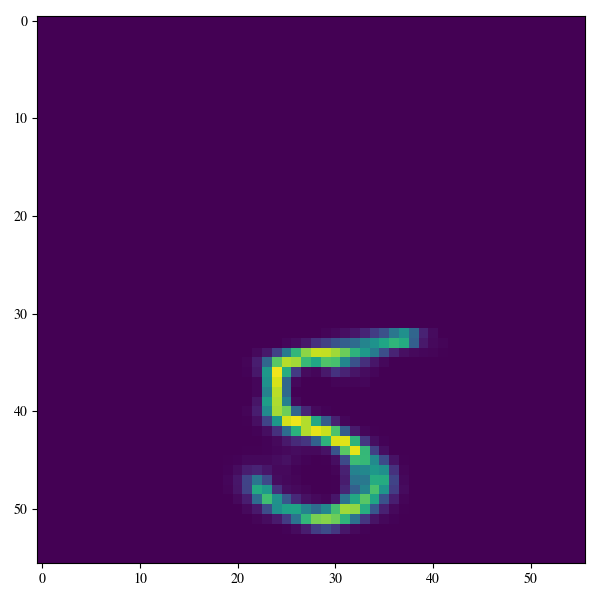}
        \caption{\gls{te-pt-tnp}-M64.}
    \end{subfigure}
    \begin{subfigure}[t]{0.19\textwidth}
        \includegraphics[width=\textwidth,trim={10, 10, 10, 10},clip]{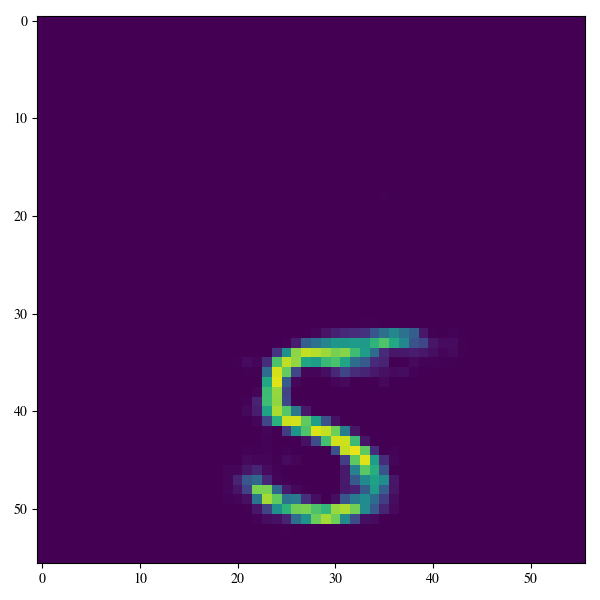}
        \caption{\gls{convcnp}.}
    \end{subfigure}
    \begin{subfigure}[t]{0.19\textwidth}
        \includegraphics[width=\textwidth,trim={10, 10, 10, 10},clip]{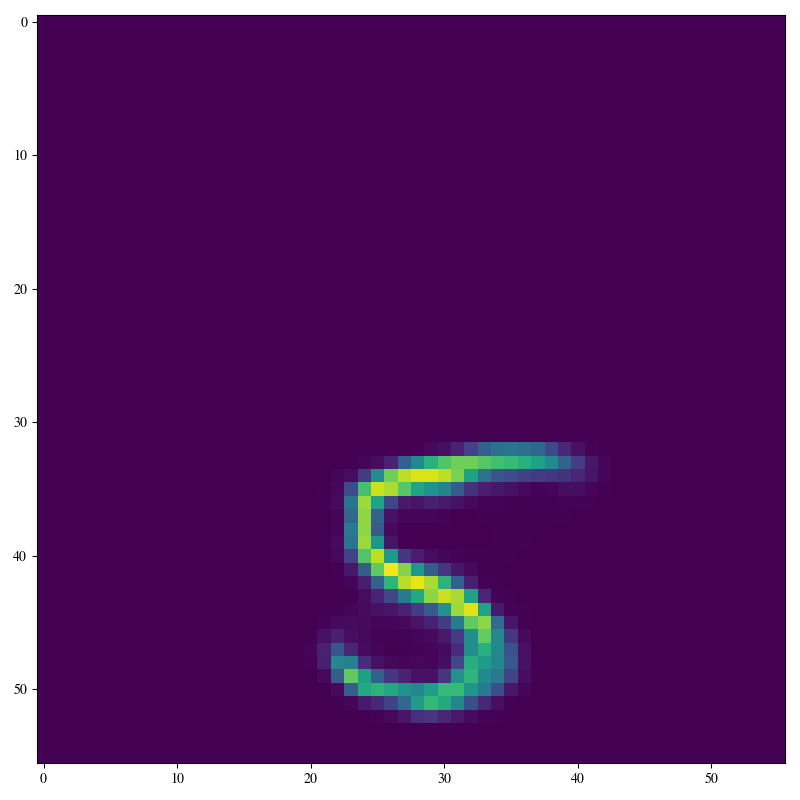}
        \caption{\gls{pt-tnp}-M64.}
    \end{subfigure}
    \caption{A comparison between the predictive mean of the \gls{te-pt-tnp}-M64 model, \gls{convcnp} model and \gls{pt-tnp}-M64 model, given the context datasets on the left.}
    \label{fig:mnist-translated}
\end{figure*}

\begin{figure*}[ht]
    \centering
    \begin{subfigure}[t]{0.19\textwidth}
        \includegraphics[width=\textwidth,trim={10, 10, 10, 10},clip]{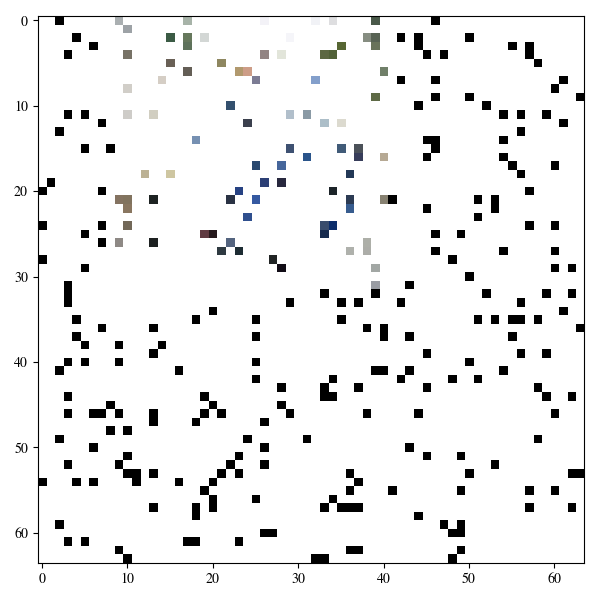}
    \end{subfigure}
    \begin{subfigure}[t]{0.19\textwidth}
        \includegraphics[width=\textwidth,trim={10, 10, 10, 10},clip]{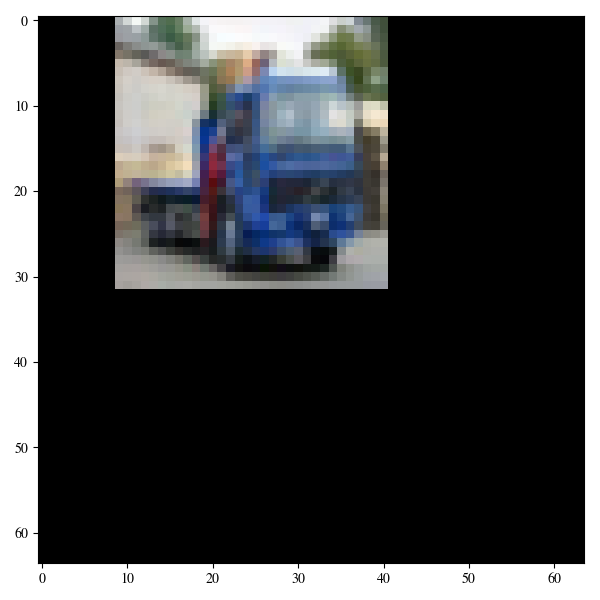}
    \end{subfigure}
    \begin{subfigure}[t]{0.19\textwidth}
        \includegraphics[width=\textwidth,trim={10, 10, 10, 10},clip]{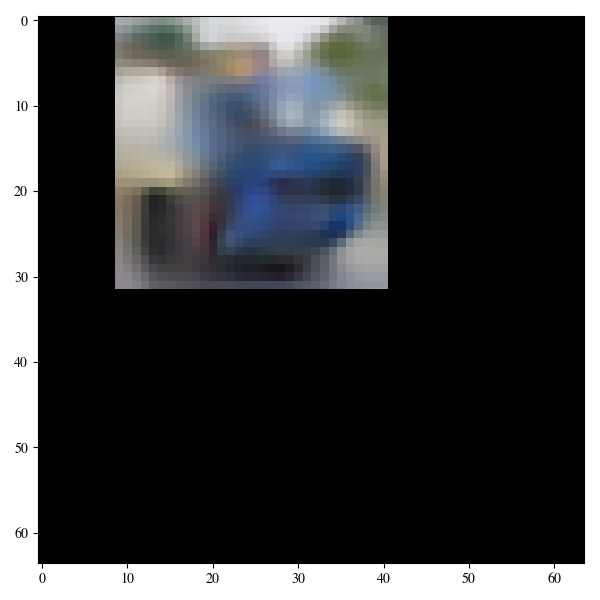}
    \end{subfigure}
    \begin{subfigure}[t]{0.19\textwidth}
        \includegraphics[width=\textwidth,trim={10, 10, 10, 10},clip]{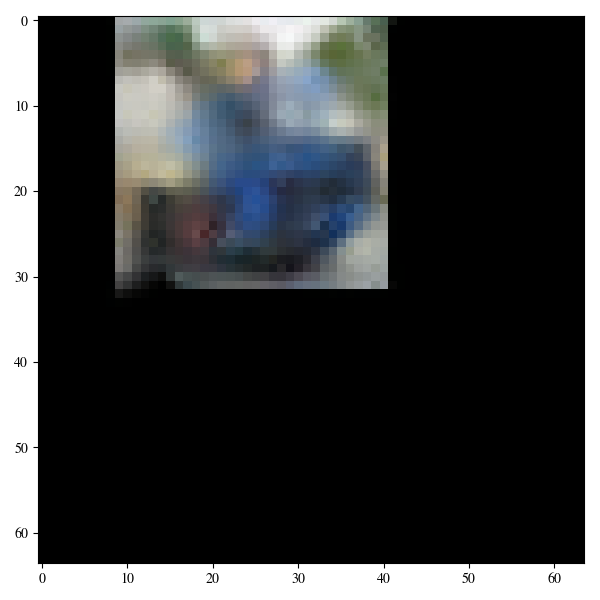}
    \end{subfigure}
    \begin{subfigure}[t]{0.19\textwidth}
        \includegraphics[width=\textwidth,trim={10, 10, 10, 10},clip]{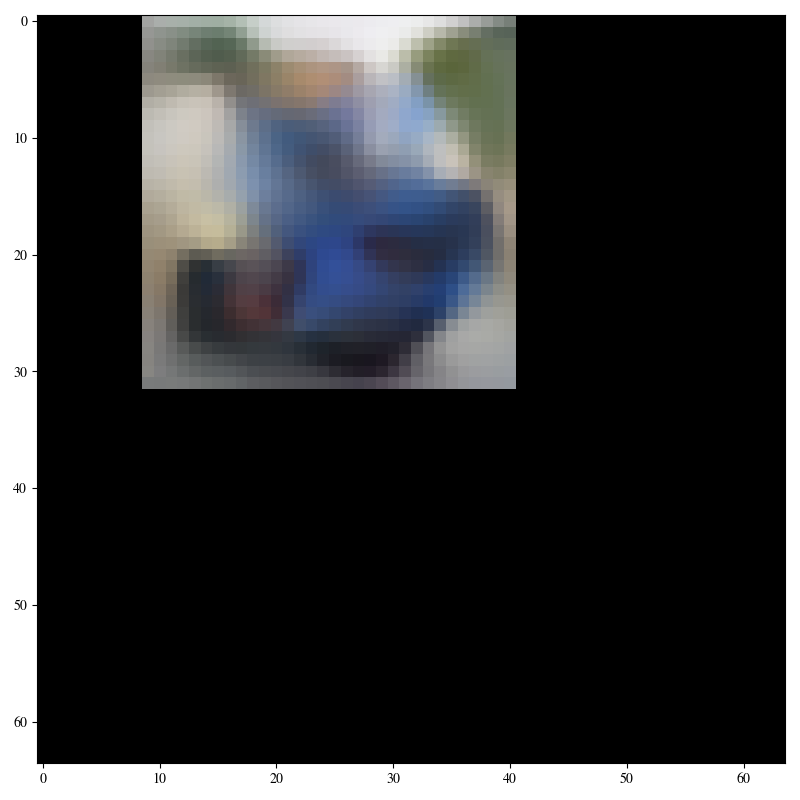}
    \end{subfigure}
    \begin{subfigure}[t]{0.19\textwidth}
        \includegraphics[width=\textwidth,trim={10, 10, 10, 10},clip]{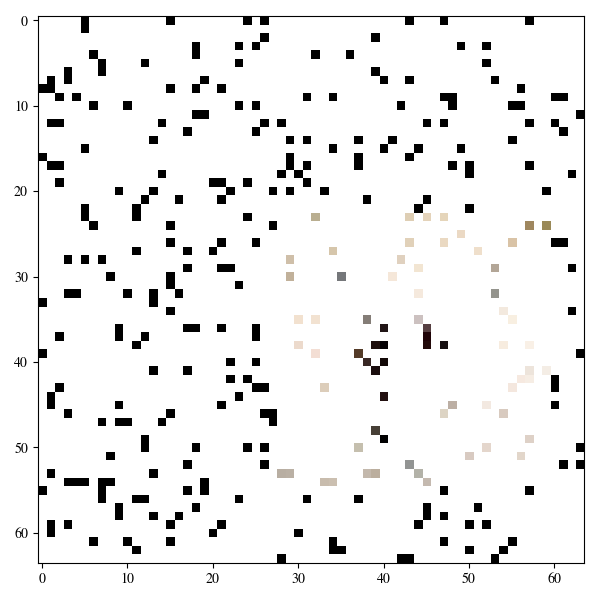}
    \end{subfigure}
    \begin{subfigure}[t]{0.19\textwidth}
        \includegraphics[width=\textwidth,trim={10, 10, 10, 10},clip]{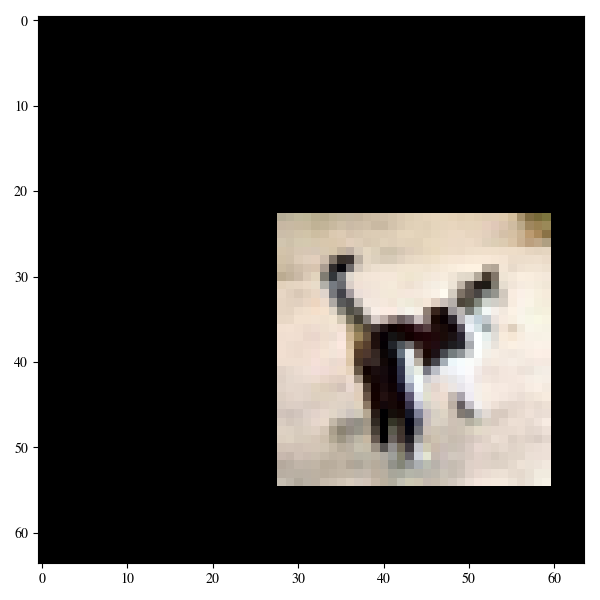}
    \end{subfigure}
    \begin{subfigure}[t]{0.19\textwidth}
        \includegraphics[width=\textwidth,trim={10, 10, 10, 10},clip]{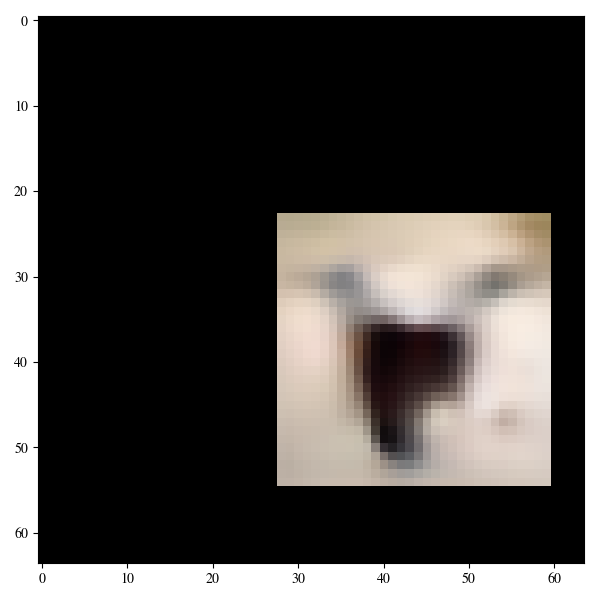}
    \end{subfigure}
    \begin{subfigure}[t]{0.19\textwidth}
        \includegraphics[width=\textwidth,trim={10, 10, 10, 10},clip]{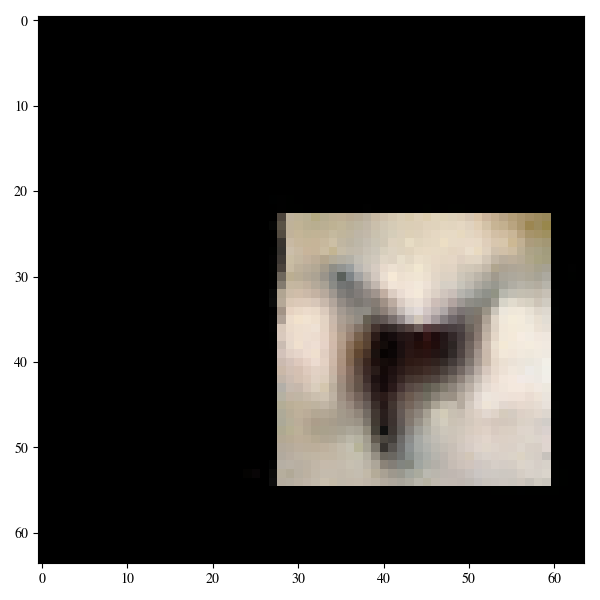}
    \end{subfigure}
    \begin{subfigure}[t]{0.19\textwidth}
        \includegraphics[width=\textwidth,trim={10, 10, 10, 10},clip]{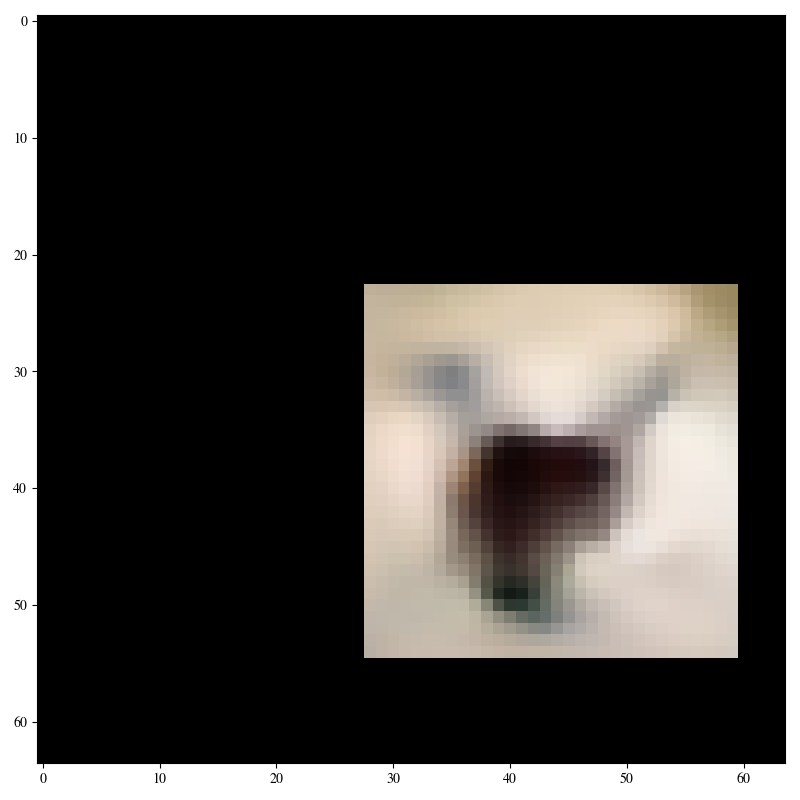}
    \end{subfigure}
    \begin{subfigure}[t]{0.19\textwidth}
        \includegraphics[width=\textwidth,trim={10, 10, 10, 10},clip]{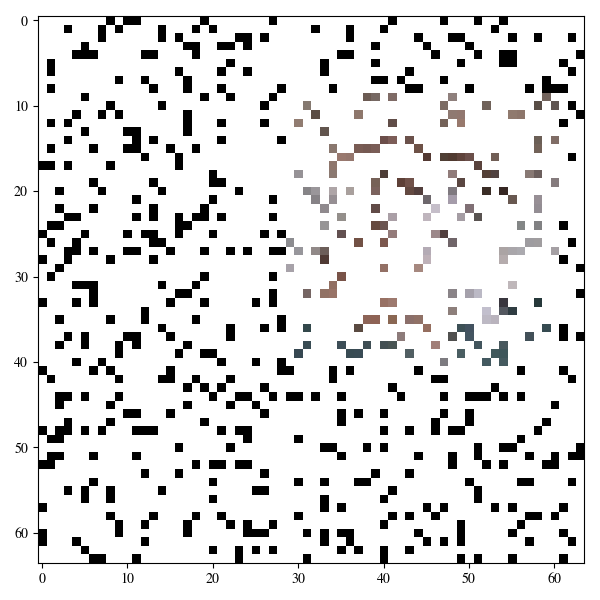}
    \end{subfigure}
    \begin{subfigure}[t]{0.19\textwidth}
        \includegraphics[width=\textwidth,trim={10, 10, 10, 10},clip]{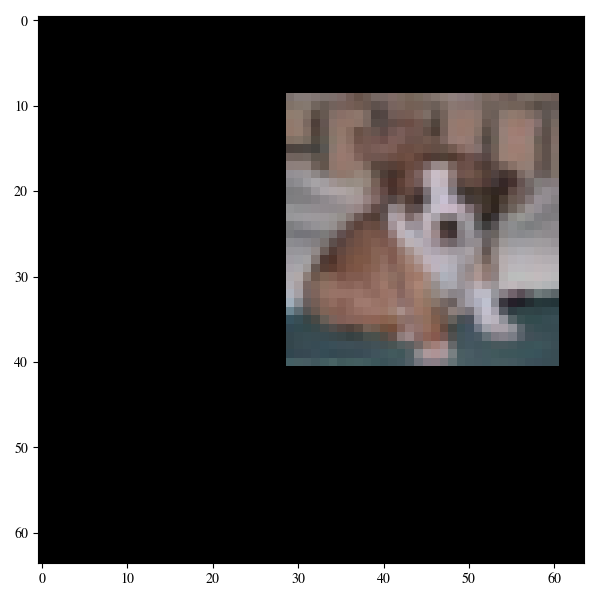}
    \end{subfigure}
    \begin{subfigure}[t]{0.19\textwidth}
        \includegraphics[width=\textwidth,trim={10, 10, 10, 10},clip]{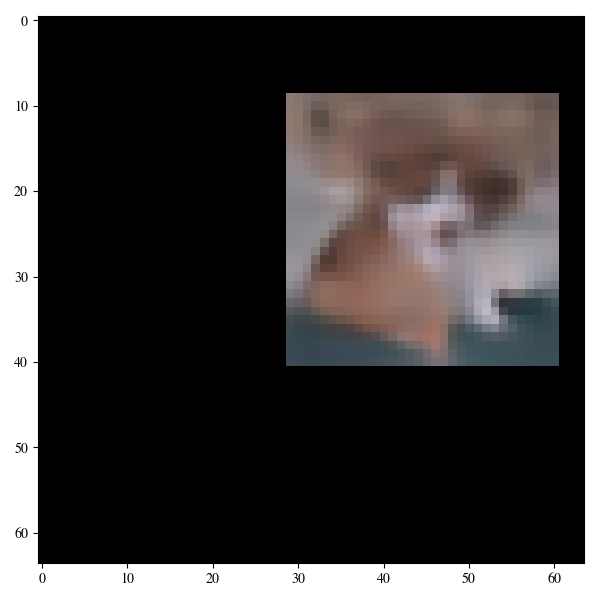}
    \end{subfigure}
    \begin{subfigure}[t]{0.19\textwidth}
        \includegraphics[width=\textwidth,trim={10, 10, 10, 10},clip]{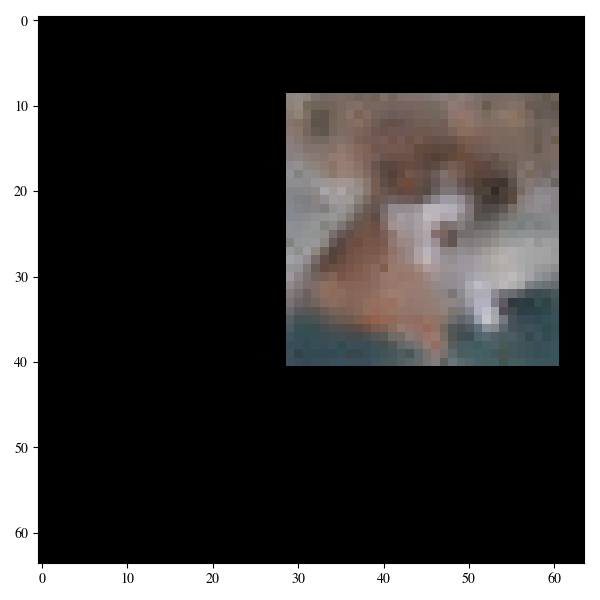}
    \end{subfigure}
    \begin{subfigure}[t]{0.19\textwidth}
        \includegraphics[width=\textwidth,trim={10, 10, 10, 10},clip]{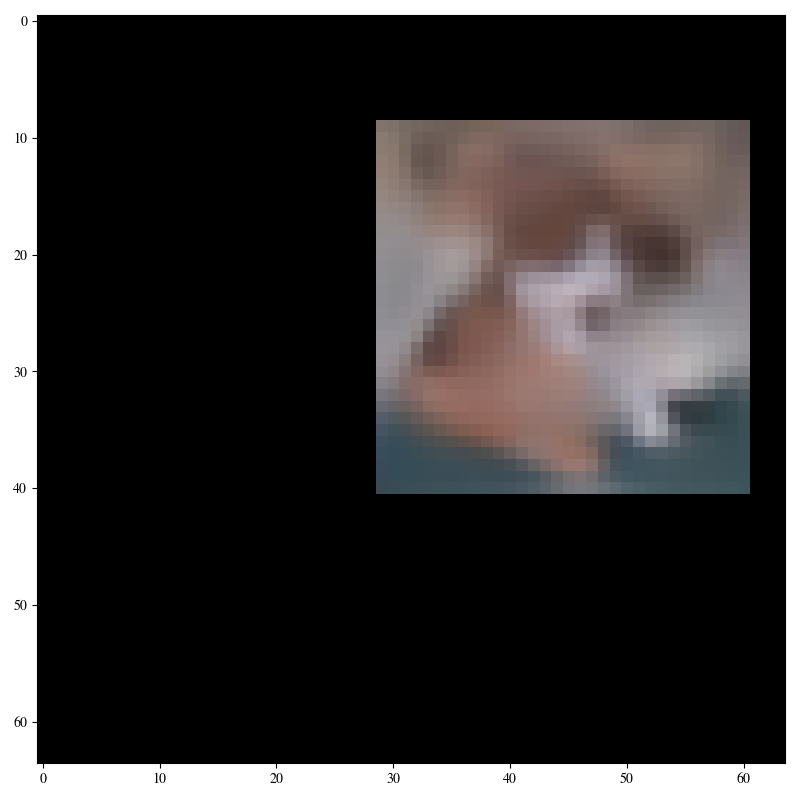}
    \end{subfigure}
    \begin{subfigure}[t]{0.19\textwidth}
        \includegraphics[width=\textwidth,trim={10, 10, 10, 10},clip]{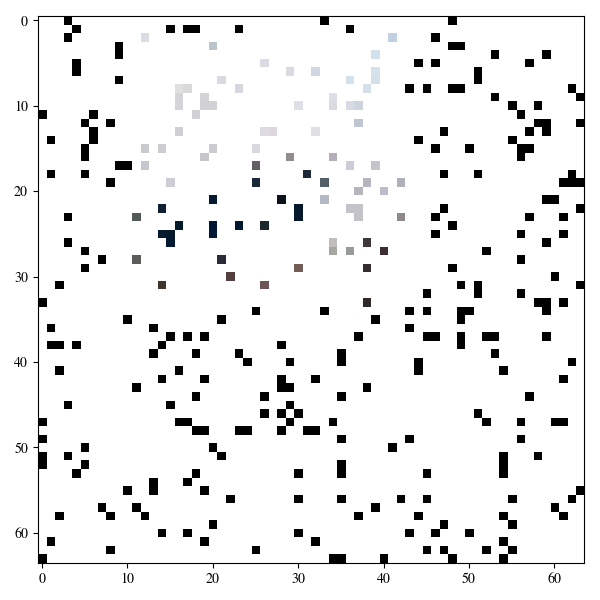}
    \end{subfigure}
    \begin{subfigure}[t]{0.19\textwidth}
        \includegraphics[width=\textwidth,trim={10, 10, 10, 10},clip]{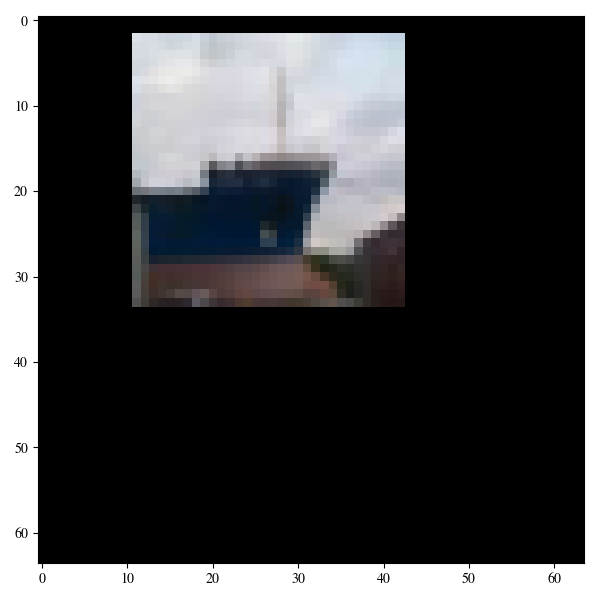}
    \end{subfigure}
    \begin{subfigure}[t]{0.19\textwidth}
        \includegraphics[width=\textwidth,trim={10, 10, 10, 10},clip]{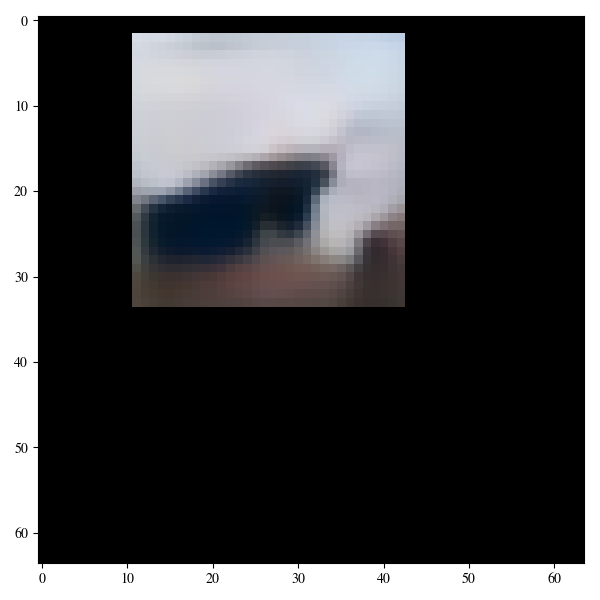}
    \end{subfigure}
    \begin{subfigure}[t]{0.19\textwidth}
        \includegraphics[width=\textwidth,trim={10, 10, 10, 10},clip]{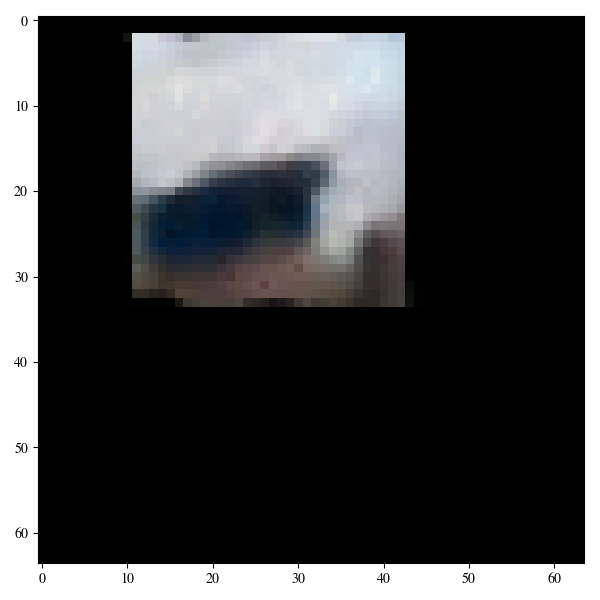}
    \end{subfigure}
    \begin{subfigure}[t]{0.19\textwidth}
        \includegraphics[width=\textwidth,trim={10, 10, 10, 10},clip]{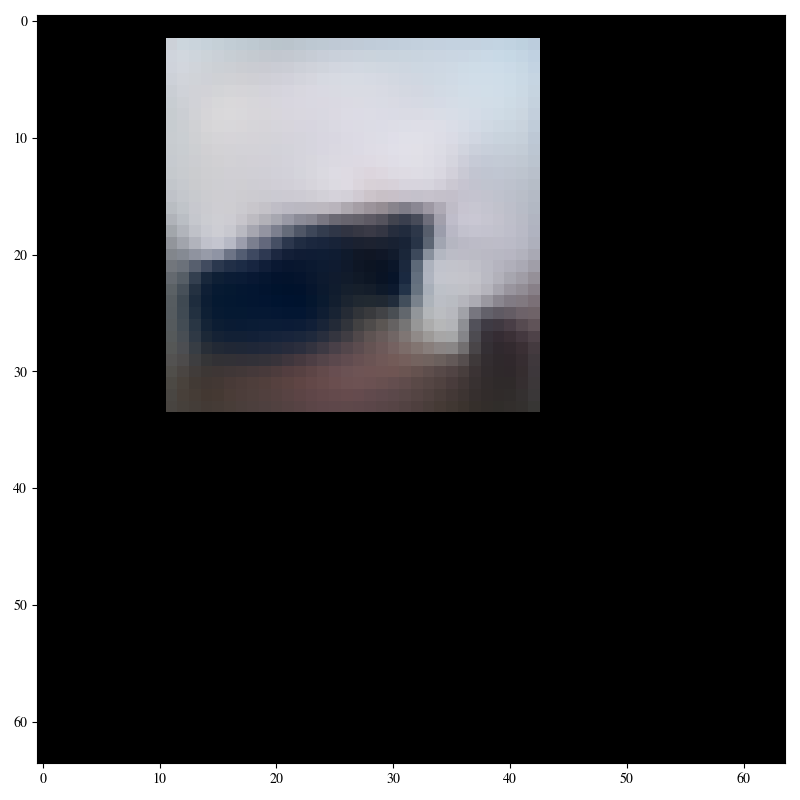}
    \end{subfigure}
    \begin{subfigure}[t]{0.19\textwidth}
        \includegraphics[width=\textwidth,trim={10, 10, 10, 10},clip]{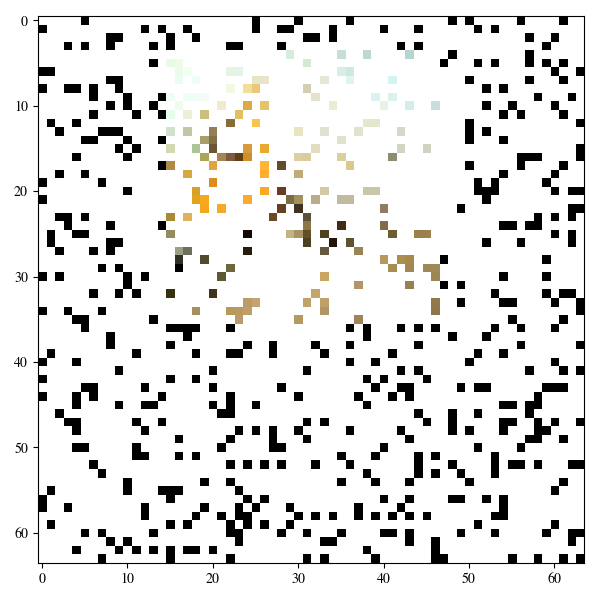}
        \caption{Context dataset.}
    \end{subfigure}
    \begin{subfigure}[t]{0.19\textwidth}
        \includegraphics[width=\textwidth,trim={10, 10, 10, 10},clip]{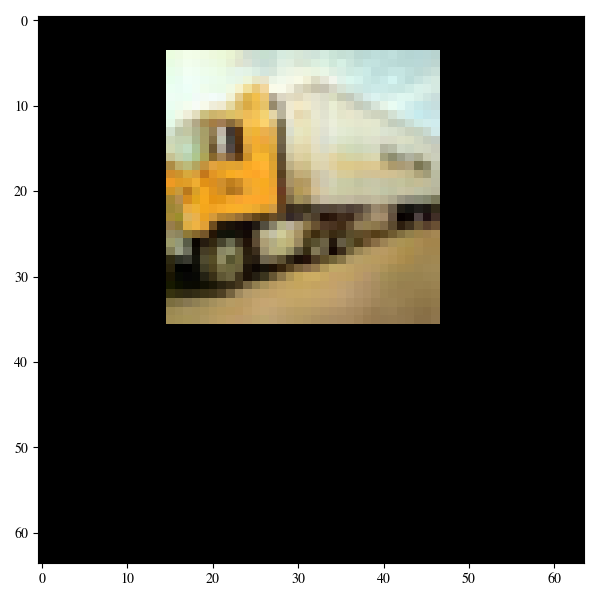}
        \caption{Ground truth data.}
    \end{subfigure}
    \begin{subfigure}[t]{0.19\textwidth}
        \includegraphics[width=\textwidth,trim={10, 10, 10, 10},clip]{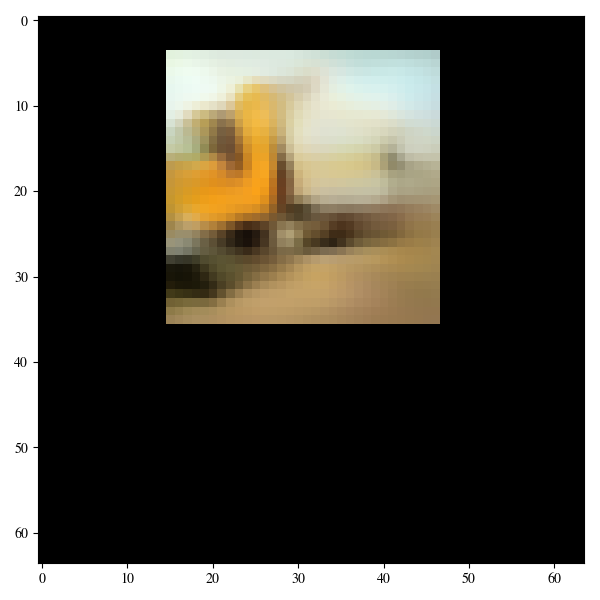}
        \caption{\gls{te-pt-tnp}-M128}
    \end{subfigure}
    \begin{subfigure}[t]{0.19\textwidth}
        \includegraphics[width=\textwidth,trim={10, 10, 10, 10},clip]{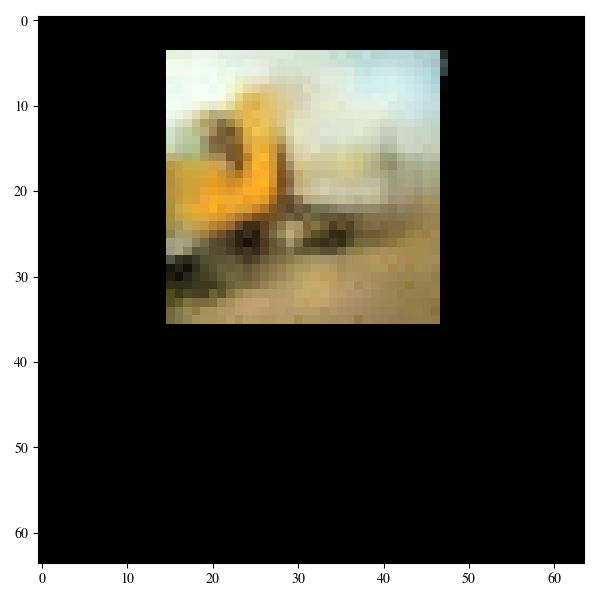}
        \caption{\gls{convcnp}.}
    \end{subfigure}
    \begin{subfigure}[t]{0.19\textwidth}
        \includegraphics[width=\textwidth,trim={10, 10, 10, 10},clip]{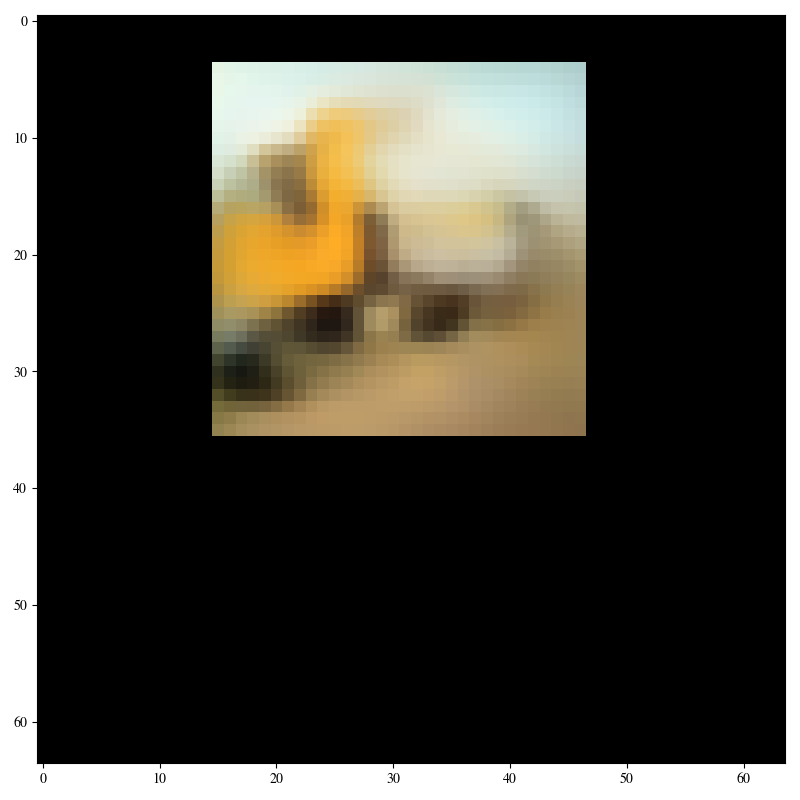}
        \caption{\gls{pt-tnp}-M128.}
    \end{subfigure}
    \caption{A comparison between the predictive mean of the \gls{te-pt-tnp}-M128 model, \gls{convcnp} model and \gls{pt-tnp}-M128 model, given the context datasets on the left.}
    \label{fig:cifar10-translated}
\end{figure*}

\subsection{Kolmogorov Flow}
\label{subapp:kolmogorov-flow}
The 2-D Kologorov flow PDE is defined as
\begin{equation}
    \delta_t \bfu + \nabla \cdot (\bfu \otimes \bfu) = \nu \nabla^2 \bfu - \frac{1}{\rho}\nabla p + \bff
\end{equation}
where $\bfu$ is the velocity field, $\otimes$ the tensor product, $\nu$ the kinematic viscosity, $\rho$ the fluid density, $p$ the pressure field, and $\bff$ the external forcing. We choose a 2-D domain with periodic boundary conditions. We use the data made available by \citet{rozet2023score}, in which the PDE is solved on a $256 \times 256$ grid, coarsened to a $64\times 64$ resolution and integration time between snapshots of $\Delta = 0.2$ time units, with 64 snapshots in total for each trajectory. The overall dataset consists of 1,024 independent trajectories of 64 states, of which 819 are used for training and 102 for testing. We sub-sample $16\times 16\times 16$ regions from these $64 \times 64 \times 64$ trajectories to construct individual tasks. We seek to model the velocity field, $\bfu(\bfx, t) \in \R^2$. For each task, we sample $N_c\sim U(1, 500)$ and set $N_t$ to all remaining points. Input values are normalised to lie in the range $[-3, 3]$.

For all models, we use an embedding / token size of $D_z = 32$. We use a decoder consisting of an MLP with two hidden layers of dimension $D_z$. The decoder parameterises the mean and pre-softplus variance of a Gaussian likelihood with heterogeneous noise. Model specific architectures are as follows:

\paragraph{\gls{pt-tnp}}
Same as \Cref{subapp:1d-regression}. 

\paragraph{\gls{rcnp}}
Same as \Cref{subapp:1d-regression}.

\paragraph{\gls{cnp}}
Same as \Cref{subapp:1d-regression}.

\paragraph{\gls{te-pt-tnp}}
Same as \Cref{subapp:1d-regression}.

\paragraph{Multi-task GP}
For the multi-task GP baseline, we model the covariance between the $i$-th output at $\bfx$ and $j$-th output at $\bfx'$ using the multi-task kernel with diagonal observation noise:
\begin{equation}
    k([\bfx, i], [\bfx', j]) = k_{\text{se}}(\bfx, \bfx') \times k_{\text{tasks}}(i, j) + \sigma_n^2\delta(\bfx - \bfx', i - j).
\end{equation}
where $k_{\text{tasks}}$ is the inter-task covariance, which in this case is a $2\times 2$ lookup table, and $k_{\text{se}}$ is an SE kernel with independent lengthscales for each input dimension:
\begin{equation}
    k_{\text{se}}(\bfx, \bfx') = \sigma^2 \exp\left(-\sum_{i=1}^{D_x}\frac{(x_i - x'_i)^2}{2\ell_i^2}\right) + \sigma^2_n\delta\left(\bfx - \bfx'\right).
\end{equation}
The GPs are implemented using GPytorch \citep{gardner2018gpytorch}, and optimisation of hyperparameters is performed using Adam \citep{kingma2014adam} for 1,000 iterations with a learning rate of $1\times 10^{-1}$.

\paragraph{Training Details}
For all \gls{np} models, we optimise the model parameters using AdamW \citep{loshchilov2017decoupled} with a learning rate of $5\times 10^{-4}$ and batch size of 16 (8 for the \gls{te-pt-tnp} and \gls{rcnp}). Gradient value magnitudes are clipped at 0.5. We train for a maximum of 500 epochs, with each epoch consisting of 10,000 iterations. We evaluate the performance of each model on the entire test set.

\subsection{Environmental Data}
\label{subapp:environmental-data}
The environmental dataset consists of surface air temperatures derived from the fifth generation of the European Centre for Medium-Range Weather Forecasts (ECMWF) atmospheric reanalyses (ERA5) \citep{cccs2020}. The data has a latitudinal and longitudinal resolution of $0.5^\circ$, and temporal resolution of an hour. We consider data collected in 2019, sub-sampled at a temporal resolution of six hours. The training set consists of data within the latitude / longitude range of $[42^{\circ},\ 53^{\circ}]$ / $[8^{\circ},\ 28^{\circ}]$ (roughly corresponding to central Europe), and the test sets consists of two non-overlapping regions: western Europe ($[42^{\circ},\ 53^{\circ}]$ / $[-4^{\circ},\ 8^{\circ}]$), and northern Europe ($[53^{\circ},\ 62^{\circ}]$ / $[8^{\circ},\ 28^{\circ}]$). Individual datasets are obtained by sub-sampling the large regions, with each dataset consists of a $[15, 15, 5]$ grid spanning $7.5^\circ$ across each axis and 30 hours. We also provide surface elevation as additional inputs, such that $D_x = 4$. The inputs and outputs are standardised using the mean and standard deviation values obtained from data within the training region. Each dataset consists of a maximum of $N = 1,125$ datapoints, from which the number of context points are sampled according to $N_c \sim U(\frac{N}{100}, \frac{N}{3})$, with the remaining set as target points.

For all models, we use an embedding / token size of $D_z = 32$. As with the image-completion experiment, we were limited by the hardware available. We use a decoder consisting of an MLP with two hidden layers of dimension $D_z$. The decoder parameterises the mean and pre-softplus variance of a Gaussian likelihood with heterogeneous noise. Model specific architectures are as follows:

\paragraph{\gls{pt-tnp}}
Same as \Cref{subapp:1d-regression}. 

\paragraph{\gls{rcnp}}
Same as \Cref{subapp:1d-regression}.

\paragraph{\gls{cnp}}
Same as \Cref{subapp:1d-regression}.

\paragraph{\gls{te-pt-tnp}}
Same as \Cref{subapp:1d-regression}.

\paragraph{GP}
For the GP baseline, we model the observations using an SE kernel with independent lengthscales for each input dimension plus observation noise:
\begin{equation}
    k(\bfx, \bfx') = \sigma^2 \exp\left(-\sum_{i=1}^{D_x}\frac{(x_i - x'_i)^2}{2\ell_i^2}\right) + \sigma^2_n\delta\left(\bfx - \bfx'\right).
\end{equation}
The GPs are implemented using GPytorch \citep{gardner2018gpytorch}, and optimisation of hyperparameters is performed using Adam \citep{kingma2014adam} for 1,000 iterations with a learning rate of $1\times 10^{-1}$.

\paragraph{Training Details}
For all \gls{np} models, we optimise the model parameters using AdamW \citep{loshchilov2017decoupled} with a learning rate of $5\times 10^{-4}$ and batch size of 16 (8 for the \gls{te-pt-tnp} and \gls{rcnp}). Gradient value magnitudes are clipped at 0.5. We train for a maximum of 500 epochs, with each epoch consisting of 10,000 iterations. We evaluate the performance of each model on the entire test set.

\subsection{The Effectiveness of Dynamically Updating Input Locations}
\label{subapp:dynamic-input-updates}
Here, we perform a simple ablation study to determine the effectiveness of dynamically updating input locations using \Cref{eq:pseudo-locations} and \Cref{eq:te-x-update} in the \gls{te-pt-tnp}. We first consider models trained on the synthetic-1D regression dataset in \Cref{subapp:1d-regression} evaluated on two test sets: one drawn from the same distribution as the train tasks, and another for which the inputs are sampled according to the hierarchical model:
\begin{equation}
\begin{aligned}
    c &\sim \operatorname{Bernoulli}(0.5) \\
    x &\sim \begin{cases}
        U(-4, -1) &\text{if $c = 0$}, \\
        U(1, 4) &\text{if $c = 1$}.
    \end{cases}
\end{aligned}
\end{equation}
In both cases, observations are sampled as described in \Cref{subapp:1d-regression}. The difficulty in this second task is that the bimodality of the input distribution wants the pseudo-locations to also be bimodal, however, the initial pseudo-tokens will have a roughly equal pull in both directions due to symmetry about 0. We therefore posit that it becomes increasingly necessary to dynamically update the input locations to account for this. We consider the \gls{te-pt-tnp}-M32 model described in \Cref{subapp:1d-regression}, and the same \gls{te-pt-tnp}-M32 model with $\psi(\bfu_m, \bfz_n)$ in \Cref{eq:pseudo-locations} set to $\psi(\bfu_m, \bfz_n) = 1 / N$ and no \Cref{eq:te-x-update}. Note that this model is trained from scratch on the same training dataset. The results are shown in \Cref{tab:ablation-synthetic-1d}. We observe that for the uniformly distributed inputs, there is no significant different in the performance of the two models. However, for the bimodal inputs the model without input adjustment performs significantly worse.

\begin{table}[htb]
    \centering
    \caption{Average log-likelihood (\textcolor{OliveGreen}{$\uparrow$}) on the two test datasets.}
    \begin{tabular}{r c c}
    \toprule
         Model & Uniform Inputs & Bimodal Inputs \\
         \midrule
         \gls{te-pt-tnp}-M32 & $\mathbf{-0.44 \pm 0.01}$ & $\mathbf{-0.67 \pm 0.01}$ \\
         \gls{te-pt-tnp}-M32 no input adjustment & $\mathbf{-0.46 \pm 0.01}$ & $-0.76 \pm 0.01$ \\
         \bottomrule
    \end{tabular}
    \label{tab:ablation-synthetic-1d}
\end{table} 


\end{document}